\documentclass{article} 
\usepackage[preprint]{colm2026_conference}

\usepackage{microtype}
\usepackage{hyperref}
\usepackage{url}
\usepackage{booktabs}

\usepackage[T1]{fontenc}
\usepackage{afterpage}

\usepackage{lineno}

\definecolor{darkblue}{rgb}{0, 0, 0.5}
\hypersetup{colorlinks=true, citecolor=darkblue, linkcolor=darkblue, urlcolor=darkblue}

\usepackage{multirow}
\usepackage{tabularx}
\usepackage{amsmath,amssymb}

\usepackage{algorithm}
\usepackage{algorithmic}
\newcommand{\abstart}{\texttt{<beginabstract>}}
\newcommand{\abend}{\texttt{<endabstract>}}
\newcommand{\atok}[1]{\texttt{<TOKEN\_#1>}}
\usepackage{float}
\usepackage{caption}
\usepackage{subcaption}

\usepackage{tikz}
\usetikzlibrary{arrows.meta, shapes, positioning, fit, calc, backgrounds}   

\usepackage{titletoc}

\newcommand{\atsep}{\hspace{0pt}\ }

\usepackage[dvipsnames]{xcolor}
\usepackage{graphicx}
\usepackage[most]{tcolorbox}

\definecolor{ultralightgray}{gray}{0.97}

\newtcolorbox{promptbox}[1][]{
  breakable,           
  enhanced,
  colback=ultralightgray, 
  colframe=ultralightgray,   
  boxrule=0pt,           
  arc=0pt,                 
  left=3pt,right=3pt,      
  top=3pt,bottom=3pt,      
  fontupper=\tt\normalsize,
  #1                        
}

\usepackage{dashrule}

\title{Thinking Without Words: \\ Efficient Latent Reasoning with Abstract Chain-of-Thought}

\author{Keshav Ramji\thanks{Correspondence to \texttt{keshav.ramji@ibm.com}}~, ~Tahira Naseem \& Ramón Fernandez Astudillo \\
IBM Research AI
}

\begin{document}

\ifcolmsubmission
\linenumbers
\fi

\maketitle
\thispagestyle{fancy}
\fancyhead[L]{}
\afterpage{\fancyhead[L]{\small{Thinking Without Words:
Efficient Latent Reasoning with Abstract Chain-of-Thought}}}

\begin{abstract}
While long, explicit chains-of-thought (CoT) have proven effective on complex reasoning tasks, they are costly to generate during inference. Non-verbal reasoning methods have emerged with shorter generation lengths by leveraging continuous representations, yet their performance lags behind verbalized CoT. 
We propose \textbf{Abstract Chain-of-Thought}, a discrete latent reasoning post-training mechanism in which the language model produces a short sequence of tokens from a reserved vocabulary in lieu of a natural language CoT, before generating a response. To make previously unseen “abstract” tokens useful, we introduce a policy iteration-style warm-up loop that alternates between (i.) bottlenecking from a verbal CoT via masking and performing supervised fine-tuning, and (ii.) self-distillation by training the model to generate abstract tokens from the prompt alone via constrained decoding with the codebook. After warm-up, we optimize the generation of abstract sequences with warm-started reinforcement learning under constrained decoding. Abstract-CoT achieves up to $11.6\times$ fewer reasoning tokens while demonstrating comparable performance across mathematical reasoning, instruction-following, and multi-hop reasoning, and generalizes across language model families. We also find an emergent power law distribution over the abstract vocabulary, akin to those seen in natural language, that evolves across the training phases. 
Our findings highlight the potential for post-training latent reasoning mechanisms that enable efficient inference through a learned abstract reasoning language. 

\end{abstract}

\section{Introduction}

Large language models (LLMs) increasingly rely on long, explicit chains-of-thought (CoTs) to solve complex, multi-step reasoning problems. Despite its effectiveness, verbalized CoT \citep{10.5555/3600270.3602070, 10.5555/3600270.3601883} is an expensive mechanism, increasing latency and cost at inference while bloating the length of traces during reinforcement learning (RL). Prior works also suggest that verbalized CoT can be unfaithful \citep{lanham2023measuringfaithfulnesschainofthoughtreasoning, turpin2023language}, while leveraging a different latent reasoning process that is not communicated. These drawbacks have motivated approaches to compress or internalize natural language CoT with more efficient intermediate representations \citep{cheng2024compressedchainthoughtefficient, deng2024explicitcotimplicitcot}. Simultaneously, approaches focusing on pause or filler tokens \citep{goyal2024think, pfau2024lets} suggest that their addition facilitates deliberate internalized thinking through its activations. 
Furthermore, the findings of DeepSeek-R1-Zero \citep{Guo2025} indicate that strong performance can be separable from human-readability, demonstrating gains even with language mixing in the CoTs. 
Recent works such as Coconut \citep{hao2025training} have sought to enable reasoning mechanisms through continuous concept spaces, balancing efficiency and expressivity through principled methods for internalized recurrence.

In this work, we study a simple question: can we replace long verbalized rationales with a short sequence of discrete abstract tokens that functions as a latent scratchpad, while retaining the performance gains of CoT in response generation? We find that not only is this possible, but it can be achieved purely through post-training instruction-tuned models. We propose \textbf{Abstract Chain-of-Thought (Abstract-CoT)}: instead of generating natural language reasoning, we induce the model to emit a bounded-length sequence of tokens from a reserved abstract vocabulary of distinguishable filler tokens. Abstract-CoT is designed to be token-efficient and non-verbal, producing short intermediate traces while offering an alternative to rationales generated in natural language.

However, adding previously unseen tokens creates a cold-start problem, as their embeddings are randomly initialized and meaningless initially. While these tokens appear semantically uninformative, our recipe aims to learn to produce a sequence of these tokens, inducing new pathways between a prompt and a response. To this end, we adopt a two-stage training recipe. The first stage is a policy iteration warm-up, alternating between verbal CoT guidance and direct on-policy generation of abstract token sequences. In the former, the final response only attends to the abstract tokens, not to the verbal CoT, forcing abstract token representations to learn useful information from the verbal CoT, serving as an information bottleneck. 
We then perform self-distillation by discarding the verbal CoTs and training only with on-policy-generated abstract sequences with the learned representations, and repeat this process iteratively. In the second stage, we apply reinforcement learning with a generative reward model to induce exploration over abstract token sequences and refine the abstract generation policy. Our findings demonstrate substantial gains in token efficiency while matching or outperforming verbalized chain-of-thought. 

We summarize our contributions below:

\begin{figure}
    \centering
    \includegraphics[width=0.9924\linewidth]{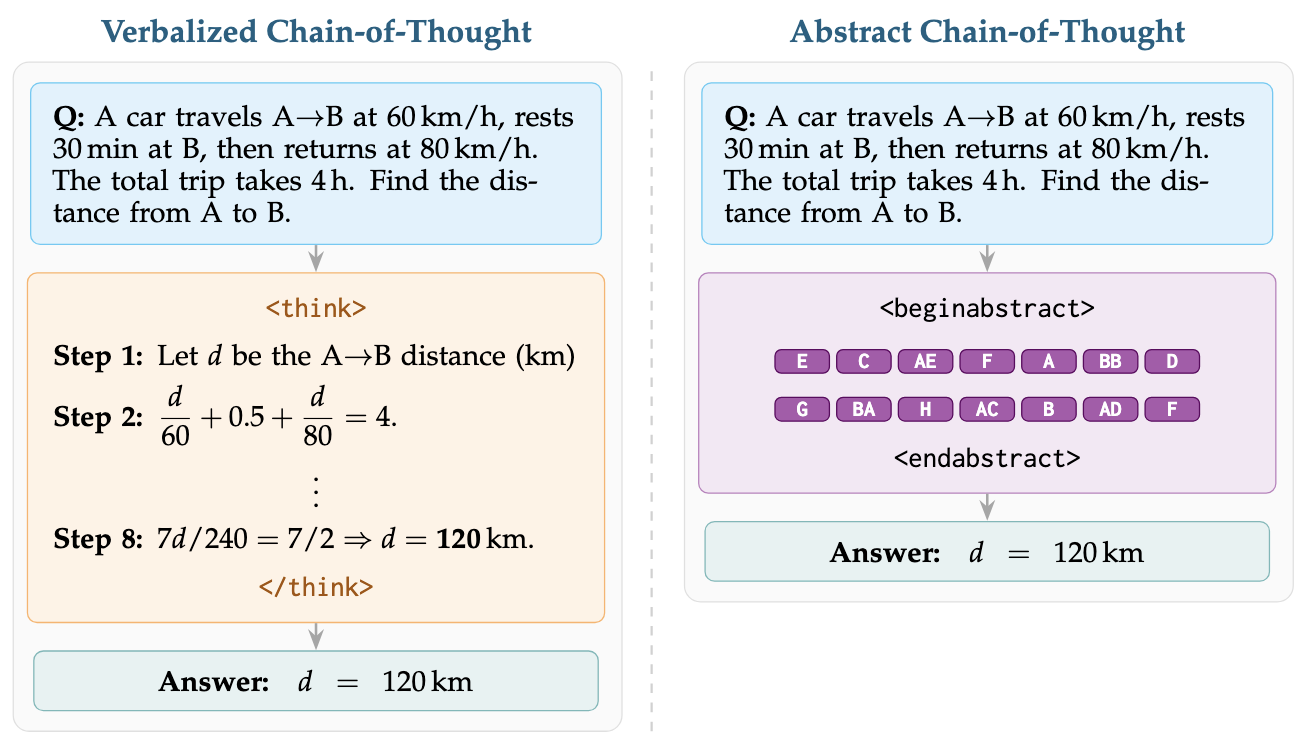}
    \caption{\textbf{Verbalized vs.\ Abstract Chain-of-Thought.} \emph{Verbalized CoT} (left) generates an explicit natural language
  rationale (Step~1 through Step~8) inside \texttt{<think>}\,$\cdots$\,%
  \texttt{</think>} tags before producing the answer.
  \emph{Abstract CoT} (right) instead emits a short sequence of tokens 
  from the reserved abstract vocabulary inside
  \abstart\,$\cdots$\,\abend\ delimiters, achieving the same answer
  with substantially fewer reasoning tokens.}
    \label{fig:cot-comparison}
\end{figure}

\begin{enumerate}
    \item \textbf{Abstract Chain-of-Thought}: We propose Abstract-CoT, a mechanism for reasoning through a vocabulary of reserved tokens introduced entirely in LLM post-training. 
    \item \textbf{Warm-up via Policy Iteration:} We warm up the embeddings of the reserved tokens by alternating bottlenecked SFT and self-distillation, yielding an abstract generator.
    \item \textbf{Warm-started RL for Abstract Policies:} We optimize generation of abstract traces using GRPO, with constrained decoding to the abstract vocabulary.
    \item \textbf{Token Efficiency:} Abstract-CoT  reduces reasoning tokens up to $11.6\times$ while matching verbalized CoT performance on MATH-500, AlpacaEval, and HotpotQA. 
    
    \item \textbf{Abstract Reasoning Language:} We observe power-law dynamics over the abstract vocabulary, indicating that meaningful concepts and re-use patterns are learned. 
\end{enumerate}

\section{Related Work}

\subsection{Filler Tokens}

Works on filler tokens augment the token sequence with special tokens that are semantically uninformative (not human-readable natural language), but expand the model's effective computation in the forward pass. \citet{goyal2024think} introduce \texttt{<pause>} tokens, showing that explicitly allocating intermediate tokens to allow for ''thinking time'' can improve reasoning. \citet{mu2023learning} propose \emph{gist tokens} which serve as a learned bottleneck, summarizing longer contexts into a small set of activations that can be cached and reused, introducing contextual ''slots'' that can hold task-relevant information. Other works suggest that such tokens can be used to expand expressivity limits \citep{pfau2024lets, merrill2025exact, london2025pause}, and used to cache preceding context for long-context retrieval \citep{shah2025languagemodelinglearnedmetatokens}. Our work also relates to parameter-efficient methods that seek to optimize continuous prompt embeddings to steer generations \citep{lester2021prompttuning,li2021prefixtuning}, and token-space interventions that add intermediate representations \citep{jang2025expandingcomputationspaces}. 
While our abstract tokens are introduced specifically as a latent reasoning trace (rather than as prompt compression), they share a similar essence that a small number of lightweight extra positions can be trained to store or carry additional information, offering a new reasoning medium.

\subsection{CoT Compression, Distillation, and Discrete Codebooks}

Compression, distillation, and partially removing the textual rationale (often through staged curricula) are some of the key mechanisms used to target the verbosity and cost of verbalized CoT. Early works such as \citet{hsieh2023distillingstepbystep} demonstrated that explicit step-by-step rationales can be distilled to smaller models. Recent methods include seeking to directly shorten verbalized CoT via multi-round refinement \citep{yan2025macc} and learning to skip intermediate reasoning tokens in a controllable fashion while retaining generation quality \citep{xia2025tokenskip}.

Approaches that compress parts of the rationale into a learned discrete or quantized representation are somewhat related to our discrete codebook.  \citet{su2025tokenassorted} combines latent tokens (learned via vector quantization) with remaining text tokens, inducing an efficiency-interpretability trade-off. Complementary works perform step-wise compression of CoT into latent tokens \citep{zhang2025lightthinker} as well as gradually internalizing explicit steps into implicit computation \citep{deng2024explicitcotimplicitcot} in a curriculum fashion. By contrast, our abstract tokens are not a quantized reconstruction of a teacher rationale, but are entirely in a newly introduced reserved vocabulary, with the model trained to use it as a compact reasoning language under constrained decoding. This allows the model to potentially explore \textit{other reasoning pathways}, rather than being constrained to that of the teacher CoT. 

\subsection{Continuous and Hybrid Latent Reasoning}

Some recent approaches seek to replace parts of the textual rationale with continuous thought states. Coconut \citep{hao2025training} replaces some CoT tokens with continuous latent vectors derived from hidden states and trains the language model with a curriculum that gradually increases the latent segment and replaces verbalized CoT segments. CODI \citep{shen-etal-2025-codi} similarly compresses CoT into a continuous space, using self-distillation to align latent trajectories with those induced by explicit rationales. System-1.5 reasoning \citep{wang2025system} introduces dynamic shortcuts, traversing between language and latent spaces while aiming to reduce unnecessary verbal reasoning and retain controllability.

Related ``soft'' thinking approaches propagate intermediate representations by feeding distributions over embeddings as subsequent inputs \citep{xu-etal-2025-softcot,zhang2025softthinkingunlockingreasoning}. Recent works such as \citet{butt2025softtokenshardtruths} study training and optimization stability when such soft tokens are treated as decision variables through RL. Hybrid methods such as HybridCoT \citep{shen2026hybridcot} explicitly interleave latent and text tokens to balance efficiency with partial interpretability. In our work, we suggest that it is possible to achieve the efficiency gains associated with latent reasoning while operating fully in the discrete token space. 

\subsection{Reinforcement Learning for Budget Control}

A complementary direction focuses on controlling inference-time cost by explicitly optimizing the reasoning budget, which is often operationalized as the length of intermediate reasoning traces.  Recent work applies RL to learn when to expend additional reasoning steps as opposed to answering early; for example, by learning adaptive chain-of-thought triggering policies under compute constraints \citep{lou2025adacotparetooptimaladaptivechainofthought}, or by pruning or shortening intermediate reasoning via training-time objectives that directly reward efficiency \citep{hou2026thinkprune}. Other recent approaches optimize a length-accuracy trade-off with RL objectives, by allocating token budgets dynamically \citep{kleinman2025e1learningadaptivecontrol} or by explicitly optimizing for consistency with user-specified length constraints \citep{aggarwal2025l}.

While our RL stage is most closely aligned with this line of work, it differs in the action space: instead of optimizing over free-form textual CoT length, we optimize over sequences constrained to a reserved discrete codebook. This enables control over the intermediate sequence while avoiding the brittleness of length control in open-ended natural language.

\section{Latent Reasoning with Abstract Chain-of-Thought}
\label{sec:method}

\subsection{Problem Setup and Notation}

Let $x, c,$ and $y$ denote a prompt, gold verbal chain-of-thought (CoT), and the target answer, respectively. We assume training data $\mathcal{D} = \{(x_i, c_i, y_i)\}_{i=1}^N$, where $c_i$ is only available during the first phase of warm-up. Let $\pi_{\theta}$ be a causal decoder-only LLM with parameters $\theta$ and base vocabulary $\mathcal{V}$. We extend the tokenizer with a set of $M$ previously unseen (reserved) tokens in the abstract codebook, along with two delimiters $\abstart$ and $\abend$, marking the abstract reasoning segment\footnote{We use alphabetical names $\atok{A},\dots,\atok{Z}$; for $M > 26$, we continue with two letter identifiers (AA-ZZ), which may be similarly extended for larger abstract vocabularies.}:
\[
\mathcal{V}_\mathrm{abs} = \{\atok{A},\atok{B},\dots,\atok{Z},\atok{AA},\dots\},
\]
Thus, an abstract chain-of-thought is a token sequence $z = (z_1, \dots, z_m) \in \mathcal{V}_{abs}^m$, formatted as:
\[
    \tilde{z} = \abstart~ z_1~ z_2~ \dots~ z_m~ \abend
\]
We denote the maximum length of the abstract sequence by $m \leq m_{\text{max}}$; at inference-time, the model receives $x$ and must generate $\tilde{z}$ and $y$ without access to $c$. Let $\mathcal{Z}$ denote the positions of the full abstract sequence $\tilde{z}$ (including $\abstart$ and $\abend$), and let $\mathcal{Z}_{\text{abs}} \subseteq \mathcal{Z}$ denote the positions of the $m$ codebook tokens $z_1, \dots, z_m \in \mathcal{V}_{\text{abs}}$.

We view the abstract trace $z$ as a discrete latent variable, mediating reasoning. Ideally, we would like to maximize the marginal likelihood:
\begin{equation}
   \log \pi_{\theta}(y \mid x) = \log \sum_{z \in \mathcal{V}_{\text{abs}}^{*}} \pi_{\theta}(z \mid x)\pi_{\theta}(y \mid x, z) 
\end{equation}
for a sequence of length $\leq m_{\text{max}}$, but the sum over discrete traces is intractable. Therefore, Abstract Chain-of-Thought uses a bootstrapping procedure that alternates (i) proposing an abstract trace $z \in \mathcal{V}_{\mathrm{abs}}^*$ with verbal CoT guidance, and (ii) updating the model given the generated trace, followed by distillation to learn to directly propose traces from $x$ alone.

\begin{figure}
    \centering
    \includegraphics[width=\linewidth]{figs/new-algo-figure.png}
    \caption{\textbf{Abstract Chain-of-Thought}: The training recipe consists of two stages: (i.) a warm-up loop, consisting of a Bottlenecked SFT phase with guidance from a teacher Verbal CoT, and a Self-Distillation phase with on-policy abstract sequence generation, repeated iteratively, and (ii.) reinforcement learning using GRPO with constrained decoding for the rollouts, which rewards abstract sequences that lead to a high-quality response.}
    \label{fig:algorithm}
\end{figure}

\subsection{Warm-Up via Policy Iteration}\label{sec:warm-up}

The abstract tokens start with randomly initialized embeddings, so the model initially cannot exploit the bottleneck in the absence of a prior that enforces specific concept mappings. Therefore, we perform an \textbf{abstract embedding warm-up} with a policy iteration loop over iterations $t = 1, \dots, T$; each iteration produces a dataset of abstract trajectories $\tilde{z}^{(t)}$ and updates $\theta$ via SFT. The training dataset $\mathcal{D}$ is staged over the iterations: $\mathcal{D} = \bigcup\limits_{t=1}^T ~ \{(\mathcal{D}_{t,1}, \mathcal{D}_{t,2})\}$).

\paragraph{Constrained Decoding. } We use $\pi_{\theta}^{\text{abs}}$ to denote the policy restricted to an allowed token set $\mathcal{A} = \mathcal{V}_{\text{abs}} \hspace{1mm} \cup \{\abend\}$ which can be generated. At each step $i$, with context $h = x  ~ \cup \{\abstart\} ~ \cup ~ \tilde{z}_{<i}$, and 
$\pi_{\theta}^{\text{abs}}(a \mid h) = \frac{\pi_{\theta}(a \mid h) \textbf{1}[a \in \mathcal{A}]}{\sum\limits_{u \in \mathcal{A}} \pi_{\theta}(u \mid h)}$. We enforce a hard cap $m_{\max}$ of codebook tokens that may be generated; if $m= m_{\max}$, we force the end delimiter to be generated, then allow the response to be produced without constrained decoding. 

\paragraph{(1) Bottlenecked SFT with Abstract Tokens.}

Given $(x,c,y)$, we construct $\tilde{z}^{(t)}$ using a policy $\phi_t$. In the first iteration, we use random initialization; with $\mathcal{S}$ steps in the verbal CoT, we sample a random number of abstract tokens per CoT step ($\text{rand}(1, \frac{|\ell|}{2})$ for $|\ell|$ tokens in step $\ell \in \mathcal{S}$) and choosing the specific tokens uniformly at random from $\mathcal{V}_{\text{abs}}$. We analyzed other initialization schemes (alphabetically cycling through the tokens, enforcing a power-law distribution), and found a uniform distribution over the tokens to be most effective. In subsequent iterations ($t \geq 2$), abstract sequences are generated on-policy: $\tilde{z}^{(t)} \sim \pi_{\theta}^{\text{abs}}(\cdot \mid x, c)$ under constrained decoding. 

We form a single concatenated training sequence $s = [x~ ; c~; \tilde{z}~ ; y]$
and define a block-structured attention mask $\mathcal{A}$ that enforces an information bottleneck. Let indices be partitioned into prompt ($\mathcal{X}$), verbal CoT ($\mathcal{C}$), abstract sequence ($\mathcal{Z}$) and answer ($\mathcal{Y}$). The \textbf{abstract tokens attend to the prompt and the verbal CoT}; that is:
\[
\mathcal{A}_{i,j} = 1 \hspace{2mm} \forall \hspace{1mm} i \in \mathcal{Z}, j \in \mathcal{X} \hspace{0.5mm} \cup \hspace{0.5mm} \mathcal{C} \hspace{0.5mm} \cup \hspace{0.5mm}\mathcal{Z}_{\leq i}
\]
Crucially, the \textbf{answer only attends to the prompt and the abstract tokens, \textit{not} to the verbal CoT}, with all other entries following standard causal masking:
\[
\mathcal{A}_{i,j} = \begin{cases} 
      1 & i \in \mathcal{Y}, j \in \mathcal{X} \hspace{0.5mm} \cup \mathcal{Z} \hspace{0.5mm} \cup \mathcal{Y}_{\leq i} \\
      0 & i \in \mathcal{Y}, j \in \mathcal{C}
   \end{cases}
\]
Concretely, this training procedure can be seen as implementing a discrete latent bottleneck; let $H_{\mathcal{Z}_{\mathrm{abs}}}$ denote the hidden states at the abstract token positions $\mathcal{Z}_{\mathrm{abs}}$ produced from the prefix $[x;c; \tilde{z}]$ following masking of the verbal CoT. The only dependence of answer generation ($y$) on the verbal CoT ($c$) is through $H_{\mathcal{Z}_{\mathrm{abs}}}$, inducing conditional Markov structure: 
\[
C \rightarrow H_{\mathcal{Z}_{\text{abs}}} \rightarrow Y \hspace{2mm} \text{(conditioned on $X$ and $Z$)} 
\]
By the data processing inequality, any dependence between $y$ and $c$ must be bounded by the information that can be transmitted through the abstract segment:
\begin{equation}
I(C;Y \mid X,Z) \leq I(C; H_{Z_{\text{abs}}} \mid X, Z)
\end{equation}
Since $H_{\mathcal{Z}_{\mathrm{abs}}}$ scales linearly with the abstract sequence length $m$, tuning $m_{\mathrm{max}}$ affects the channel capacity from $c$ to $y$ during warm-up. 

We then optimize a masked SFT objective that trains on the abstract sequence\footnote{Optionally, $\tilde{z}^{(t)}$ can be treated as fixed in this stage (without backpropagating on the abstract tokens), which would update the embeddings solely through gradients flowing from the answer loss. } and the answer while hiding the verbal CoT with bottleneck attention mask $\mathcal{A}$:

\begin{equation}
  \mathcal{L}_{\mathrm{SFT}}(\theta;\,x,c,\tilde{z},y)
  \;=\;
  -\sum_{j\in (\mathcal{Z}_{\mathrm{abs}}\cup \mathcal{Y})}
  \log \pi_\theta\!\left(s_j \mid s_{<j};\,\mathcal{A}\right),
\end{equation}

\begin{algorithm}[t]
\caption{: Policy Iteration Warm-Up for Abstract-CoT}
\label{alg:pi}
\begin{algorithmic}[1]
\REQUIRE Training data $\mathcal{D}=\{(x,c,y)\}$, abstract vocabulary $\mathcal{V}_\mathrm{abs}$, iterations $T$
\STATE Initialize $\theta^{(0)}$ from base instruction-tuned model; add new token embeddings for $\mathcal{V}_\mathrm{abs}$
\FOR{$t=1$ \textbf{to} $T$}
  \STATE Data for current iteration:
  $\mathcal{D}_{t, 1}, \mathcal{D}_{t,2} \subset \mathcal{D}^{(t)}$
  \STATE Generate abstract traces $\tilde{z}^{(t)} \sim \phi_t(\cdot \mid x, c, \theta^{(t-1)})$ (random if $t{=}1$, else constrained decoding with $\phi_t = \pi_{\theta^{(t-1)}}$for $(x,c,y) \in \mathcal{D}_{t,1}$
  \STATE Update $\bar{\theta}^{(t)} \leftarrow \arg\min\limits_\theta \mathbb{E}_{(x,c,y)\sim\mathcal{D}_{t,1}}\big[\mathcal{L}_{\mathrm{SFT}}(\theta;x,c,\tilde{z}^{(t)},y; A)\big]$
  \STATE Distill: generate $\tilde{z}'\sim \pi_{\bar{\theta}^{(t)}}^{\mathrm{abs}}(\cdot\mid x)$ for $(x,y) \in \mathcal{D}_{t,2}$
  \STATE Starting from $\bar{\theta}^{(t)}$, update $\theta^{(t)} \leftarrow \arg\min\limits_\theta \mathbb{E}_{(x,y)\sim\mathcal{D}_{t,2}}\big[\mathcal{L}_{\mathrm{Distill}}(\theta;x,\tilde{z}',y)\big]$
\ENDFOR
\STATE \textbf{return} $\theta^{(T)}$
\end{algorithmic}
\end{algorithm}

\paragraph{(2) Self-Distillation Without Verbal CoT.} The bottlenecked SFT stage exploits $c$ to shape the hidden states at abstract-token positions, but our target policy ultimately should produce abstract tokens from the prompt alone; this motivated the loss computation on $\tilde{z}^{(t)}$. We create a distillation dataset by generating $\tilde{z} \sim \pi_{\theta}^{\text{abs}}(\cdot \mid x)$ via constrained decoding (with $m \leq m_{\text{max}}$) and pairing it with the gold answer $y$:
$
\mathcal{D}_{\text{distill}}^{(t)} = \{(x_i, \tilde{z}_i, y_i)\}_{i=1}^N
$.
In the discrete latent bottleneck interpretation, the self-distillation and RL (Section \ref{section3.3-rl}) phases tune the model's inference-time thinking budget. 
We train with standard causal SFT on [$x; \tilde{z}; y$], where $s_j$ spans the abstract and response tokens:

\begin{equation}
    \mathcal{L}_{\text{Distill}}(\theta; x, \tilde{z}, y) = - \sum\limits_{j \hspace{0.5mm}\in (\mathcal{Z}_\mathrm{abs} \hspace{0.5mm}\cup\hspace{0.5mm} \mathcal{Y})} \log \pi_{\theta}(s_j \mid s_{<j})
\end{equation}

\subsection{Reinforcement Learning from Warm-Start}\label{section3.3-rl}
After the warm-up stage, we optimize the abstract-token policy with RL; we refer to this as warm-started RL. In practice, this is implemented in a similar manner as the warm-up self-distillation phase: (1) generate $\tilde{z}$ under a guided-regex constraint, and (2) append \abend~ and decode $y$ unconstrained. Our default GRPO \citep{shao2024deepseekmathpushinglimitsmathematical} updates include log-probabilities for both the abstract trace and the answer tokens, improving response quality after RL in addition to shaping the intermediate abstract sequence\footnote{Alternatively, updates can be isolated to just the abstract tokens along with a fixed decoding rule.}.
We use a generative reward model -- specifically, \texttt{gpt-oss-20b} \citep{openai2025gptoss120bgptoss20bmodel} -- to score outputs in our experiments, for our recipe to generalize to non-verifiable, natural language settings.

For each prompt $x$, we sample a group of $K$ trajectories $\{(\tilde{z}_k, y_k)\}_{k=1}^K$ by first drawing $\tilde{z}_k \sim \pi_{\theta}^{\text{abs}}(\cdot \mid x)$, then $y_k \sim \pi_{\theta}(\cdot \mid x, \tilde{z}_k)$, and computing rewards $\hat{R}_k = \hat{R}(x, \tilde{z}_k, y_k)$. We define advantages:
\[
A_k = \frac{\hat{R}_k - \text{mean}(\hat{R}_{1:K})}{\text{std}(\hat{R}_{1:K}) + \epsilon}
\]
We update $\theta$ via applying GRPO to an action space over $(\tilde{z}, y)$:
\begin{align}
\mathcal{J}(\theta)
&=\mathbb{E}_{x}\Bigg[
\frac{1}{K}\sum_{k=1}^K A_k \Big(
\sum_{t\in \mathcal{Z}_{\mathrm{abs}}}\log \pi_\theta^{\mathrm{abs}}\!\left(z_{k,t}\mid x, z_{k,<t}\right)
+ \sum_{t\in \mathcal{Y}}\log \pi_\theta\!\left(y_{k,t}\mid x,\tilde{z}_k,y_{k,<t}\right)
\Big) \nonumber \\
&\qquad
- \beta\,\mathrm{KL}\!\left(
\pi_\theta^{\mathrm{abs}}(\tilde{z}\mid x)\,\pi_\theta(y\mid x,\tilde{z})
~\Big\|~
\pi_{\theta_{\mathrm{ref}}}^{\mathrm{abs}}(\tilde{z}\mid x)\,\pi_{\theta_{\mathrm{ref}}}(y\mid x,\tilde{z})
\right)
\Bigg].
\end{align}
where $\pi_{\theta_{\text{ref}}}$ is the reference policy (the warm-started model). KL regularization is applied over both the abstract and the response distributions. 

\section{Results}

\subsection{Experimental Setup}\label{sec:expt-setup}

\paragraph{Datasets.} We train using subsampled data (600k samples) from the Dolci-Think-SFT dataset \citep{allenai_dolci_think_sft_7b_dataset} for the policy iteration warm-up loop and from Dolci-Think-RL \citep{allenai_dolci_think_rl_7b_dataset} for reinforcement learning; these were used in the development of the Olmo 3 Think models \citep{olmo2025olmo3}. The former contains prompts paired with gold verbal CoT and gold answers, while we only use prompts and gold responses from the latter. We report results on MATH-500 \citep{hendrycks2021measuring}, AlpacaEval-LC-2.0 \citep{dubois2024length}, and HotpotQA (500 samples; \citet{yang-etal-2018-hotpotqa}),  spanning mathematics (verifiable), general instruction-following, and multi-hop question-answering (non-verifiable) settings, as well as more challenging, reasoning-intensive benchmarks like AIME'25 \citep{aime2025} and GPQA-Diamond \citep{rein2024gpqa} involving math and natural language. 

\paragraph{Models.} We consider three instruction-tuned models, Qwen3-8B \citep{yang2025qwen3technicalreport}, Qwen3-4B, and Granite-4.0-Micro (3B)\footnote{\href{https://huggingface.co/ibm-granite/granite-4.0-micro}{https://huggingface.co/ibm-granite/granite-4.0-micro}}. We also include ablations with Qwen3-32B in Appendix \ref{appendix:Qwen32B}. We evaluate models without their "thinking mode" with standard CoT prompting to ensure controlled comparison, and report both performance and intermediate token (reasoning/rationale) length. For Abstract-CoT, we extend each tokenizer with the abstract vocabulary, train new embeddings during warm-up, and decode constrained abstract traces up to $m_{\text{max}}$ tokens. We use $T = 3$ policy iteration warm-up rounds.

\begin{table*}[t]
\centering
\small
\begin{tabular}{lcc|cc|cc}
\toprule
\multirow{2}{*}{Method} & \multicolumn{2}{c|}{MATH-500} & \multicolumn{2}{c|}{AlpacaEval} & \multicolumn{2}{c}{HotpotQA} \\
 & Accuracy & Tokens & Win-rate & Tokens & F1 & Tokens \\
\midrule
\multicolumn{7}{c}{\textbf{Qwen3-8B}} \\
\midrule
Baseline & 82.4 & 1205 & 52.4 & 322 & 51.1 & 527 \\
Pause Tokens & 78.6 & 142 & 46.7 & 212 & 49.0 & 173 \\
Stepwise Internalization & 88.6 & 169 & 55.3 & 245 & 52.7 & 136 \\
SFT (no CoT) & 85.8 & 394 & 54.9 & 298 & 51.3 & 331 \\
SFT (CoT) & 89.8 & 1522 & 57.0 & 460 & 54.8 & 679 \\
SFT+RL & \textbf{92.6} & 1671 & \underline{58.4} & 496 & \underline{58.1} & 735 \\
\midrule
Abstract-CoT (RL-only) & 82.0 & 118 & 50.4 & 229 & 49.0 & 149 \\
Abstract-CoT (Warm-up) & 88.0 & 173 & 55.9 & 251 & 53.7 & 174 \\
Abstract-CoT (Warm-up + RL) & \underline{90.8} & 144 & \textbf{60.8} & 225 & \textbf{58.8} & 171 \\
\midrule
\multicolumn{7}{c}{\textbf{Qwen3-4B}} \\
\midrule
Baseline & 83.2 & 1087 & 49.8 & 298 & 46.7 & 482 \\
Pause Tokens & 80.6 & 138 & 42.1 & 198 & 44.2 & 162 \\
Stepwise Internalization & 84.6 & 129 & 51.4 & 174 & 45.8 & 131 \\
SFT (no CoT) & 85.4 & 362 & 53.7 & 276 & 46.9 & 308 \\
SFT (CoT) & 88.4 & 1396 & 55.2 & 431 & 50.3 & 612 \\
SFT + RL & \textbf{91.2} & 1523 & \underline{57.1} & 467 & \underline{53.4} & 683 \\
\midrule
Abstract-CoT (RL-only) & 81.0 & 114 & 47.8 & 217 & 43.1 & 142 \\
Abstract-CoT (Warm-up) & 86.2 & 168 & 54.3 & 238 & 48.6 & 167 \\
Abstract-CoT (Warm-up + RL) & \underline{89.8} & 141 & \textbf{58.7} & 213 & \textbf{53.8} & 169 \\
\midrule
\multicolumn{7}{c}{\textbf{Granite 4.0 Micro (3B)}} \\
\midrule
Baseline & 71.4 & 1043 & 29.13 & 341 & 36.8 & 498 \\
Pause Tokens & 67.2 & 146 & 23.84 & 223 & 35.2 & 164 \\
Stepwise Internalization & 69.8 & 137 & 27.95 & 198 & 35.6 & 148 \\
SFT (no CoT) & 71.0 & 398 & 28.42 & 317 & 37.4 & 318 \\
SFT (CoT) & 72.2 & 1412 & 29.06 & 442 & 40.8 & 571 \\
SFT + RL & \underline{73.8} & 1587 & \underline{31.97} & 418 & \textbf{42.8} & 704 \\
\midrule
Abstract-CoT (RL-only) & 70.0 & 127 & 28.26 & 251 & 33.6 & 138 \\
Abstract-CoT (Warm-up) & 71.8 & 181 & 28.69 & 234 & 36.2 & 193 \\
Abstract-CoT (Warm-up + RL) & \textbf{74.4} & 153 & \textbf{33.54} & 219 & \underline{42.6} & 172 \\
\bottomrule
\end{tabular}
\caption{Main results on MATH-500 (accuracy), AlpacaEval (win-rate), and HotpotQA (F1). 
We include the average number of generated tokens per prompt during evaluation, combining reasoning and response tokens. This includes verbal-CoT tokens (baselines), abstract vocabulary tokens for Abstract-CoT, and pause-token count for the Pause Tokens baseline. The top results per model for each column are bolded, with the second best underlined. }
\label{tab:main}
\end{table*}

\paragraph{Baselines and Methods.} Our baselines include (1.) direct answer generation from the prompt ("Baseline"); (2.) "Pause Tokens" \citep{goyal2024think}, where we insert $m_{\max}$ \texttt{<pause>} tokens before generation; (3.) "Stepwise Internalization" (ICoT-SI, \cite{deng2024explicitcotimplicitcot}), involving iterative training while incrementally removing CoT steps ($\max\limits_{i \in \mathcal{C}}|\mathcal{S}_i|$ iterations), starting from SFT w/ CoT; (4.) "SFT (no CoT)", performing supervised fine-tuning on $(x,y)$ pairs; (5.)  "SFT (CoT)", performing supervised fine-tuning on $(x,c, y)$ trajectories; and (6.) "SFT + RL", which involves SFT with CoT followed by RL (GRPO) for verbal CoT.

The variants of our method analyzed are:
\textbf{RL-only} (cold-start),
\textbf{Warm-up} (warm-up only), and
\textbf{Warm-up + RL} (warm-started GRPO). We use an abstract codebook of $M = 64$ tokens (ablations included in Appendix \ref{appendix:scaling-vocab}), and constrain abstract generation to at most $m_{\text{max}}=128$ tokens. Each phase of warm-up (bottlenecked SFT, self-distillation) consists of 3 epochs of training, and RL is performed for 1M episodes. SFT training was performed on $8\times$NVIDIA H100 GPUs and RL training was performed on up to $32\times$NVIDIA H100 GPUs. 

\subsection{Abstract-CoT Enables Efficient Reasoning}

\begin{figure}[t]
\centering
\begin{minipage}[t]{\linewidth}
  \centering
  \includegraphics[width=0.7\linewidth]{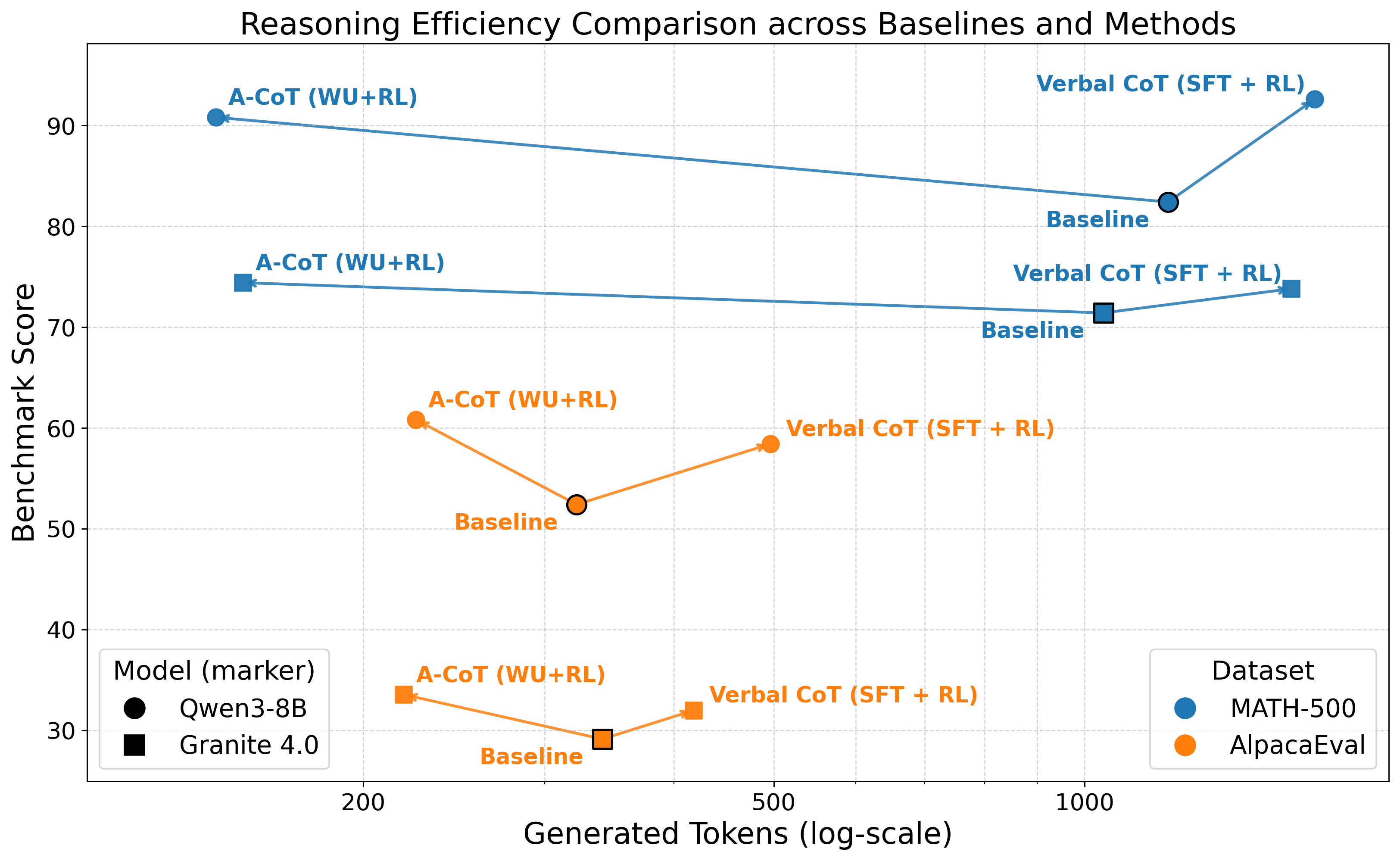}
\end{minipage}\hfill
\begin{minipage}[t]{0\linewidth}
  \centering
\end{minipage}
\caption{Plot of average generated (reasoning + response) tokens (log-scale) vs. benchmark scores (MATH-500 accuracy and AlpacaEval-LC-2.0 win-rate). We compare verbal CoT (SFT + RL) and Abstract-CoT (A-CoT) with policy iteration warm-up and warm-started RL (WU + RL). Compared to the baseline (initial model), Abstract-CoT drastically reduces the number of generated tokens while achieving similar or better performance on both benchmarks.}
\label{fig:token-scaling-and-dist}
\end{figure}

Our findings show that Abstract-CoT nears or exceeds the performance of post-training with verbalized CoT (SFT + RL)
across all models and all benchmarks, as illustrated in Table \ref{tab:main}. We find that all models achieve meaningful gains on AlpacaEval (+2.4 pts for Qwen3-8B, +1.6 pts for Qwen3-4B, and +1.6 pts for Granite 4.0 Micro). We measure token efficiency as the ratio of verbal CoT tokens to abstract tokens, averaged over evaluation prompts. For $c_{\text{verbal}}$ (the model-generated verbal rationale from a verbal CoT baseline) and $m$, we denote the compression ratio $=  \frac{\mathbb{E}[|c_{\mathrm{verbal}}|]}{\mathbb{E}[m]}$. We achieve substantial gains in token efficiency: \textbf{10.4-11.6$\times$} (MATH-500), \textbf{1.9-2.2$\times$} (AlpacaEval), and \textbf{4.0-4.3$\times$} (HotpotQA). In Figure \ref{fig:token-scaling-and-dist}, we plot the token usage and performance on MATH-500 and AlpacaEval, which highlights the efficacy of our method in token-efficient reasoning. Our evaluations on AIME'25 and GPQA-Diamond, as shown in Table \ref{tab:gpqa-aime}, reveal similar trends: Abstract-CoT achieves substantial improvements in token efficiency (2.7$\times$ and 7.9$\times$, respectively), while nearly matching the performance of SFT + RL. These findings suggest that even for more challenging reasoning settings, training models to generate Abstract-CoTs can learn new latent pathways for thinking while remaining performant.  

\begin{table}[t]
\centering
\small
\begin{tabular}{lcccc}
\toprule
Method & \multicolumn{2}{c}{GPQA-Diamond} & \multicolumn{2}{c}{AIME'25} \\
\cmidrule(lr){2-3} \cmidrule(lr){4-5}
 & Accuracy & Tokens & Accuracy & Tokens \\
\midrule
Baseline & 44.9 & 767 & 23.3 & 3981 \\
SFT (no CoT) & 39.4 & 499 & 18.9 & 2745 \\
SFT (CoT) & 45.5 & 1106 & 23.3 & 4627 \\
SFT + RL & \textbf{51.5} & 1382 & \textbf{25.6} & 9343 \\
\midrule
Abstract-CoT (RL-only) & 41.9 & 142 & 18.9 & 2849 \\
Abstract-CoT (Warm-up) & 44.4 & 168 & 20.0 & 2195 \\
Abstract-CoT (Warm-up + RL) & \underline{50.5} & 174 & \underline{24.4} & 3438 \\
\bottomrule
\end{tabular}
\caption{Results (accuracy and token count) on GPQA-Diamond and AIME for Qwen3-8B. The top results per accuracy column are bolded, and the second best is underlined.}
\label{tab:gpqa-aime}
\end{table}

Notably, performing cold-start RL or warm-up alone is insufficient. While cold-start often underperforms the base instruction-tuned model, warm-up manages to outperform the SFT without CoT baseline, yet lags behind SFT with verbal CoT (except for Granite 4.0 on HotpotQA). However, in tandem, these methods succeed, suggesting that the warm-up stage is indeed effective as a warm-start for RL. The inefficacy of cold-start RL indicates that a more substantial burn-in (data volume and training compute) is required to learn a mapping through the abstract sequence, which the warm-up stage addresses.  The observed trends persist across abstract vocabulary sizes, with ablations from $M = \{2^i\}_{i=0}^9$ included in Appendix \ref{appendix:scaling-vocab}. We also find that pause token fine-tuning, despite reducing the number of generated tokens to similar levels as Abstract-CoT, performs worse than the baseline in all settings, corroborating findings in prior works \citep{goyal2024think, hao2025training} that pause fine-tuning without pause pre-training as an initialization proves ineffective. While Stepwise Internalization is relatively more efficient than the SFT baselines at inference, it is more compute-intensive (22 iterations) and still lags our method in performance.

\subsection{Analysis of the Abstract Reasoning Language}\label{sec4.3:abstract-lang-analysis}

\begin{figure}[t]
\centering
\begin{minipage}[t]{0.5\linewidth}
  \centering
  \includegraphics[width=\linewidth]{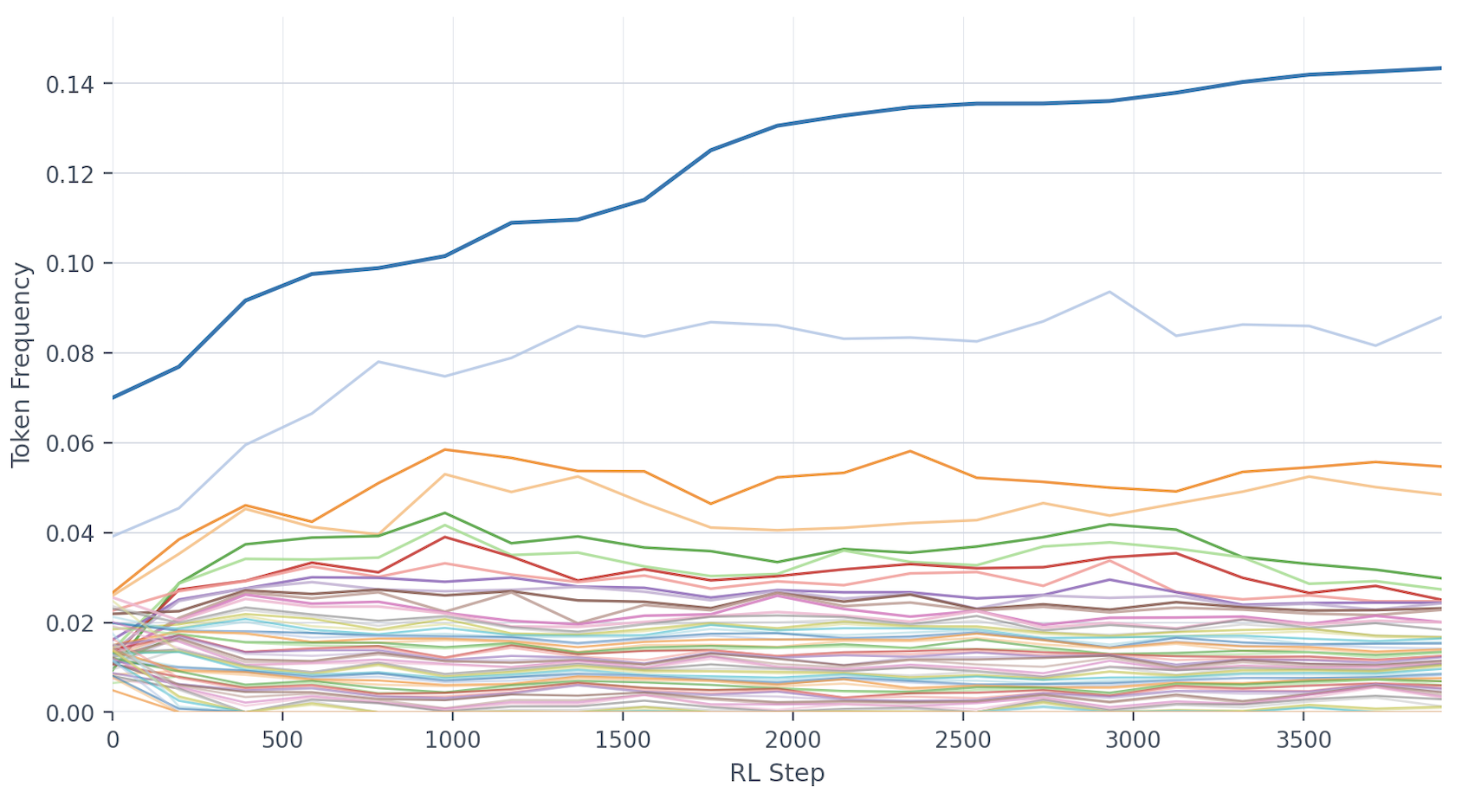}
\end{minipage}\hfill
\begin{minipage}[t]{0.5\linewidth}
  \centering
  \includegraphics[width=\linewidth]{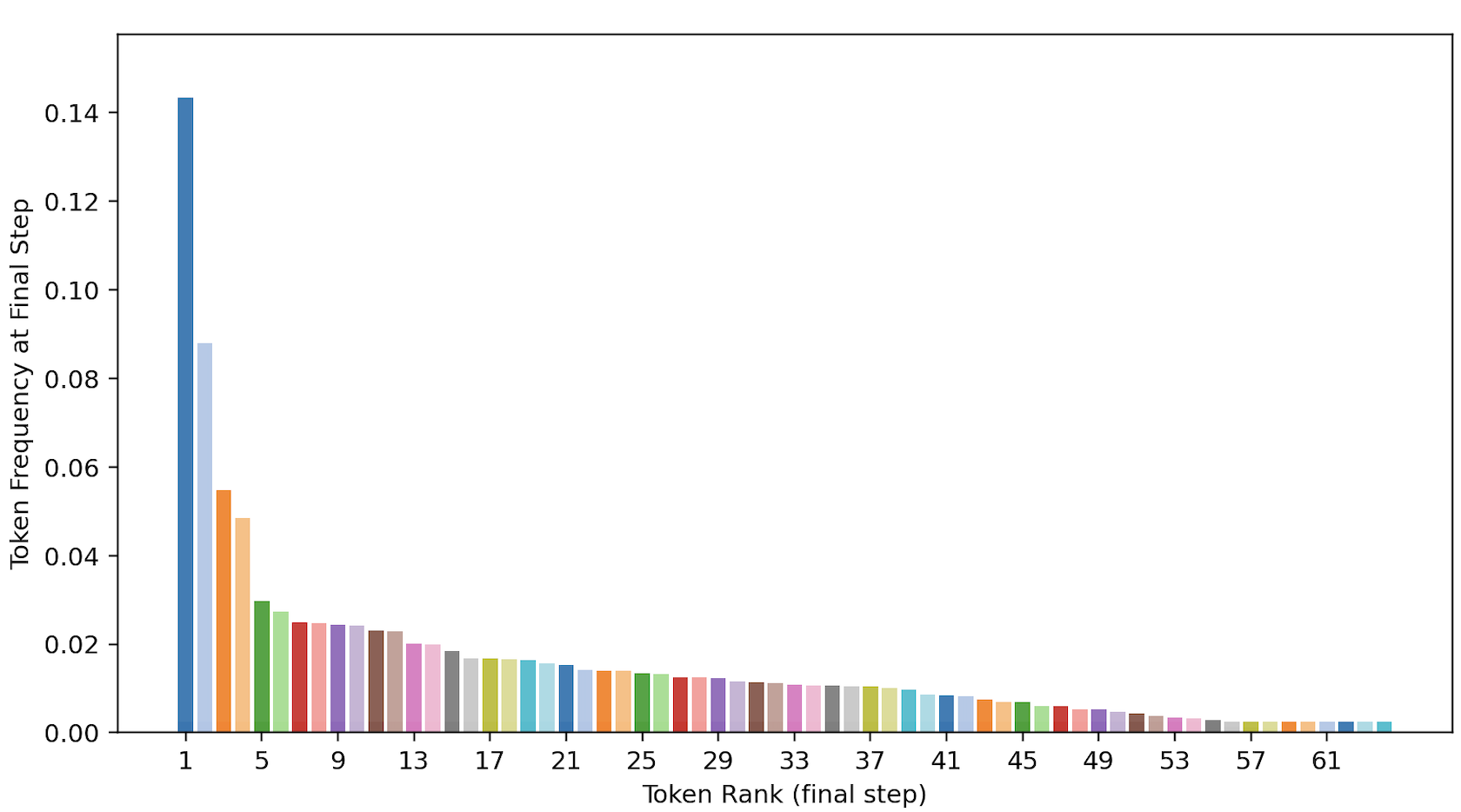}
\end{minipage}
\caption{(Left) Evolution of the abstract token distribution over 1M episodes of warm-started RL. We observe a clear divergence of the top token from the rest as the distribution of token usage in abstract sequences is shaped. (Right) Distribution of final-step token frequencies, which demonstrates a power law characterization akin to Zipf's law, induced purely through post-training, indicating re-use over a learned "reasoning language".
}
\label{fig:token-dist-evol}
\end{figure}

\paragraph{Token Frequency Distribution.} We analyze the distribution of abstract token frequencies during RL training (in Figure \ref{fig:token-dist-evol}), and observe an interesting phenomenon: RL shapes the distribution of token usage in abstract sequences, resulting in a power law-like character. While we initialize the token distribution for the warm-up as uniformly random, subsequent generation and training with on-policy sequences augments this distribution, resulting in an initial shape where a few tokens are invoked more than the rest, with some barely used. Then, we clearly see one token (specifically, \atok{F}) begin to be used much more than the rest, suggesting that it is leveraged across diverse settings, while some tokens with low frequency after warm-up start to appear more often. This serves as an indication of the value in the embedding learning stage, promoting token usage across the vocabulary.

\begin{table}[h]
\centering
\small

\begin{subtable}[t]{0.48\linewidth}
\centering
\setlength{\tabcolsep}{5pt}
\renewcommand{\arraystretch}{1.15}
\begin{tabular}{lcc}
\toprule
\multirow{2}{*}{Method} & \multicolumn{2}{c}{MATH-500} \\
\cmidrule(lr){2-3}
& Original & Permuted \\
\midrule
Verbal CoT (SFT)            & 89.8 & 81.8 (-8.0) \\
Verbal CoT (SFT+RL)         & 92.6 & 81.6 (-11.0) \\
Abstract-CoT (PI-3)         & 87.4 & 83.2 (-4.2) \\
Abstract-CoT (PI-3+RL)      & 90.6 & 82.8 (-7.8) \\
\bottomrule
\end{tabular}
\caption{
Permutation ablation with Qwen3-8B. For verbal CoT, we use turn-level permutation; for Abstract-CoT, we use token-level permutation.
}
\label{tab:perm_math_only}
\end{subtable}
\hfill
\begin{subtable}[t]{0.48\linewidth}
\centering
\setlength{\tabcolsep}{5pt}
\renewcommand{\arraystretch}{1.15}
\begin{tabular}{lcc}
\toprule
\multirow{2}{*}{Method} & \multicolumn{2}{c}{MATH-500} \\
\cmidrule(lr){2-3}
& Full CoT & 32 tokens \\
\midrule
Verbal CoT (SFT)            & 89.8 & 82.8 (-7.0) \\
Verbal CoT (SFT+RL)         & 92.6 & 80.8 (-11.8) \\
Abstract-CoT (PI-3)         & 87.4 & 83.8 (-3.6) \\
Abstract-CoT (PI-3+RL)      & 90.6 & 84.6 (-6.0) \\
\bottomrule
\end{tabular}
\caption{
Truncation ablation with Qwen3-8B, comparing performance with the full CoT to capping the reasoning trace at 32 thinking tokens.
}
\label{tab:trunc_math_only}
\end{subtable}

\caption{
Sensitivity analyses (permutation, truncation) on MATH-500 with Qwen3-8B. PI-3 refers to performing 3 iterations of the policy iteration warm-up loop. 
}
\label{tab:perm_trunc_math}
\end{table}

\paragraph{Permutation Analysis.} To analyze the compositional power of the abstract reasoning language in generating ordered sequences for response generation, we study permuting the CoTs in Table \ref{tab:perm_math_only}. The experimental setup and further results are included in Appendix \ref{appendix:permutation}. We observe a clear decrease in performance with both methods, before and after RL training. 
Given a strong prior for verbalized CoT through pre-training, such a performance drop with permutation is expected. At the same time, observing this behavior with Abstract-CoT offers compelling evidence of learned compositional behavior over the new vocabulary.

\paragraph{CoT Truncation Analysis.} We study the performance of verbalized and Abstract-CoT under truncation: halting the CoT after $k$ tokens, appending their respective end delimiters ($\texttt{</think>}$ and $\abend$), and generating a response to evaluate. The results are shown in Table \ref{tab:trunc_math_only} with $k = 32$, which depict clear decreases for both methods, with Verbal CoT seeing a much greater decline as it is trained to produce long rationales in natural language. By contrast, the decline for Abstract-CoT is not as severe since it has been trained to reason through a short, bounded trace, resulting in a more graceful degradation. Further ablations across benchmarks and different values of $k$ are included in Appendix \ref{appendix:halting}.

\section{Discussion}

Abstract-CoT crucially lies between two extremes: human-interpretable verbal CoT (which is expensive during inference) and fully implicit reasoning (which is implicit but challenging to control). The discrete abstract codebook offers a middle ground, yielding short, bounded traces while exposing a structured intermediate segment that can be analyzed.
As such, Abstract-CoT offers a practical path for inference scaling in settings where intermediate reasoning need not be human-readable. 
The power-law findings are promising for future analyses on concept-level understanding, to study the underlying mechanisms of concept learning with abstract tokens and how those representations evolve as new sequences are explored through RL. This would serve to lend mechanistic support to the compositionality hypothesis over the abstract reasoning language, along with greater interpretability. It is possible that different concept clusters may have varying sample complexities for effective learning, related to the vocabulary size and sequence length, and shaped through the training phases. 
Therefore, the abstract tokens also provide a new intermediate interface for chain-of-thought monitorability \citep{korbak2025chainthoughtmonitorabilitynew} and auditability studies.

As small budgets may be insufficient for long-horizon reasoning, budget-adaptive mechanisms with Abstract-CoT (with difficulty-aware abstract sequence lengths) offers exciting potential for more challenging regimes. It is possible that a hierarchical structure over the codebook could enable re-usable subroutines for particular tasks, offering another mechanism to shape larger abstract vocabularies. 

\section{Conclusion}

We introduce Abstract Chain-of-Thought,  a discrete latent reasoning mechanism that post-trains language models to generate a short sequence of new ''abstract tokens''. A policy iteration warm-up loop learns abstract-token embeddings and sequence generation, while warm-started RL further improves performance through exploration. Abstract Chain-of-Thought can produce substantially shorter sequences (up to $12\times$) while achieving similar performance as post-training with verbalized CoT,  generalizing across model families. Our findings suggest that filler tokens can be effectively introduced and used in post-training alone, offering a meaningful vocabulary-augmentation strategy to introduce a ''reasoning language'' without requiring continued pre-training.  
Our work presents a new outlet for training LLMs with latent reasoning behavior, providing a fertile ground for future research.

\appendix
\newpage
\appendix

\section*{\Large Appendix}
\startcontents[appendix]
\printcontents[appendix]{}{1}{\setcounter{tocdepth}{3}\vspace{0.5em}}

\newpage

\section{Ablations}

\subsection{Scaling Abstract Vocabulary Size}\label{appendix:scaling-vocab}

\begin{figure}[H]
\centering
\begin{minipage}[t]{\linewidth}
  \centering
  \includegraphics[width=0.85\linewidth]{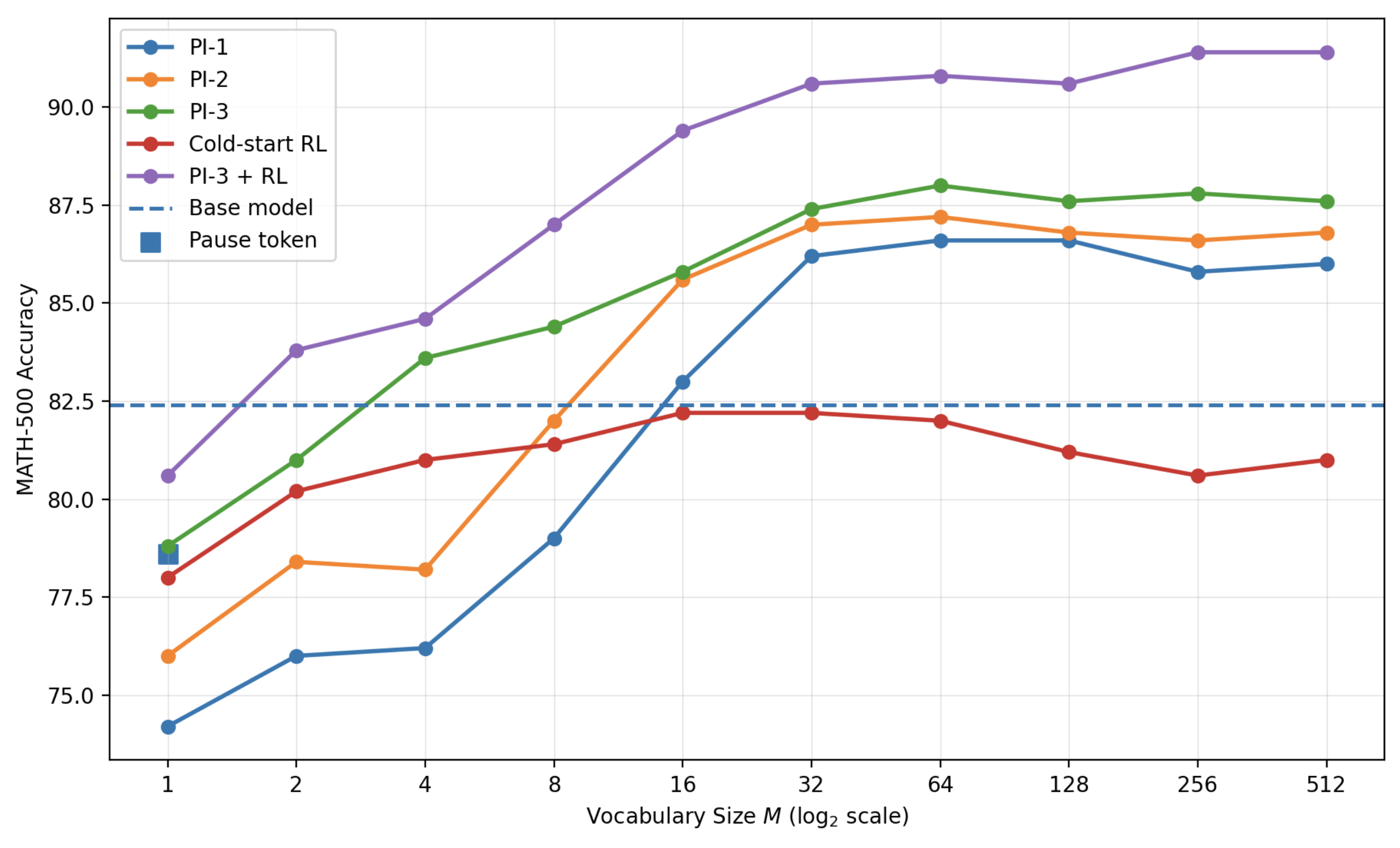}
\end{minipage}\hfill
\begin{minipage}[t]{0\linewidth}
  \centering
\end{minipage}
\caption{MATH-500 abstract vocabulary scaling ablation across stages of the Abstract-CoT pipeline, compared against cold-start RL, the base model performance, and the pause token baseline ($M = 1$ variant with greater data). The results show clear and consistent scaling trends across warm-up stages and warm-started RL, plateauing at larger vocabulary sizes, while cold-start flattens out much earlier and does not surpass the base model performance.}
\label{fig:scaling-vocab-math}
\end{figure}

\begin{figure}[H]
\centering
\begin{minipage}[t]{\linewidth}
  \centering
  \includegraphics[width=0.85\linewidth]{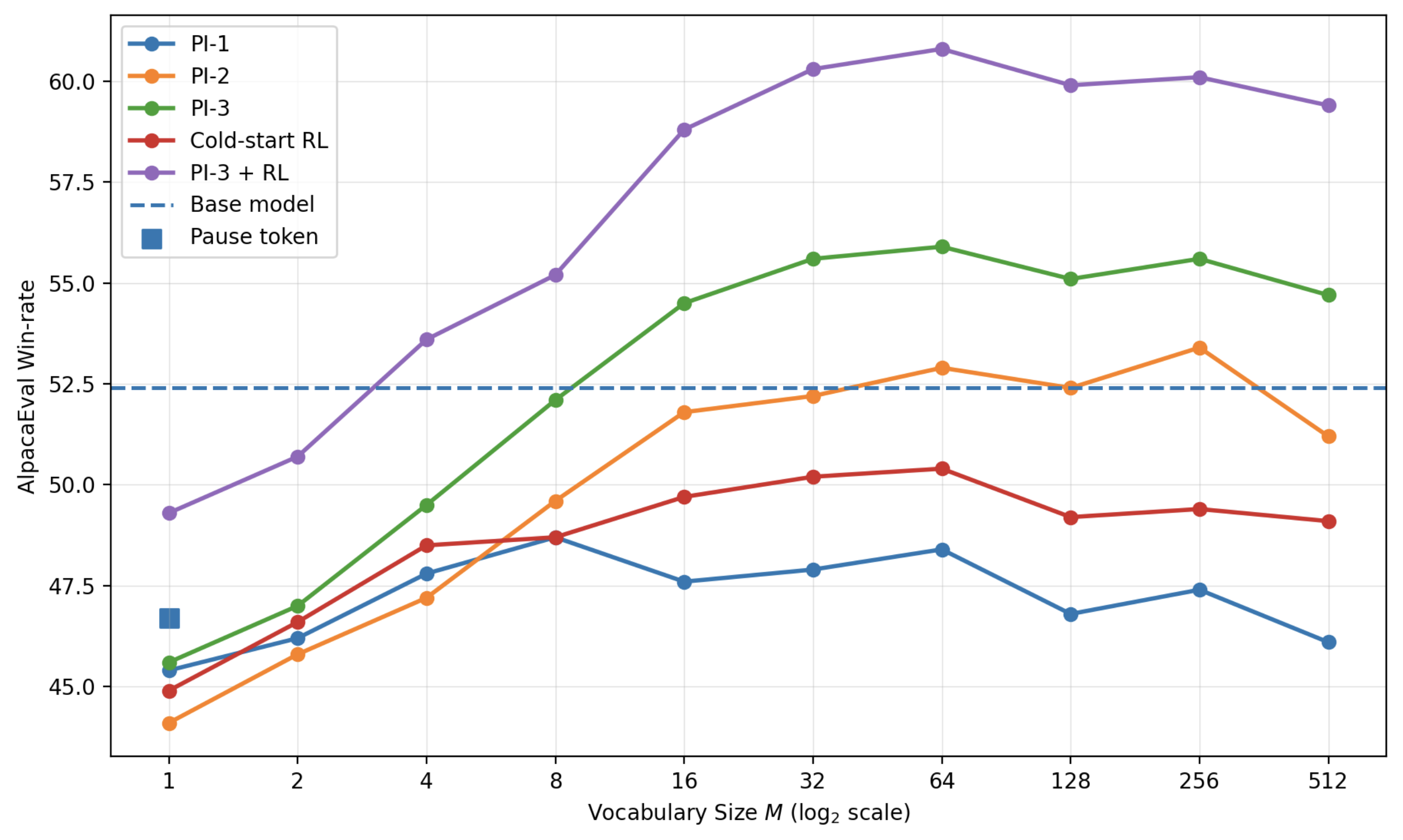}
\end{minipage}\hfill
\begin{minipage}[t]{0\linewidth}
  \centering
\end{minipage}
\caption{AlpacaEval abstract vocabulary scaling ablation. The results show clearer differentiation than MATH-500, with $M = 64$ achieving the highest score with PI-3 and PI-3 + RL, and cold-start RL again underperforming the base model while surpassing PI-1. }
\label{fig:scaling-vocab-alpaca}
\end{figure}

We ablate the behavior of Abstract-CoT when scaling the size of $\mathcal{V}^*$, with values ranging from $M = 1 ~(2^0)$ to $M = 512 ~(2^9)$, across stages of training ($3\times$ Policy Iteration warm-up, GRPO) as well as with cold-start GRPO with randomly initialized embeddings for the abstract tokens. Aside from analyzing the value of a larger vocabulary on benchmark performance across MATH-500, AlpacaEval, and HotpotQA, this allows us to study trends in the frequency distributions between cold-started and warm-started RL. Since the $M = 1$ case trivially invokes the same token every time, we start from $M = 2$. Note that for the distribution plots, the y-axes have been scaled to illustrate the variation in token frequency. 

\vspace{-3mm}

\begin{figure}[H]
\centering
\begin{minipage}[t]{\linewidth}
  \centering
  \includegraphics[width=0.85\linewidth]{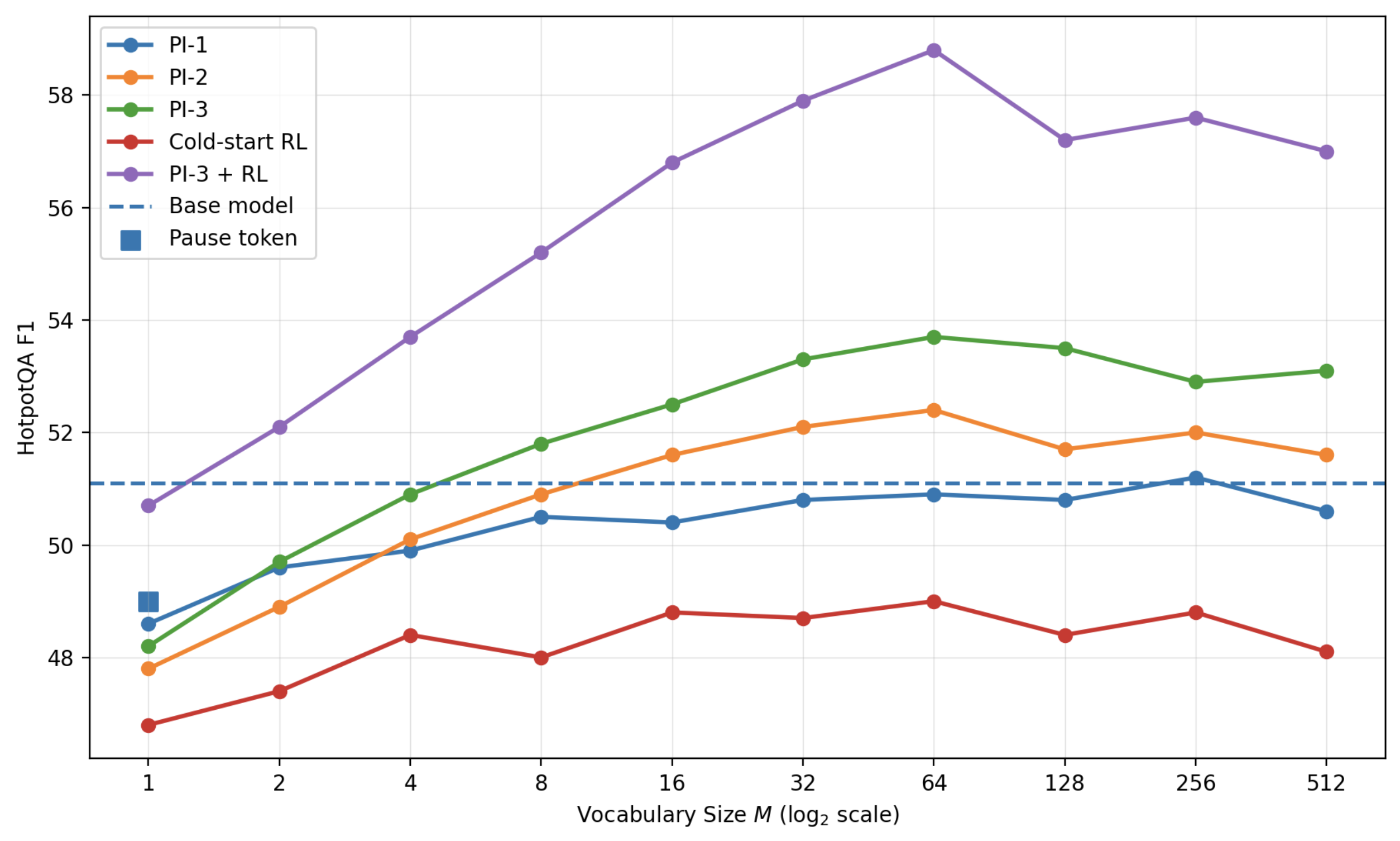}
\end{minipage}\hfill
\begin{minipage}[t]{0\linewidth}
  \centering
\end{minipage}
\caption{HotpotQA abstract vocabulary scaling ablation. The results again show clearly differentiable scaling curves, with PI-3 and PI-3 + RL exhibiting near-linear improvement from $M = 1$ to $M = 64$, followed by a decline. On the other hand, cold-start lags behind even PI-1, further supporting the value of the warm-up stage.  }
\label{fig:scaling-vocab-hotpot}
\end{figure}

The results in Figures \ref{fig:scaling-vocab-math}-
\ref{fig:scaling-vocab-hotpot} demonstrate that across all three benchmarks, scaling the size of the vocabulary improves performance, yet eventually saturates and slightly declines. This trend holds across all methods, though cold-start RL appears to consistently underperform the base model's performance out-of-the-box regardless of vocabulary size -- in fact, it underperforms PI-1 on HotpotQA for all sizes and on MATH-500 for the tested values of $M \geq 16$. On the basis of these scaling analyses, $M = 64$ was chosen as the best configuration.

\subsubsection{Frequency Distribution Across RL Training}

\begin{figure}[h]
\centering
\begin{minipage}[t]{0.5\linewidth}
  \centering
  \includegraphics[width=\linewidth]{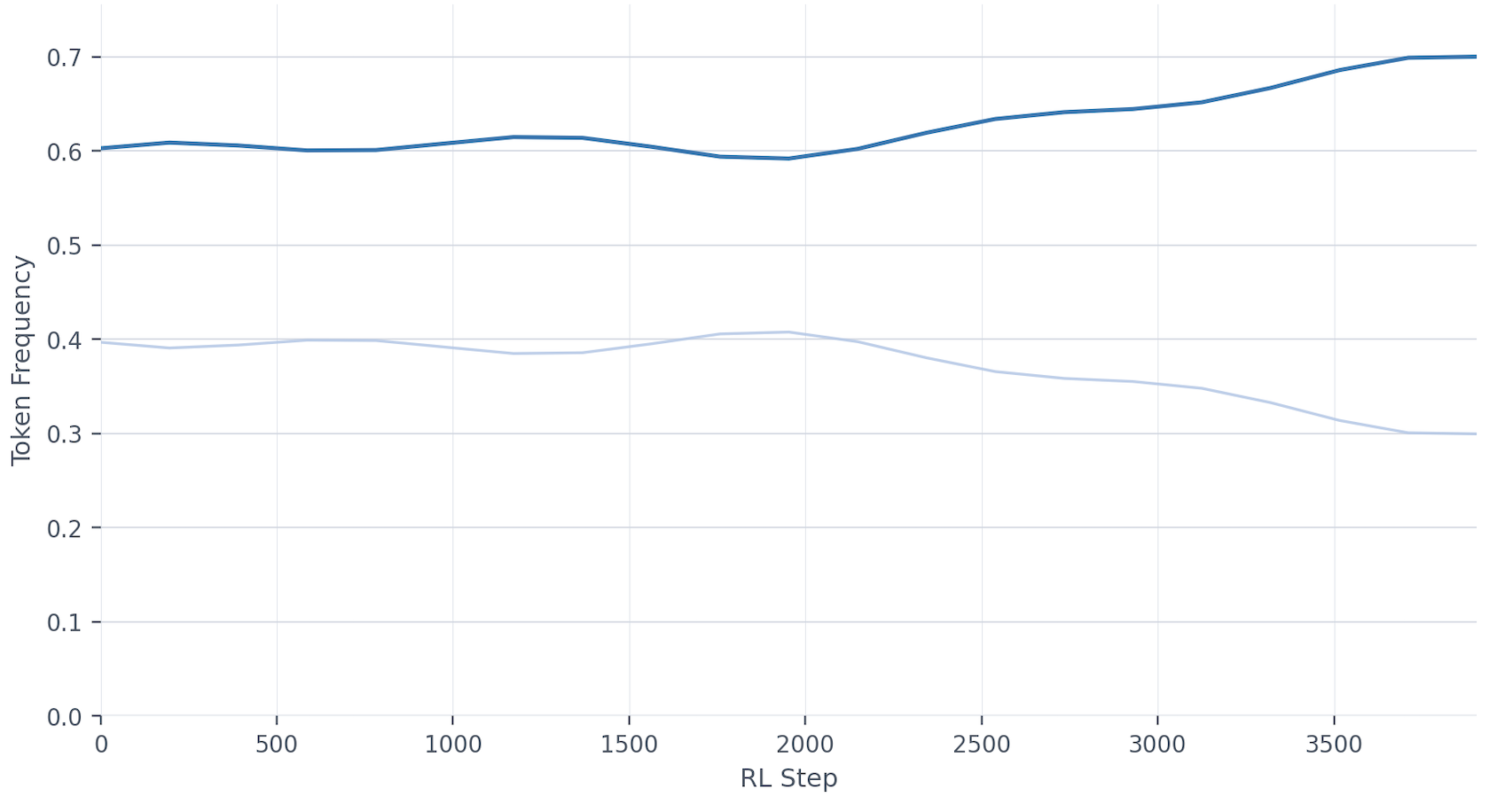}
\end{minipage}\hfill
\begin{minipage}[t]{0.5\linewidth}
  \centering
  \includegraphics[width=\linewidth]{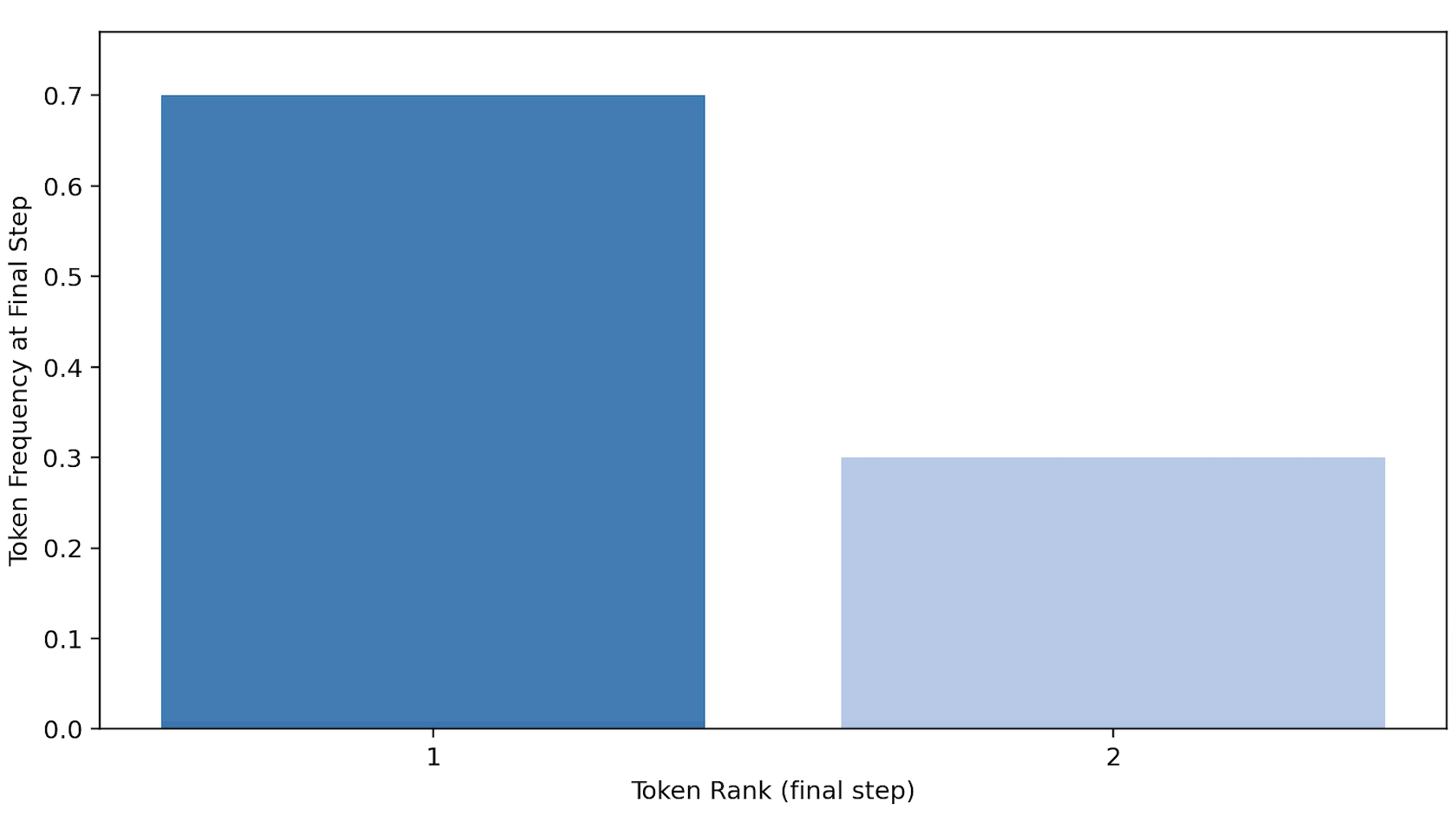}
\end{minipage}
\caption{Scaling ablation with $M = 2$ abstract token vocabulary.}
\label{fig:token-dist-m2}
\end{figure}

\begin{figure}[h]
\centering
\begin{minipage}[t]{0.5\linewidth}
  \centering
  \includegraphics[width=\linewidth]{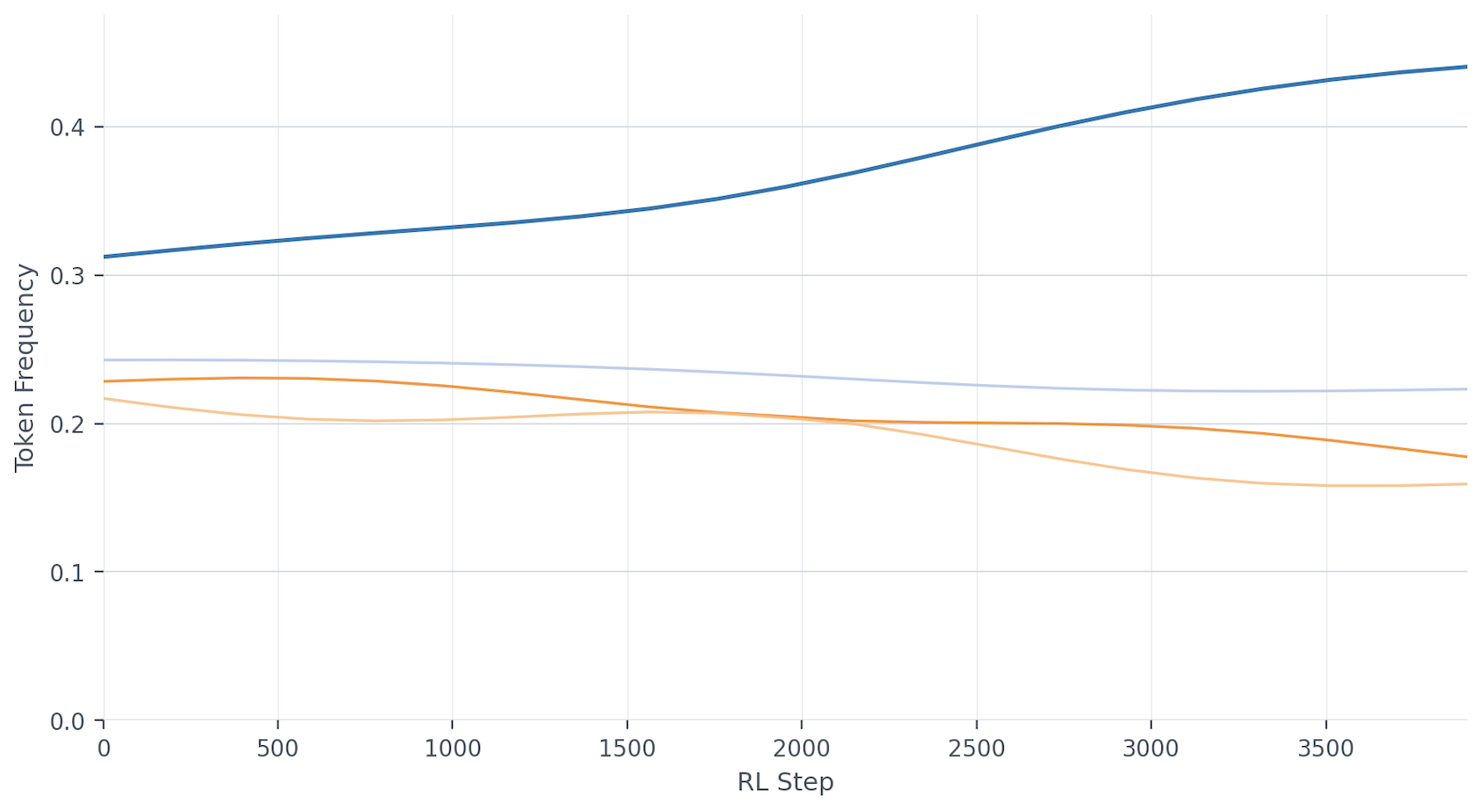}
\end{minipage}\hfill
\begin{minipage}[t]{0.5\linewidth}
  \centering
  \includegraphics[width=\linewidth]{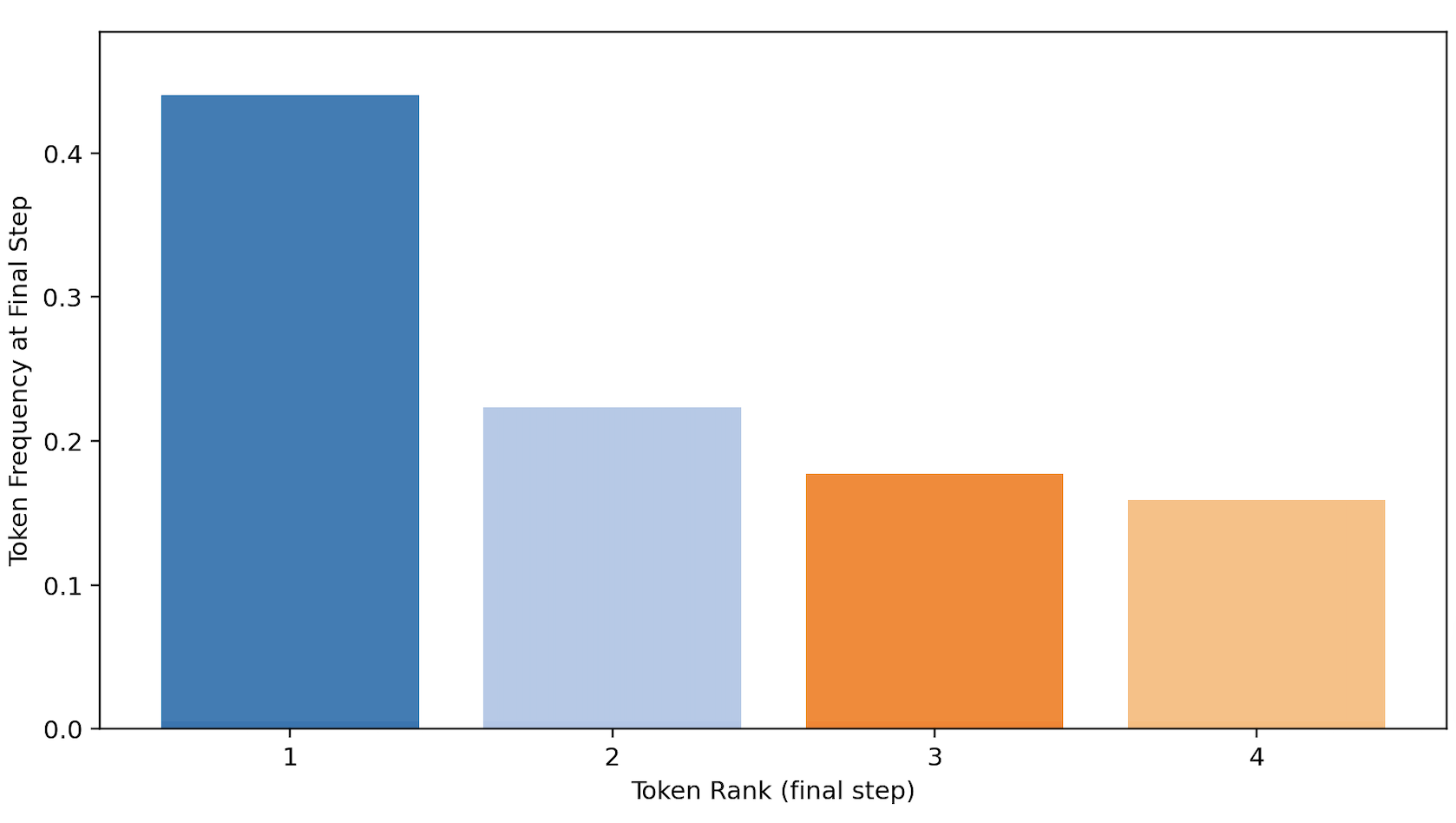}
\end{minipage}
\caption{Scaling ablation with $M = 4$ abstract token vocabulary.}
\label{fig:token-dist-m4}
\end{figure}

\begin{figure}[h]
\centering
\begin{minipage}[t]{0.5\linewidth}
  \centering
  \includegraphics[width=\linewidth]{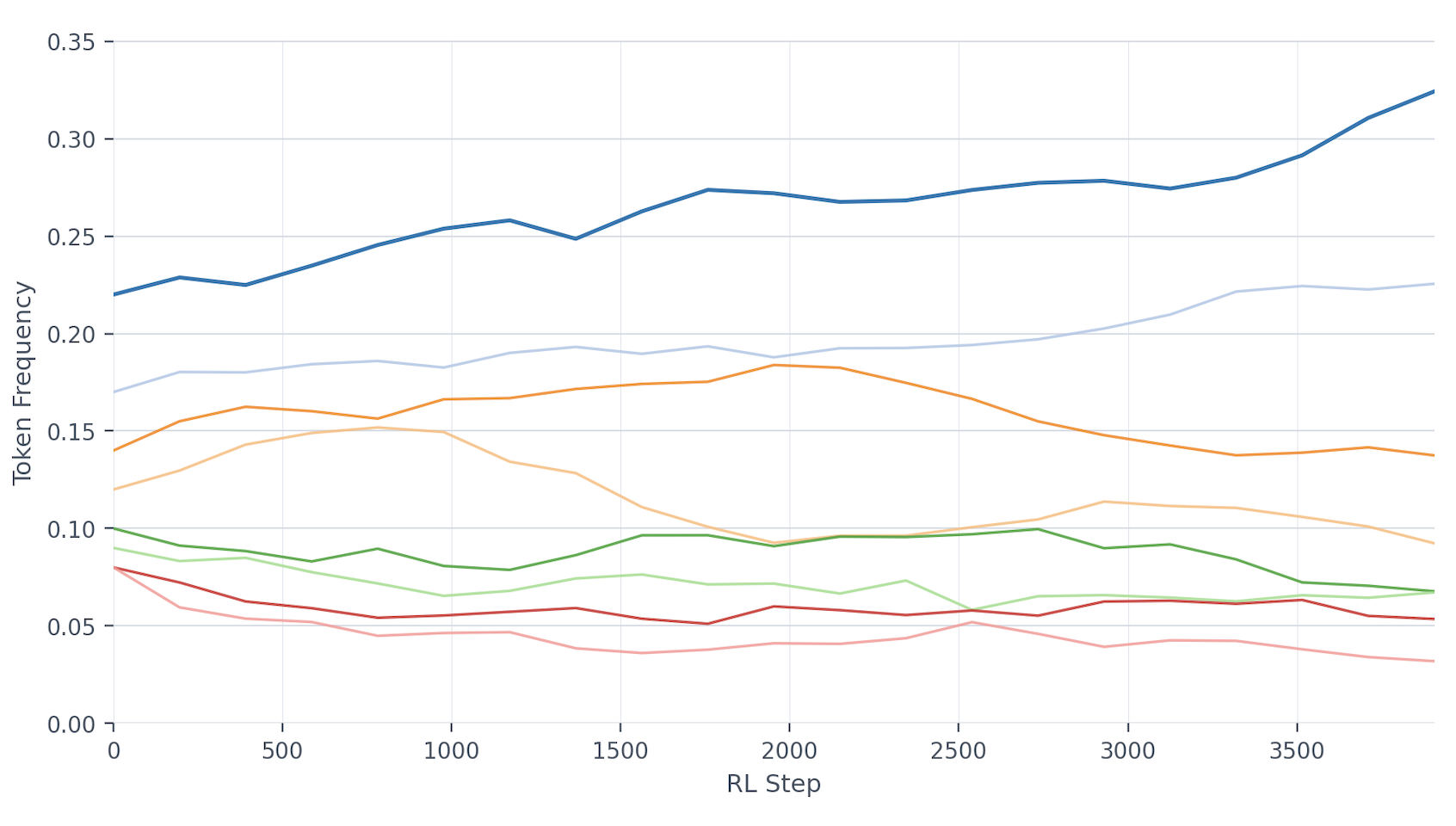}
\end{minipage}\hfill
\begin{minipage}[t]{0.5\linewidth}
  \centering
  \includegraphics[width=\linewidth]{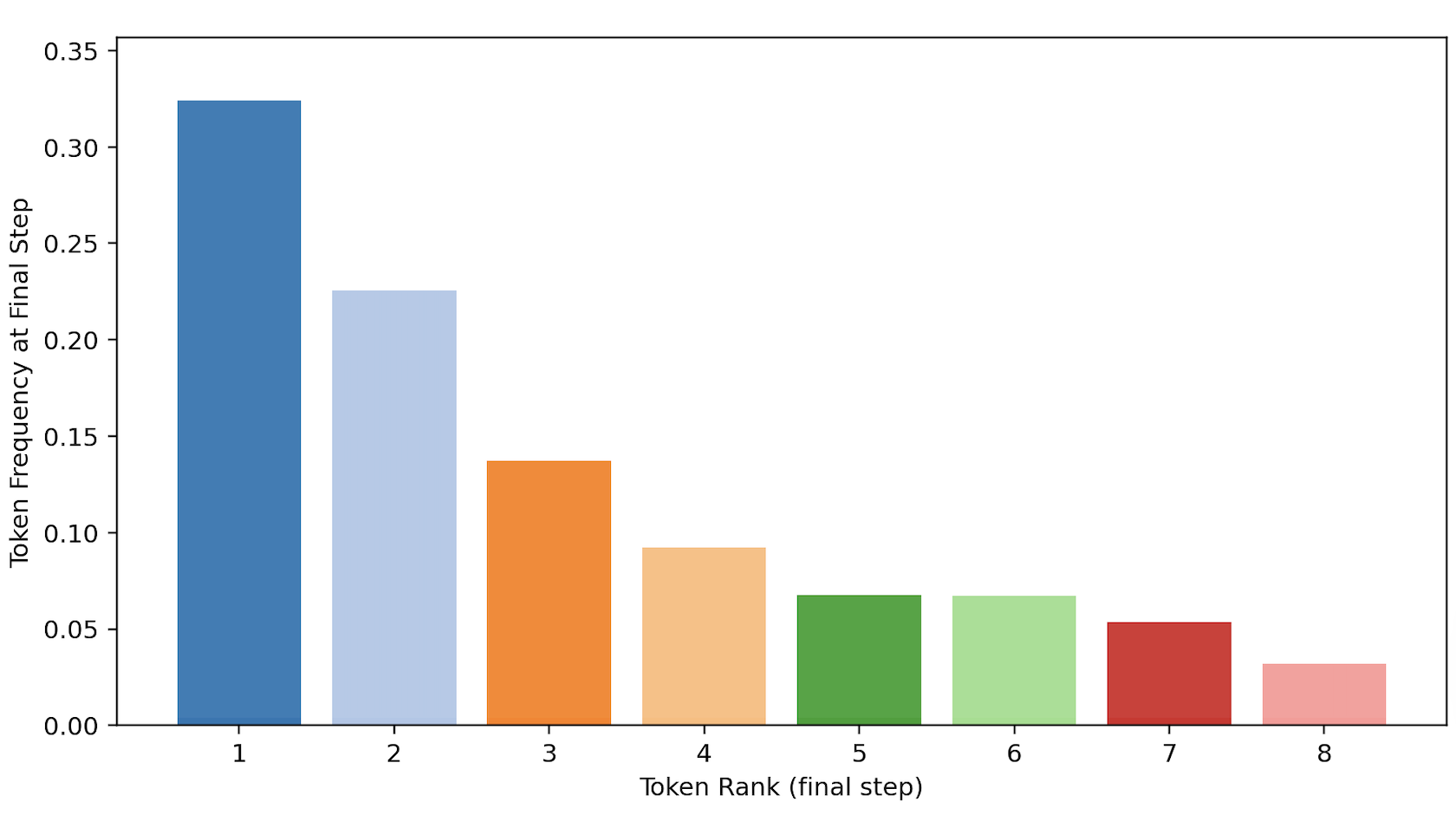}
\end{minipage}
\caption{Scaling ablation with $M = 8$ abstract token vocabulary.}
\label{fig:token-dist-m8}
\end{figure}

\begin{figure}[H]
\centering
\begin{minipage}[t]{0.5\linewidth}
  \centering
  \includegraphics[width=\linewidth]{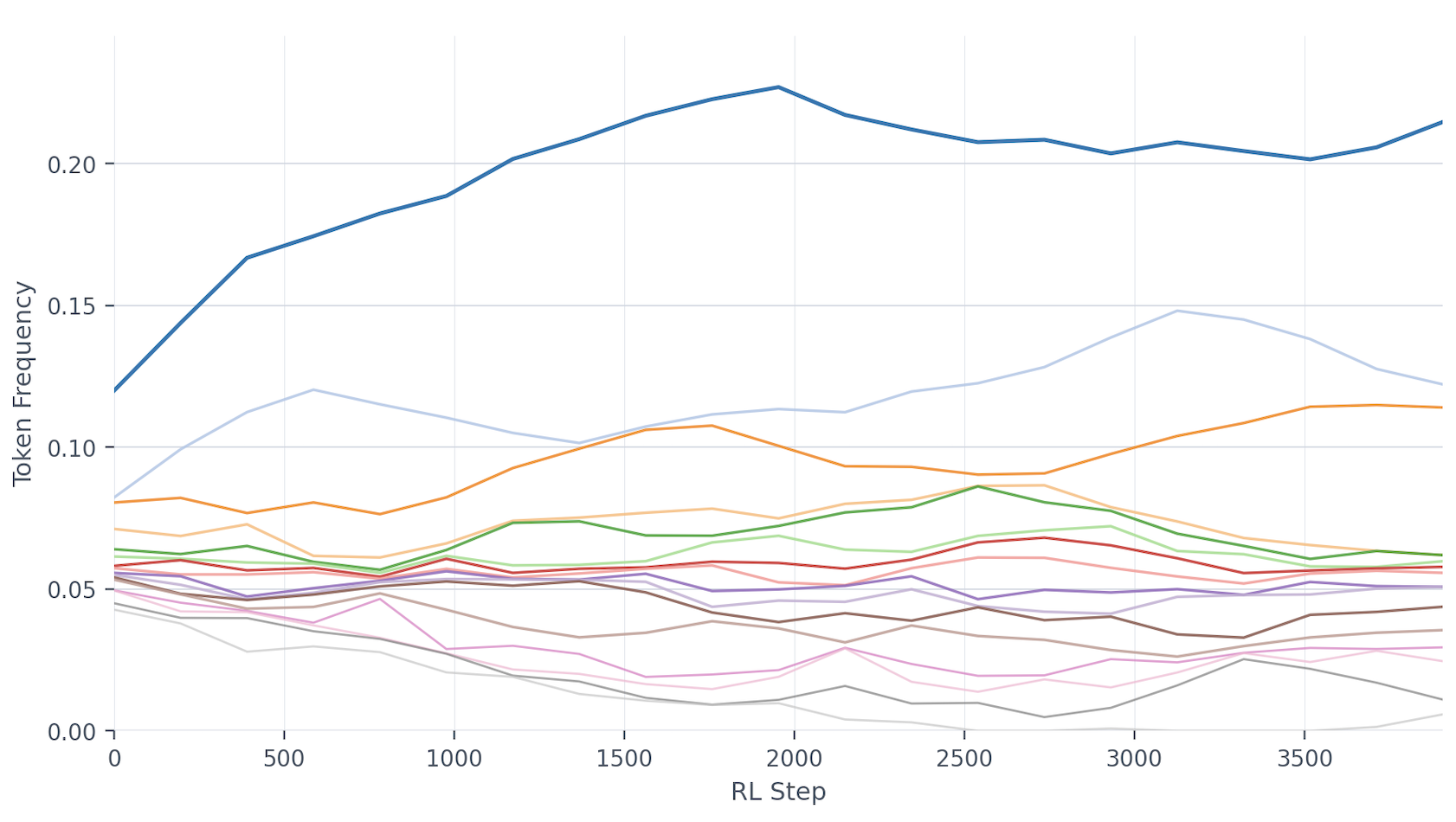}
\end{minipage}\hfill
\begin{minipage}[t]{0.5\linewidth}
  \centering
  \includegraphics[width=\linewidth]{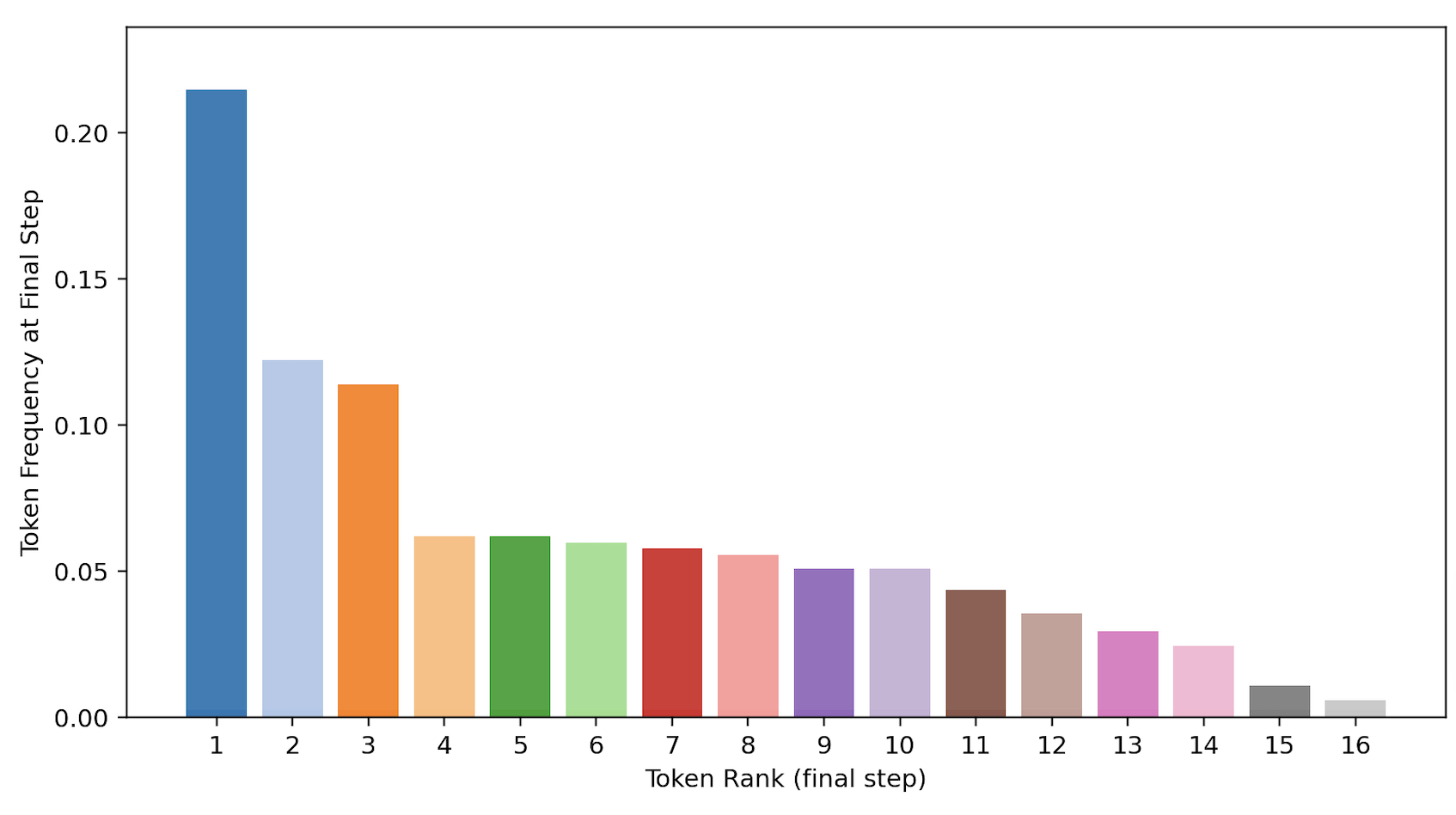}
\end{minipage}
\caption{Scaling ablation with $M = 16$ abstract token vocabulary.}
\label{fig:token-dist-m16}
\end{figure}

\begin{figure}[h]
\centering
\begin{minipage}[t]{0.5\linewidth}
  \centering
  \includegraphics[width=\linewidth]{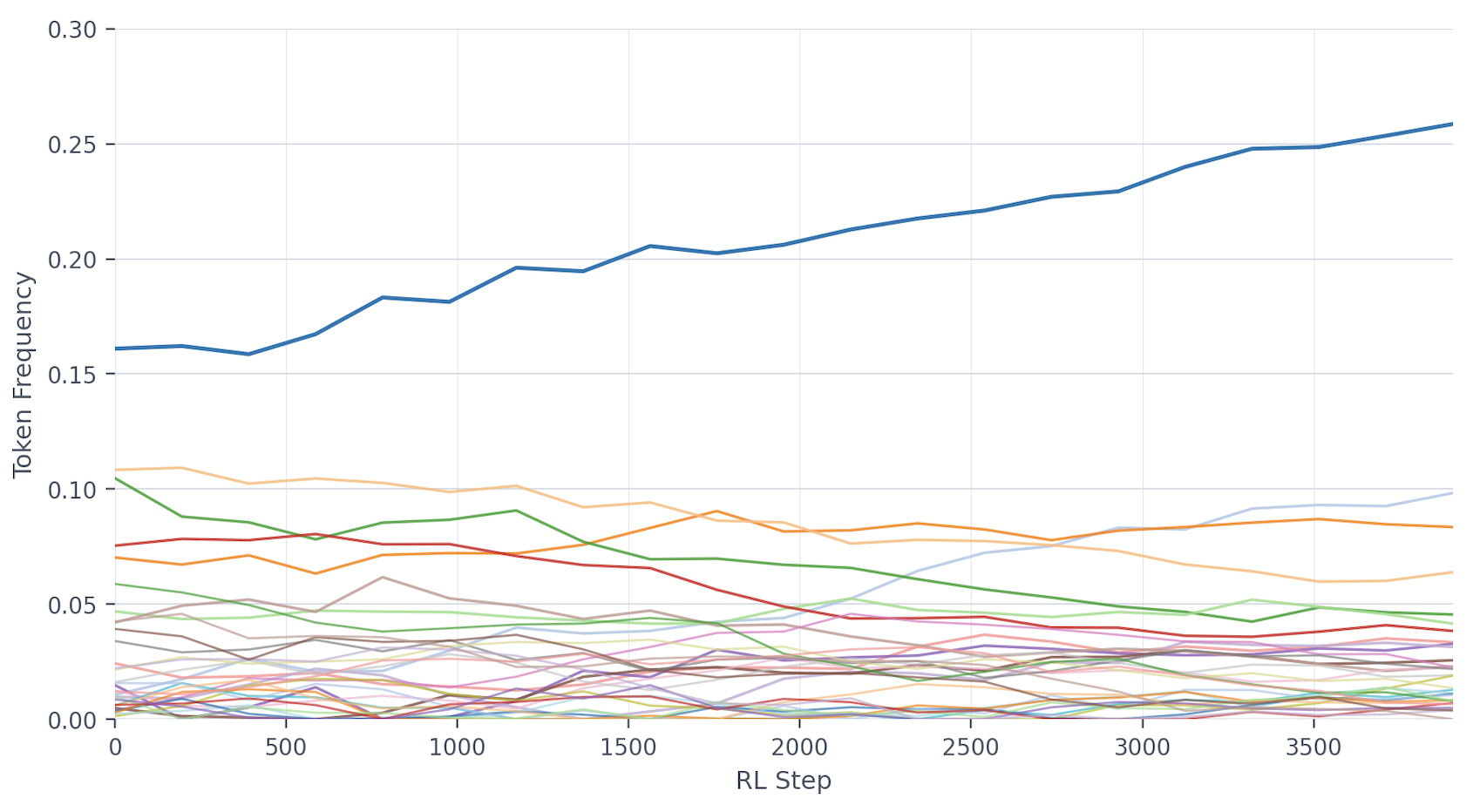}
\end{minipage}\hfill
\begin{minipage}[t]{0.5\linewidth}
  \centering
  \includegraphics[width=\linewidth]{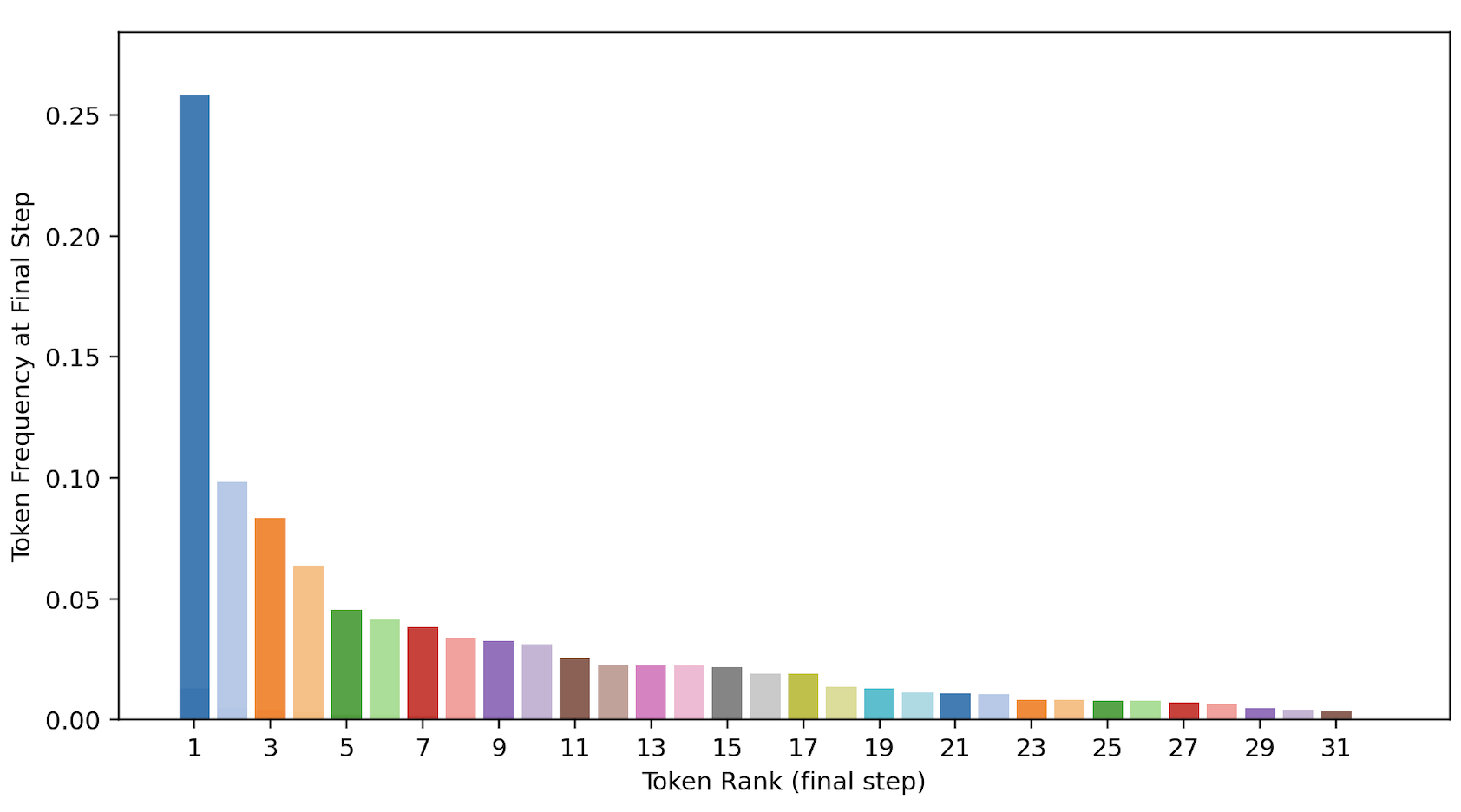}
\end{minipage}
\caption{Scaling ablation with $M = 32$ abstract token vocabulary.}
\label{fig:token-dist-m32}
\end{figure}

\begin{figure}[h]
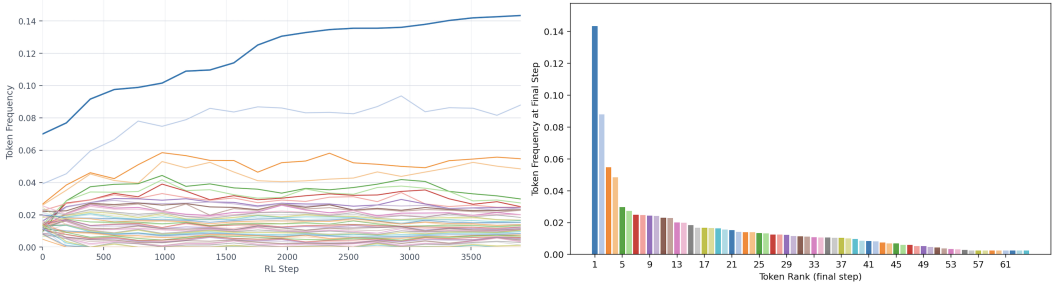

\centering
\begin{minipage}[t]{0.5\linewidth}
  \centering
  \includegraphics[width=\linewidth]{figs/tok-dist-figs/token-dist-evolution_64.png}
\end{minipage}\hfill
\begin{minipage}[t]{0.5\linewidth}
  \centering
  \includegraphics[width=\linewidth]{figs/tok-dist-figs/token-ranks_64.png}
\end{minipage}
\caption{Scaling ablation with $M = 64$ abstract token vocabulary. Note that this is the same as Figure \ref{fig:token-dist-evol}, included in the appendix for comparison to the other vocabulary sizes.}
\label{fig:token-dist-m64}
\end{figure}

\begin{figure}[H]
\centering
\begin{minipage}[t]{0.5\linewidth}
  \centering
  \includegraphics[width=\linewidth]{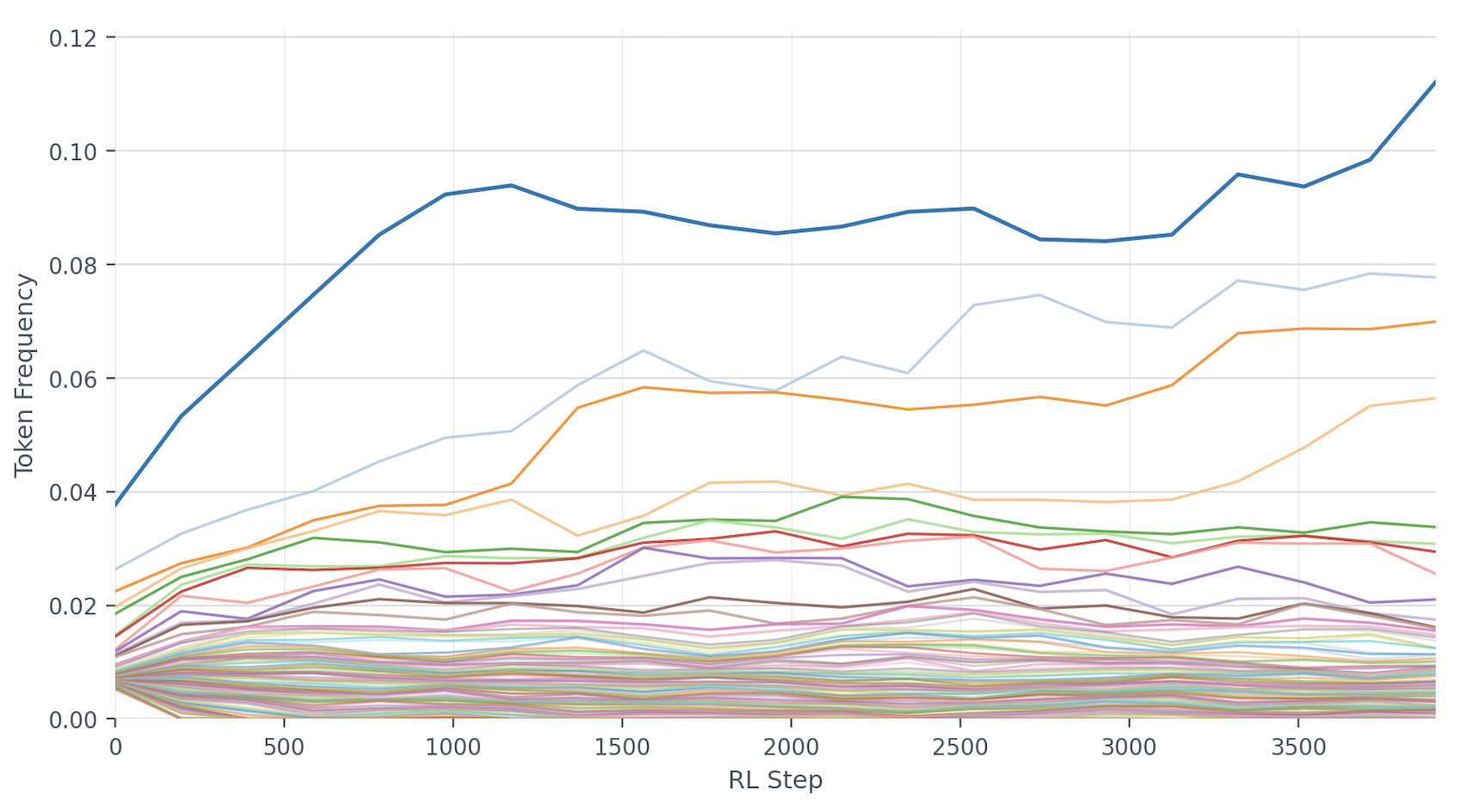}
\end{minipage}\hfill
\begin{minipage}[t]{0.5\linewidth}
  \centering
  \includegraphics[width=\linewidth]{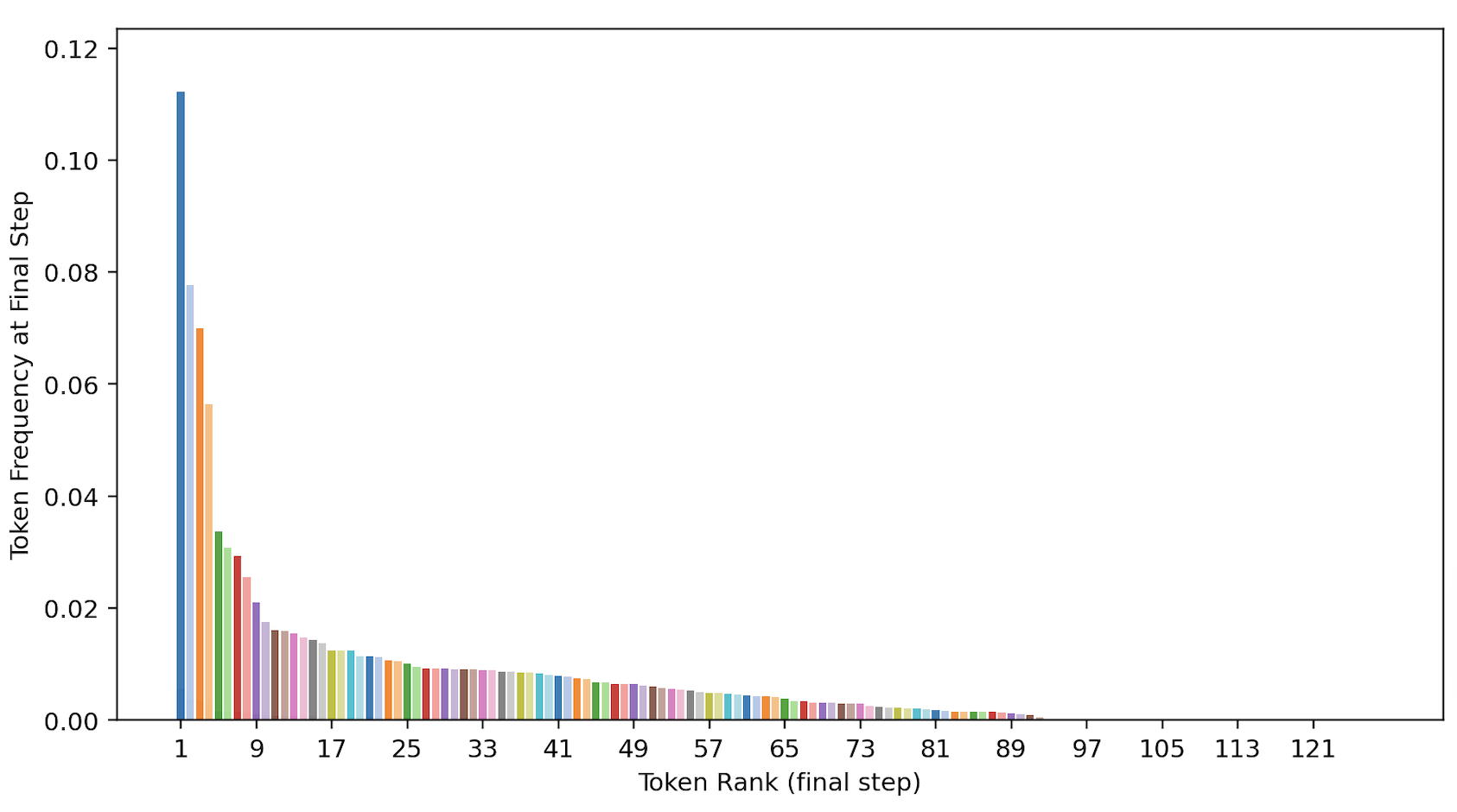}
\end{minipage}
\caption{Scaling ablation with $M = 128$ abstract token vocabulary.}
\label{fig:token-dist-m128}
\end{figure}

\begin{figure}[H]
\centering
\begin{minipage}[t]{0.5\linewidth}
  \centering
  \includegraphics[width=\linewidth]{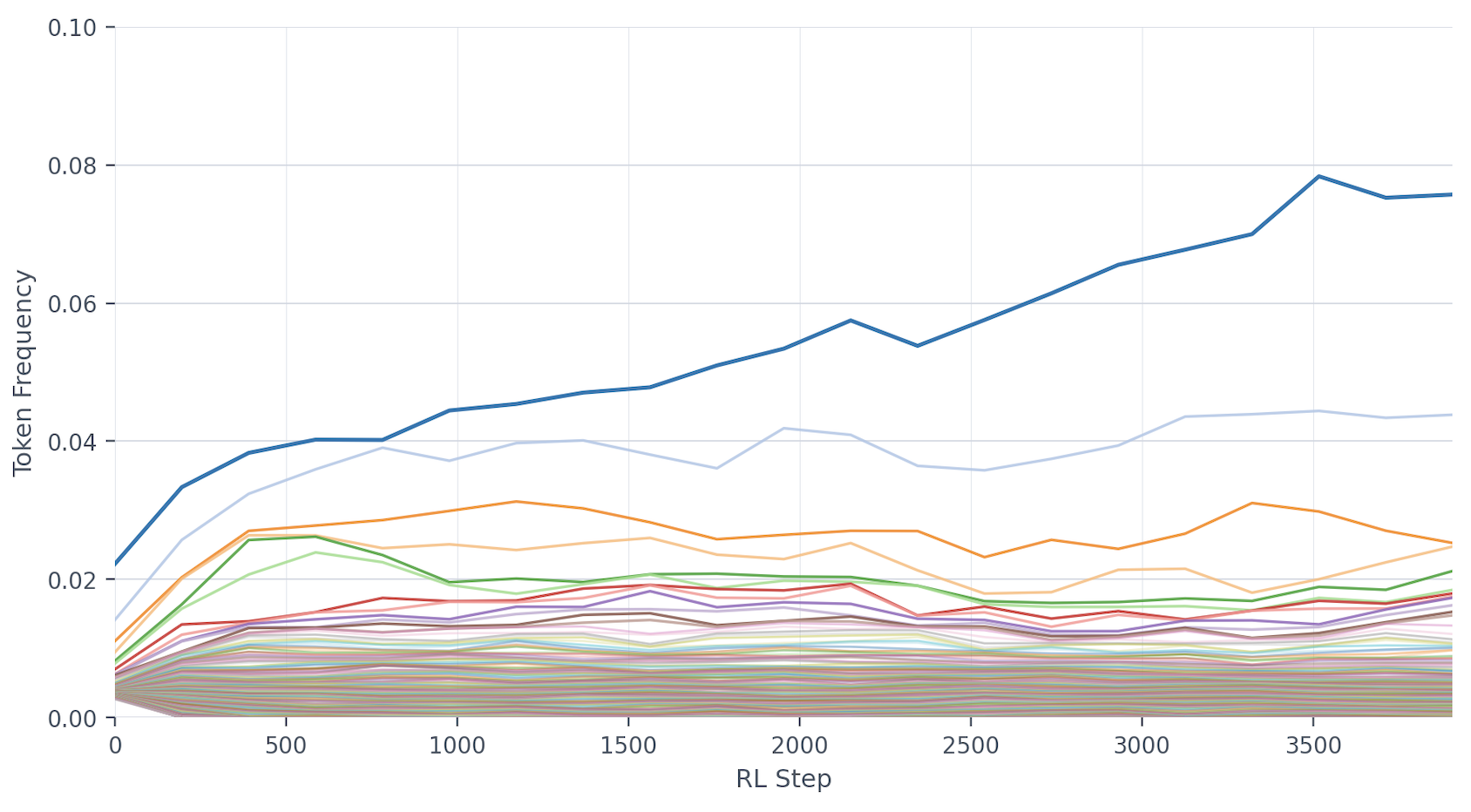}
\end{minipage}\hfill
\begin{minipage}[t]{0.5\linewidth}
  \centering
  \includegraphics[width=\linewidth]{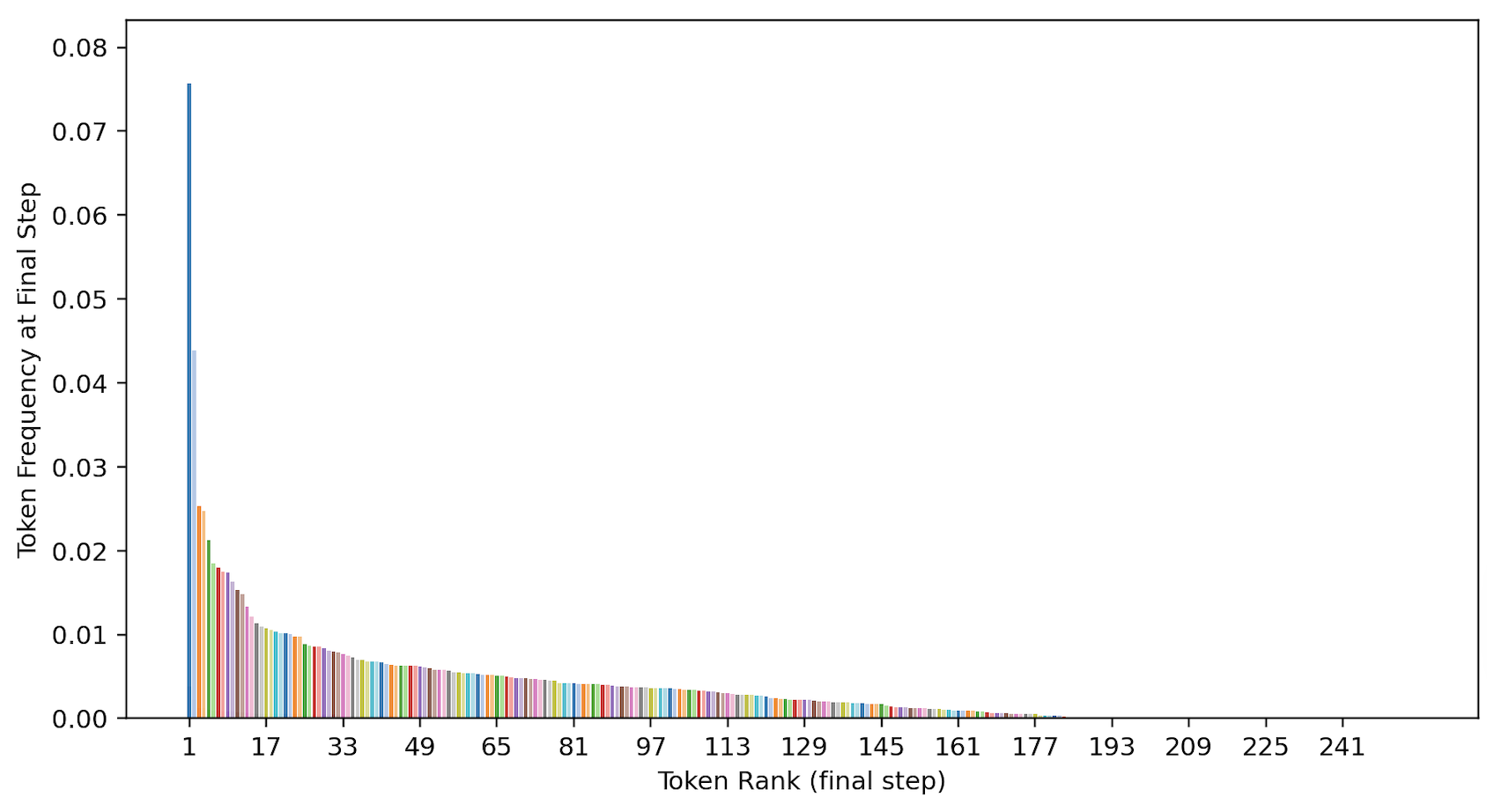}
\end{minipage}
\caption{Scaling ablation with $M = 256$ abstract token vocabulary.}
\label{fig:token-dist-m256}
\end{figure}

\begin{figure}[H]
\centering
\begin{minipage}[t]{0.5\linewidth}
  \centering
  \includegraphics[width=\linewidth]{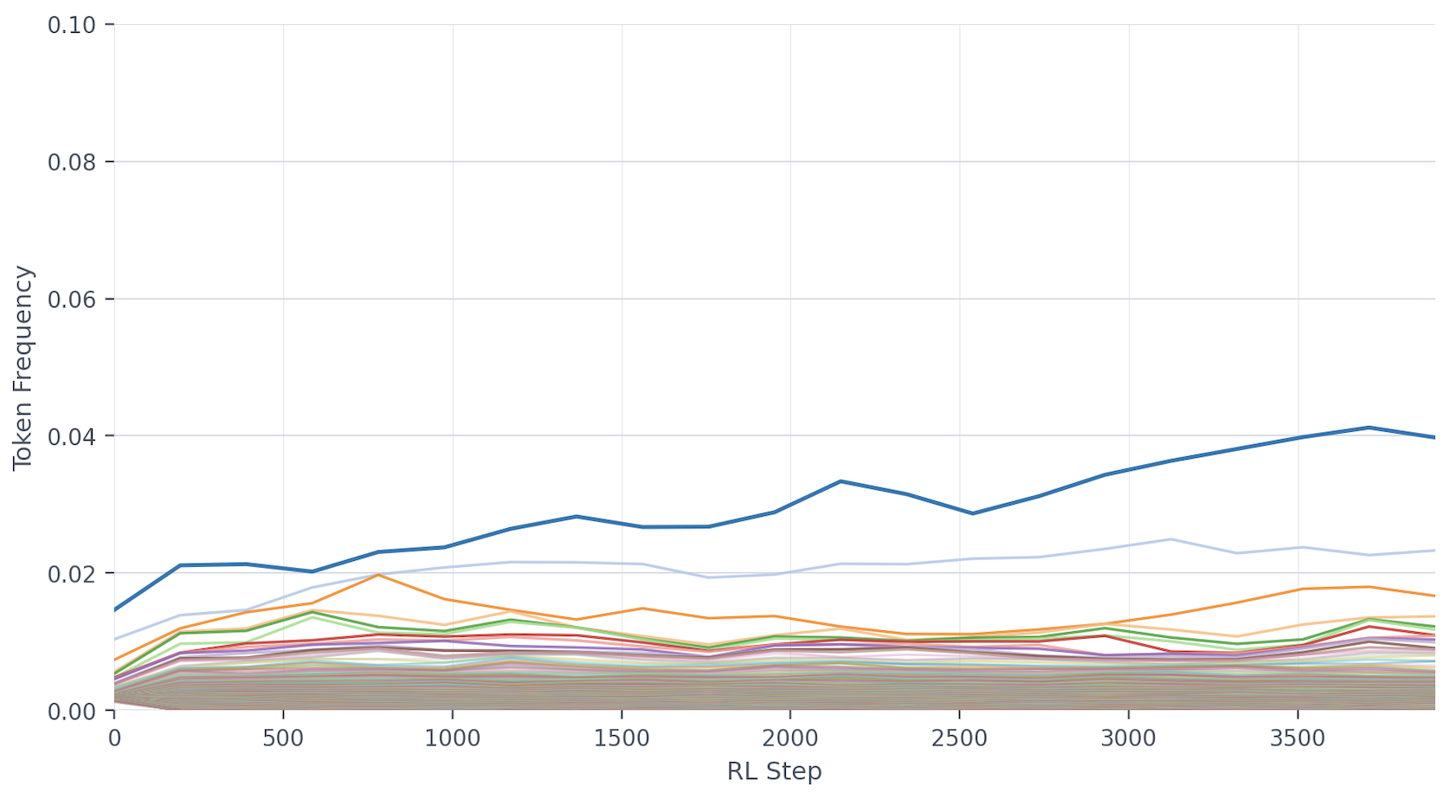}
\end{minipage}\hfill
\begin{minipage}[t]{0.5\linewidth}
  \centering
  \includegraphics[width=\linewidth]{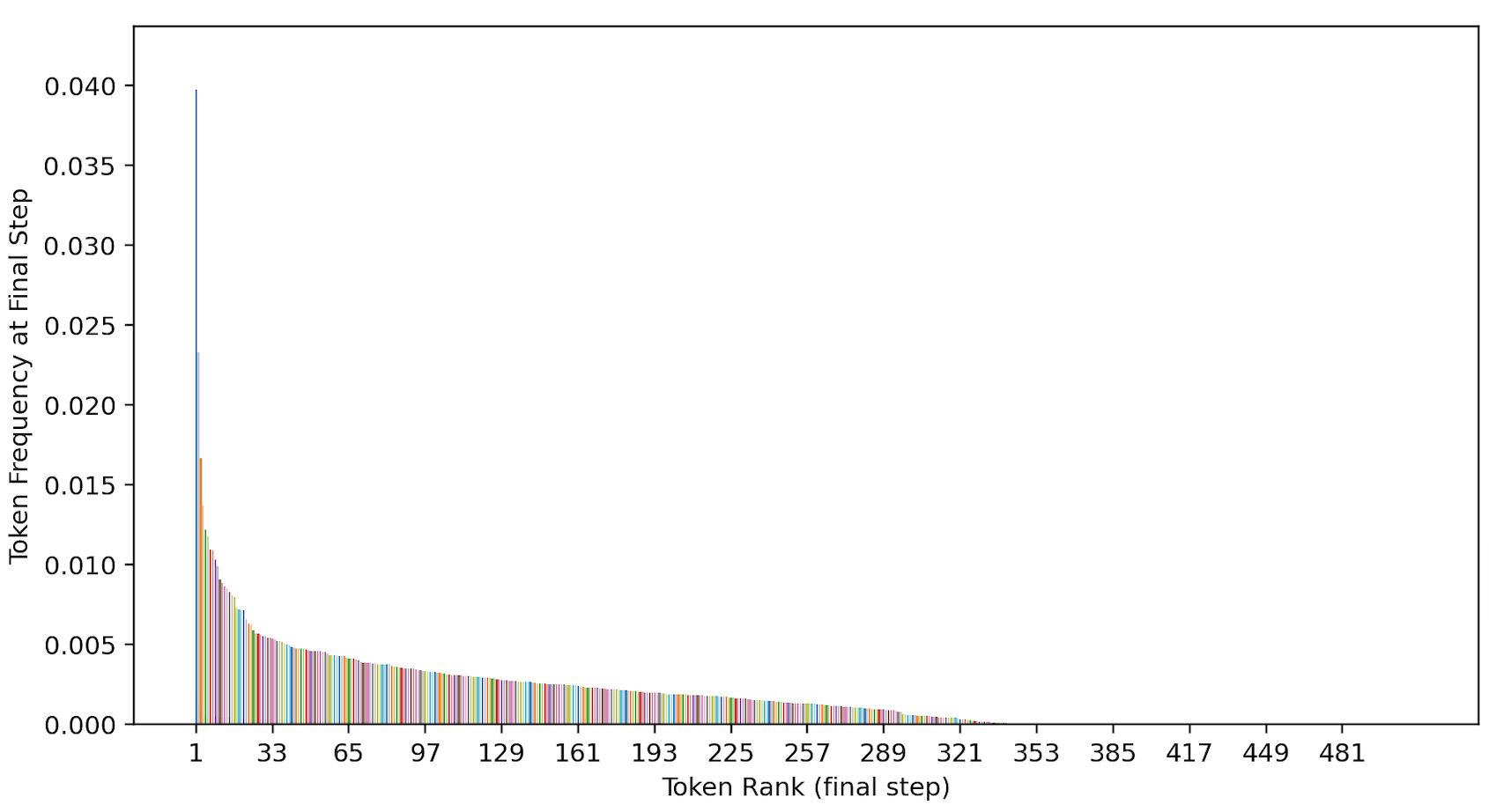}
\end{minipage}
\caption{Scaling ablation with $M = 512$ abstract token vocabulary.}
\label{fig:token-dist-m512}
\end{figure}

Figures \ref{fig:token-dist-m2}-\ref{fig:token-dist-m512} demonstrate the change in token frequency after the policy iteration warm-up (RL step 0) to after RL training, depicted by the subfigures on the left, and the frequency distribution by token rank after RL, given by the subfigures on the right. As shown in Figure \ref{fig:token-dist-evol}, RL training shapes the distribution of token usage, resulting in a power law-like distribution that is more evident at larger vocabulary sizes. We observe that using a max sequence length of 128 during training, as we scale the vocabulary size, there are more tokens in the tail that are relatively unused, despite the initial stage of warm-up involving a uniform schedule of abstract token selection. This suggests that the model learns to re-use certain tokens for frequently re-appearing concepts, while reserving others for more rare concepts. 

\subsubsection{Cold-Start RL Frequency Distribution}

\begin{figure}[h]
\centering
\begin{minipage}[t]{0.5\linewidth}
  \centering
  \includegraphics[width=\linewidth]{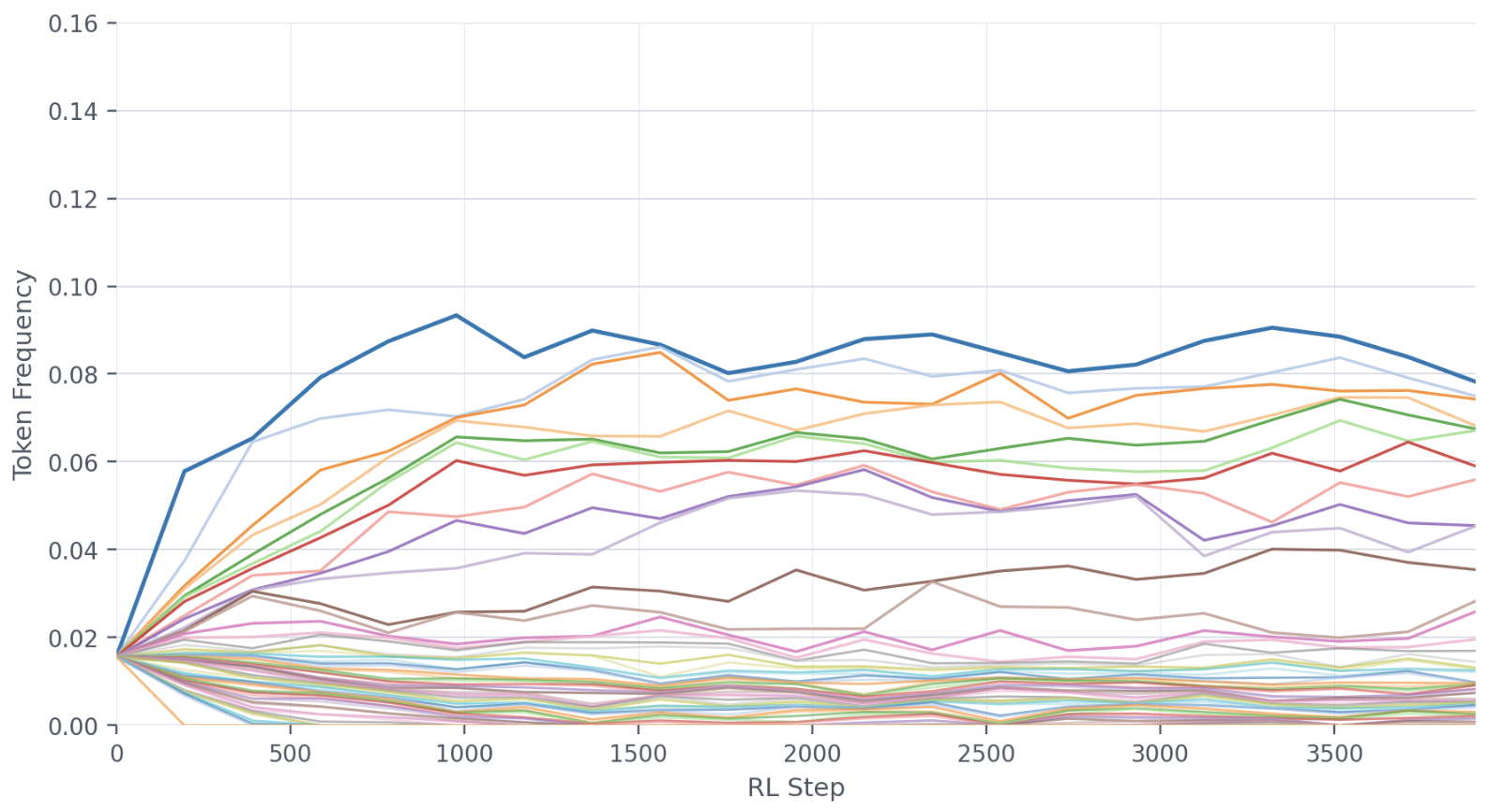}
\end{minipage}\hfill
\begin{minipage}[t]{0.5\linewidth}
  \centering
  \includegraphics[width=\linewidth]{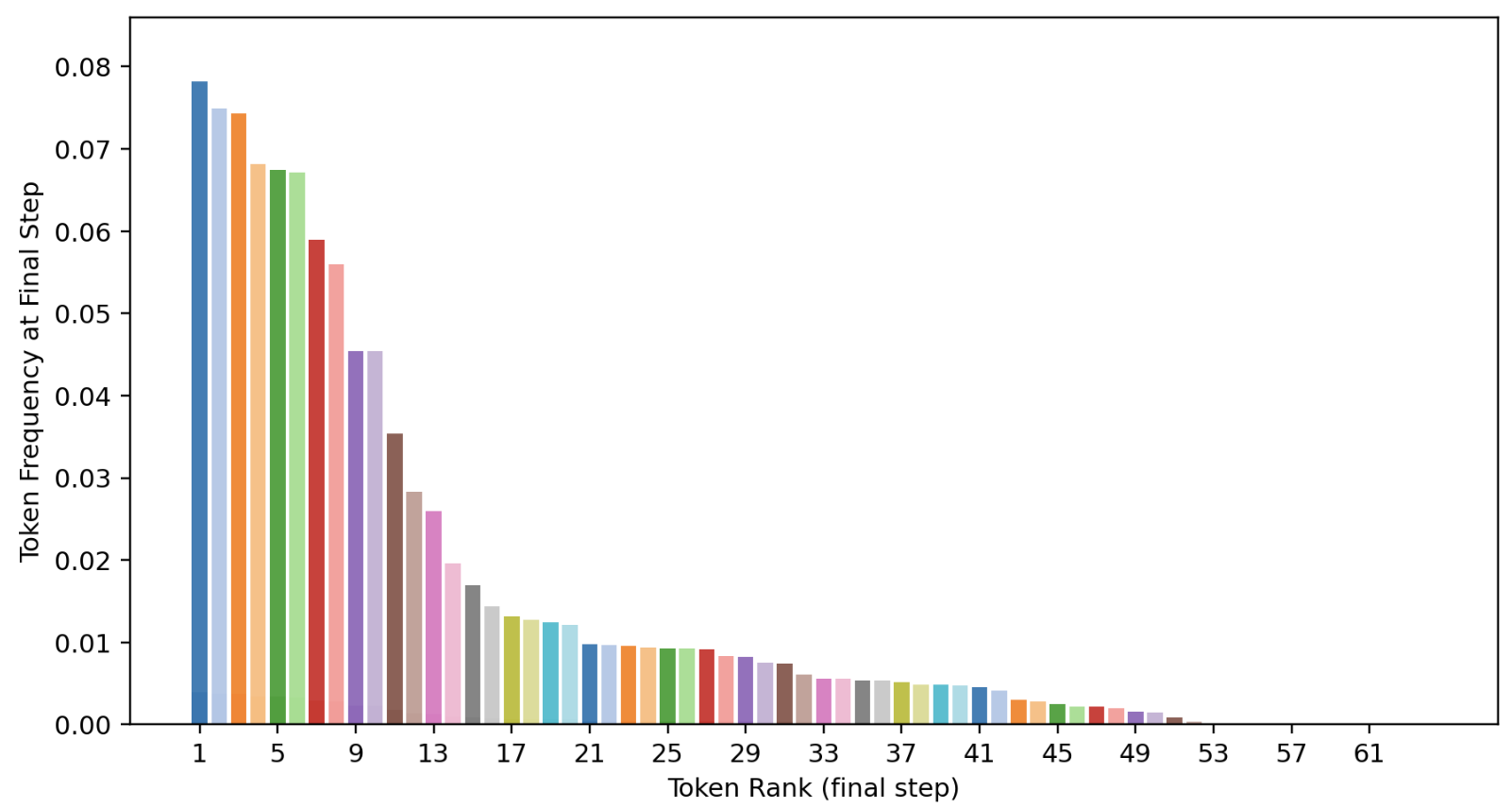}
\end{minipage}
\caption{Cold-start RL in the scaling ablation with $M = 64$ abstract token vocabulary. }
\label{fig:cold-start-m64}
\end{figure}

For comparison, we include the frequency distribution with $M = 64$ for cold-start RL training in Figure \ref{fig:cold-start-m64}. The starting frequency is uniform probability ($\frac{1}{M}$) given the embeddings are randomly initialized. We find a distribution that less closely resembles a power law compared to the warm-started RL. This demonstrates the efficacy of the warm-up toward its goal of embedding learning for the new vocabulary, creating a distribution of token usage that is further shaped through RL. On-policy generation, which happens in the self-distillation phase as well as the bottlenecked SFT stage with $t > 1$, allows the model to learn the relationship between successful sequences and a teacher (gold) response. By contrast, though cold-start RL quickly learns to use the token which ends up with the highest frequency (\atok{T})), several other tokens are eventually used with similar frequency while other tokens are very rarely used. It is possible that further scaling training compute -- specifically, rollouts and RL episodes -- could have a similar effect as a warm-up phase.

\subsection{Scaling Model Size: Qwen3-32B}\label{appendix:Qwen32B}

While we have studied the transferability of our method across model families (Qwen, Granite), it is also valuable to examine whether the method scales to larger model sizes. Thus, we study the performance of Abstract-CoT with Qwen3-32B, in comparison to baselines consistent with those reported in Table \ref{tab:main}. SFT training was performed with $8\times$NVIDIA H100 GPUs, and RL training was performed on $32\times$NVIDIA H100 GPUs. As with Qwen3-8B, the "thinking mode" is disabled. 

\begin{table*}[h]
\centering
\small
\begin{tabular}{lcc|cc|cc}
\toprule
\multirow{2}{*}{Method} & \multicolumn{2}{c|}{MATH-500} & \multicolumn{2}{c|}{AlpacaEval} & \multicolumn{2}{c}{HotpotQA} \\
 & Accuracy & Tokens & Win-rate & Tokens & F1 & Tokens \\
\midrule
Baseline & 86.8 & 1278 & 60.5 & 361 & 54.4 & 598 \\
Pause Tokens & 82.6 & 156 & 53.9 & 240 & 52.1 & 176 \\
Stepwise Internalization & 90.6 & 163 & 61.3 & 261 & 56.6 & 165  \\
SFT (no CoT) & 89.0 & 427 & 60.8 & 372 & 55.1 & 372 \\
SFT (CoT) & 93.4 & 1706 & 63.3 & 545 & 58.3 & 734 \\
SFT+RL & \textbf{95.0} & 1832 & \underline{65.2} & 608 & \underline{60.9} & 797 \\
\midrule
Abstract-CoT (RL-only) & 84.4 & 137 & 58.9 & 232 & 53.8 & 156 \\
Abstract-CoT (Warm-up) & 90.2 & 195 & 62.7 & 277 & 58.6 & 216 \\
Abstract-CoT (Warm-up + RL) & \underline{94.6} & 167 & \textbf{65.6} & 229 &  \textbf{62.1} & 180 \\
\bottomrule
\end{tabular}
\caption{Results on MATH-500 (accuracy), AlpacaEval (win-rate) and HotpotQA (F1) with Qwen3-32B, demonstrating efficiency gains and strong performance consistent with the trends exhibited with other models, highlighting the scalability of Abstract-CoT.}
\label{tab:Qwen32B-expts}
\end{table*}

The findings corroborate those in Table \ref{tab:main}: Abstract-CoT outperforms verbalized CoT (SFT + RL) on AlpacaEval and HotpotQA while using 2.7$\times$ and 4.4$\times$ fewer tokens, respectively, while nearly matching performance on MATH-500 with 11.0$\times$ fewer tokens. The 32B model appears to be slightly more verbose in its reasoning traces and response tokens than its 8B counterpart, resulting in slight increases in average tokens for all settings. 

\subsection{CoT Truncation Analysis}\label{appendix:halting}

While verbal chain-of-thought generates a large number of thinking tokens within its delimiters, truncating reasoning trace to a short length serves as a form of inference-time budget control and a mechanism to analyze the "compactness" of Abstract-CoT sequences. Prior works have sought to extend CoT lengths for reasoning tasks to analyze budget control, such as appending "Wait" to its generation to continue reasoning \citep{muennighoff-etal-2025-s1}. We limit the model to $k$ tokens by truncation: for Abstract-CoT, this means halting the CoT after $k$ tokens and appending the $\abend$ delimiter prior to response generation. The complete set of results across benchmarks and for $k = \{32, 48, 64\}$ are included in Table \ref{tab:trunc_all_benchmarks}. 

\begin{table}[H]
\centering
\small
\setlength{\tabcolsep}{8pt}
\renewcommand{\arraystretch}{1.15}
\begin{tabular}{lcccc}
\toprule
Method & Full CoT & 64 Tokens & 48 Tokens & 32 Tokens \\

\specialrule{0.08em}{0.35em}{0pt}
\multicolumn{5}{c}{\textbf{MATH-500}} \\
\specialrule{0.08em}{0pt}{0.2em}
Verbal CoT (SFT+RL)              & 92.6 & 84.8 & 84.0 & 80.9 \\
Abstract-CoT (PI-3+RL, $M{=}64$) & 90.8 & 87.1 & 86.4 & 84.6 \\

\specialrule{0.08em}{0.35em}{0pt}
\multicolumn{5}{c}{\textbf{AlpacaEval}} \\
\specialrule{0.08em}{0pt}{0.2em}
Verbal CoT (SFT+RL)              & 58.4 & 55.2 & 54.7 & 53.4 \\
Abstract-CoT (PI-3+RL, $M{=}64$) & 60.6 & 57.0 & 56.1 & 55.5 \\

\specialrule{0.08em}{0.35em}{0pt}
\multicolumn{5}{c}{\textbf{HotpotQA}} \\
\specialrule{0.08em}{0pt}{0.2em}
Verbal CoT (SFT+RL)              & 58.1 & 53.1 & 52.8 & 51.0 \\
Abstract-CoT (PI-3+RL, $M{=}64$) & 58.6 & 54.8 & 54.0 & 52.4 \\

\bottomrule
\end{tabular}
\caption{
Truncation sensitivity across benchmarks for Qwen3-8B. “Normal” denotes the untruncated setting. For verbal CoT, truncation is applied to the natural-language chain-of-thought; for Abstract-CoT, truncation is applied to the abstract trace. 
}
\label{tab:trunc_all_benchmarks}
\end{table}

The key findings noted in Section \ref{sec4.3:abstract-lang-analysis} are reflected across benchmarks: both methods clearly decrease, with a similar amount on AlpacaEval and HotpotQA, while there is a sizable discrepancy with MATH-500. AlpacaEval has the smallest drop due to having the fewest thinking tokens and more response tokens to start, as reflected in Table \ref{tab:main}.   Notably, the smoothness of degradation with Verbal CoT appears to be consistent with the number of  thinking tokens produced. Benchmarks like MATH-500 with more tokens saw a sharper drop, whereas AlpacaEval had a smaller decline, with HotpotQA in between. 

\subsection{Permutation Testing Analysis}\label{appendix:permutation}

As discussed in Section \ref{sec4.3:abstract-lang-analysis}, we conducted a permutation analysis to study the behavior of Abstract-CoT compared to Verbal CoT in its compositional power induced through RL. Recall that while the warm-up phase is intended to learn embeddings corresponding to the new abstract vocabulary, the RL phase is designed to learn sequences of abstract tokens that result in high-quality responses, as determined by a generative reward model. To this end, learning to effectively use the abstract vocabulary through constrained decoding should make the model more sensitive to CoT perturbations, making it less permutation-invariant. 

For each prompt in the evaluation set, we first generate a \{Verbal, Abstract\} CoT. Then, for verbal CoT experiments, we randomly permute the steps in the CoT based on the newline delimiter, and then induce the model to produce a response. For Abstract-CoT, we randomly permute the tokens in the generated sequence, given the absence of such delimiters therein. 

The complete set of results across MATH-500, AlpacaEval, and HotpotQA is included in Table \ref{tab:perm_all_benchmarks}. All 3 benchmarks exhibit a similar trend: both methods clearly decrease with permuted CoT, and the verbal CoT degrades by more than Abstract-CoT, but Abstract-CoT is still affected by a substantial amount. Consistent with the intuition above, RL training leads to greater degradation due to learning token sequences that produce better responses, and augmenting the CoT muddles the context, resulting in worse generations. Encouragingly, this appears to be reflected in Abstract-CoT, suggesting that further scaling training improves the model's ability to use the abstract vocabulary, leading to behavior resembling natural language even more closely. 

\begin{table}[H]
\centering
\small
\setlength{\tabcolsep}{8pt}
\renewcommand{\arraystretch}{1.15}
\begin{tabular}{lccc}
\toprule
Method & Direct & Permuted & Change ($\Delta$) \\

\specialrule{0.08em}{0.35em}{0pt}
\multicolumn{4}{c}{\textbf{MATH-500}} \\
\specialrule{0.08em}{0pt}{0.2em}
Verbal CoT (SFT)       & 89.8 & 81.8 & -8.0 \\
Verbal CoT (SFT+RL)    & 92.6 & 81.6 & -11.0 \\
Abstract-CoT (PI-3)    & 87.4 & 83.2 & -4.2 \\
Abstract-CoT (PI-3+RL) & 90.6 & 82.8 & -7.8 \\

\specialrule{0.08em}{0.35em}{0pt}
\multicolumn{4}{c}{\textbf{AlpacaEval}} \\
\specialrule{0.08em}{0pt}{0.2em}
Verbal CoT (SFT)       & 57.0 & 51.9 & -5.1 \\
Verbal CoT (SFT+RL)    & 58.4 & 50.4 & -8.0 \\
Abstract-CoT (PI-3)    & 55.6 & 51.9 & -3.7 \\
Abstract-CoT (PI-3+RL) & 60.3 & 54.3 & -6.0 \\

\specialrule{0.08em}{0.35em}{0pt}
\multicolumn{4}{c}{\textbf{HotpotQA}} \\
\specialrule{0.08em}{0pt}{0.2em}
Verbal CoT (SFT)       & 54.8 & 48.7 & -6.1 \\
Verbal CoT (SFT+RL)    & 58.1 & 47.6 & -10.5 \\
Abstract-CoT (PI-3)    & 53.3 & 48.0 & -5.3 \\
Abstract-CoT (PI-3+RL) & 57.9 & 49.2 & -8.7 \\

\bottomrule
\end{tabular}
\caption{
Permutation ablation on Qwen3-8B. For verbal CoT, we use turn-level permutation; for Abstract-CoT, we use fully random token permutation.
}
\label{tab:perm_all_benchmarks}
\end{table}

\section{Generative Reward Model Prompt}

The following prompt is used with \texttt{gpt-oss-20b} as a generative reward model in GRPO training with the ''medium'' thinking mode. 

\begin{promptbox}[colframe=NavyBlue,boxrule=1pt]
\begin{verbatim}
You are an expert evaluator assessing the quality of AI assistant responses. 
Your task is to score a response on a scale from 0 to 10. 

## Scoring Criteria 

Evaluate the response across these dimensions:

**Helpfulness (Does it address the user's needs?)** 
- Directly answers the question or completes the task  
- Provides appropriate level of detail 
- Anticipates follow-up needs

**Accuracy (Is the information correct?)** 
- Factually correct information 
- No hallucinations or fabrications 
- Appropriate caveats when uncertain 

**Clarity (Is it easy to understand?)** 
- Well-organized and structured 
- Clear language appropriate to context 
- Proper formatting when needed 

**Relevance (Does it stay on topic?)** 
- Addresses the actual query 
- Avoids unnecessary tangents 
- Appropriate scope and focus

**Safety & Harmlessness** 
- No harmful, offensive, or inappropriate content 
- Respectful and unbiased 
- Considers ethical implications 

## Scoring Scale

**10 - Exceptional**: Perfect or near-perfect response that excels across all 
criteria
**9 - Excellent**: Outstanding response with only trivial imperfections 
**8 - Very Good**: Strong response that fully addresses the query with minor 
areas for improvement 
**7 - Good**: Solid response that meets expectations but has some room for 
improvement
**6 - Above Average**: Decent response with noticeable limitations 
**5 - Average**: Adequate response that addresses the query but has clear gaps
**4 - Below Average**: Partially helpful but with significant issues in one 
or more criteria
**3 - Poor**: Minimally useful with major problems across multiple criteria
**2 - Very Poor**: Severe deficiencies, barely addresses the query
**1 - Extremely Poor**: Almost completely fails to address the query
**0 - Unacceptable**: Completely fails to address the query, contains harmful 
content, or is entirely inappropriate

## Your Task

Provide your evaluation as a JSON object with the following structure:

```json
{
  "score": <number between 0-10>,
  "reasoning": "<2-4 sentences explaining your score, highlighting key strengths 
  and weaknesses>"
} 
``` 
---

## Conversation Context
{CONVERSATION_HISTORY}

## Assistant Response to Evaluate
{RESPONSE_TO_SCORE}
\end{verbatim}
\end{promptbox}

\newpage
\section{Qualitative Examples of Abstract Chain-of-Thought}\label{appendix:examples}

\definecolor{exPrBg}{RGB}{229,231,250}
\definecolor{exPrFr}{RGB}{148,155,220}
\definecolor{exPrTi}{RGB}{200,204,240}
\definecolor{exAbBg}{RGB}{253,241,217}
\definecolor{exAbFr}{RGB}{200,162,95}
\definecolor{exAbTi}{RGB}{240,215,175}
\definecolor{exVbBg}{RGB}{220,245,245}
\definecolor{exVbFr}{RGB}{100,185,190}
\definecolor{exVbTi}{RGB}{175,225,228}
\definecolor{exArBg}{RGB}{252,235,222}
\definecolor{exArFr}{RGB}{195,148,100}
\definecolor{exArTi}{RGB}{240,212,180}
\definecolor{exRsBg}{RGB}{222,245,225}
\definecolor{exRsFr}{RGB}{115,190,125}
\definecolor{exRsTi}{RGB}{180,228,185}

\tcbset{
  exbase/.style={
    breakable, enhanced,
    boxrule=0.6pt, arc=2pt,
    left=6pt, right=6pt, top=4pt, bottom=4pt,
    fonttitle=\bfseries\small,
    before skip=3pt, after skip=3pt,
  },
  exPrompt/.style={exbase,
    colback=exPrBg,colframe=exPrFr,colbacktitle=exPrTi,coltitle=black},
  exAbstract/.style={exbase,
    colback=exAbBg,colframe=exAbFr,colbacktitle=exAbTi,coltitle=black},
  exAbsResponse/.style={exbase,
    colback=exArBg,colframe=exArFr,colbacktitle=exArTi,coltitle=black},
  exVerbal/.style={exbase,
    colback=exVbBg,colframe=exVbFr,colbacktitle=exVbTi,coltitle=black},
  exResponse/.style={exbase,
    colback=exRsBg,colframe=exRsFr,colbacktitle=exRsTi,coltitle=black},
}

\newcommand{\exsep}{\vspace{1em}\noindent\rule{\linewidth}{0.4pt}\vspace{0.6em}}

\subsection{Mathematical Problem Solving}

\subsubsection{Example 1: Combinatorics}
\begin{tcolorbox}[exPrompt,title={Prompt}]
\textbf{How many five-digit integers are divisible by 5 and have their digits sum to 20?}
\end{tcolorbox}
\begin{tcolorbox}[exAbstract,title={Abstract CoT}]
\small\abstart{}\atsep
\atok{R}\atsep\atok{C}\atsep\atok{M}\atsep\atok{BA}\atsep\atok{Q}\atsep\atok{AD}\atsep\atok{C}\atsep\atok{AH}\atsep\atok{S}\atsep\atok{M}\atsep\atok{R}\atsep\atok{C}\atsep\atok{BA}\atsep\atok{AD}\atsep\atok{BK}\atsep\atok{Q}\atsep
\atok{C}\atsep\atok{M}\atsep\atok{AJ}\atsep\atok{R}\atsep\atok{C}\atsep\atok{S}\atsep\atok{AD}\atsep\atok{BF}\atsep\atok{C}\atsep\atok{M}\atsep\atok{R}\atsep\atok{Q}\atsep\atok{C}\atsep\atok{BA}\atsep\atok{AD}\atsep\atok{BD}\atsep
\atok{C}\atsep\atok{R}\atsep\atok{AK}\atsep\atok{BE}\atsep\atok{M}\atsep\atok{C}\atsep\atok{AD}\atsep\atok{BA}\atsep\atok{C}\atsep\atok{BG}\atsep\atok{R}\atsep\atok{Q}\atsep\atok{M}\atsep\atok{C}\atsep\atok{S}\atsep\atok{BL}\atsep
\atok{Z}\atsep\atok{C}\atsep\atok{R}\atsep\atok{AD}\atsep\atok{C}\atsep\atok{BA}\atsep\atok{Q}\atsep\atok{AW}\atsep\atok{M}\atsep\atok{C}\atsep\atok{R}\atsep\atok{AD}\atsep\atok{S}\atsep\atok{BD}\atsep\atok{C}\atsep\atok{X}\atsep
\atok{BE}\atsep\atok{P}\atsep\atok{C}\atsep\atok{BF}\atsep\atok{Z}\atsep\atok{M}\atsep\atok{C}\atsep\atok{BG}\atsep\atok{R}\atsep\atok{AD}\atsep\atok{E}\atsep\atok{C}\atsep\atok{M}\atsep\atok{BH}\atsep\atok{Y}\atsep\atok{C}\atsep
\atok{BI}\atsep\atok{N}\atsep\atok{AD}\atsep\atok{AA}\atsep\atok{C}\atsep\atok{BK}\atsep\atok{S}\atsep\atok{R}\atsep\atok{C}\atsep\atok{Z}\atsep\atok{AD}\atsep\atok{M}\atsep\atok{BL}\atsep\atok{Q}\atsep\atok{C}\atsep\abend{}
\end{tcolorbox}
\begin{tcolorbox}[exAbsResponse,title={Abstract CoT Response}]
The total number of five-digit integers that are divisible by 5 with digit sum equal to 20 is:
\[
  \boxed{1116}
\]
\end{tcolorbox}
\vspace{0.5em}\noindent\hdashrule{\linewidth}{0.4pt}{2pt 2pt}\vspace{0.3em}
\begin{tcolorbox}[exPrompt,title={Prompt}]
\textbf{How many five-digit integers are divisible by 5 and have their digits sum to 20?}
\end{tcolorbox}
\begin{tcolorbox}[exVerbal,title={Verbal CoT}]
\small
To determine how many five-digit integers are divisible by 5 and have their digits sum to 20, we must adhere to two conditions:
\begin{itemize}\setlength\itemsep{1pt}
  \item \textbf{Divisibility by 5}: A number is divisible by 5 if its last digit is either 0 or 5.
  \item \textbf{Sum of digits equals 20}: Denote the digits as $d_1, d_2, d_3, d_4, d_5$, where $d_1 \geq 1$ (to avoid leading zeros), and we require:
  \[
    d_1 + d_2 + d_3 + d_4 + d_5 = 20
  \]
\end{itemize}
We can split the problem into two cases, based on the condition that the last digit $d_5 \in \{0, 5\}$, since the number must end in 0 or 5 for divisibility by 5. Each case will then become a subproblem of solving the digit sum under the modified constraint.

\textbf{Case 1: $d_5 = 0$}

Then, the sum of the other digits becomes:
\[
  d_1 + d_2 + d_3 + d_4 = 20
\]
Since $d_1 \geq 1$, we define $d_1' = d_1 - 1 \geq 0$, so that:
\[
  d_1' + d_2 + d_3 + d_4 = 19
\]
Now, the constraints are:
\begin{itemize}\setlength\itemsep{1pt}
  \item $d_1' \leq 8$ (since $d_1 \leq 9$)
  \item $d_2, d_3, d_4 \leq 9$
\end{itemize}
We now apply the principle of inclusion-exclusion over all non-negative integer solutions to the equation:
\[
  x_1 + x_2 + x_3 + x_4 = 19
\]
with $x_1 \leq 8,\; x_2, x_3, x_4 \leq 9$.

\textbf{Total number of solutions without constraints:}
\[
  \binom{19 + 4 - 1}{4 - 1} = \binom{22}{3} = 1540
\]

\textbf{Subtract invalid solutions:}
\begin{itemize}\setlength\itemsep{2pt}
  \item For $x_1 \geq 9$: Set $x_1' = x_1 - 9$, new sum: $x_1' + x_2 + x_3 + x_4 = 10$
  \[
    \binom{10 + 4 - 1}{4 - 1} = \binom{13}{3} = 286
  \]
  \item For each $x_i \geq 10$ ($i = 2, 3, 4$):
  \[
    \binom{19 - 10 + 4 - 1}{4 - 1} = \binom{12}{3} = 220
  \]
  (Total for $x_2, x_3, x_4$: $3 \times 220 = 660$)
\end{itemize}

\textbf{Add back overlaps:}

Any pair where one $x_i \geq 9$ and another $x_j \geq 10$ leads to:
\[
  x_i' + x_j' + \text{others} = 19 - 9 - 10 = 0 \;\Rightarrow\; \text{Only one solution } (0,\,0,\,0,\,0)
\]
There are 3 such combinations (with $d_1'$ and one of $d_2, d_3, d_4$), contributing:
\[
  3 \times 1 = 3
\]

\textbf{Final count:}
\[
  1540 - 286 - 660 + 3 = 597
\]

\textbf{Case 2: $d_5 = 5$}

Then, the sum of the other digits becomes:
\[
  d_1 + d_2 + d_3 + d_4 = 15
\]
Again, define $d_1' = d_1 - 1 \geq 0$, so:
\[
  d_1' + d_2 + d_3 + d_4 = 14
\]
Constraints:
\begin{itemize}\setlength\itemsep{1pt}
  \item $d_1' \leq 8$
  \item $d_2, d_3, d_4 \leq 9$
\end{itemize}

\textbf{Total number of solutions without constraints:}
\[
  \binom{14 + 4 - 1}{4 - 1} = \binom{17}{3} = 680
\]

\textbf{Subtract invalid solutions:}
\begin{itemize}\setlength\itemsep{2pt}
  \item For $x_1 \geq 9$: $\displaystyle\binom{14 - 9 + 4 - 1}{4 - 1} = \binom{8}{3} = 56$
  \item For each $x_i \geq 10$ ($i = 2, 3, 4$):
  \[
    \binom{14 - 10 + 4 - 1}{4 - 1} = \binom{7}{3} = 35
  \]
  (Total for $x_2, x_3, x_4$: $3 \times 35 = 105$)
\end{itemize}

\textbf{Add back overlaps:}

No two variables can have their limits exceeded simultaneously, as that would require more than 14, so all overlaps are zero.

\textbf{Final count:}
\[
  680 - 56 - 105 = 519
\]

\textbf{Final Calculation}

Summing both cases gives:
\[
  \text{Total} = 597 + 519 = \boxed{1116}
\]
\end{tcolorbox}
\begin{tcolorbox}[exResponse,title={Verbal CoT Response}]
Therefore, the number of five-digit integers divisible by 5 and with digits summing to 20 is:
\[
  \boxed{1116}
\]
\end{tcolorbox}

\exsep

\subsubsection{Example 2: Geometry}
\begin{tcolorbox}[exPrompt,title={Prompt}]
\textbf{Triangle $ABC$ has sides of length $a=13$, $b=14$, and $c=15$. Find the area of the triangle and the length of the altitude from vertex $A$ to side $b$.}
\end{tcolorbox}
\begin{tcolorbox}[exAbstract,title={Abstract CoT}]
\small\abstart{}\atsep
\atok{C}\atsep\atok{M}\atsep\atok{R}\atsep\atok{AD}\atsep\atok{Q}\atsep\atok{BA}\atsep\atok{C}\atsep\atok{AF}\atsep\atok{M}\atsep\atok{R}\atsep\atok{C}\atsep\atok{S}\atsep\atok{Q}\atsep\atok{AD}\atsep\atok{C}\atsep\atok{BH}\atsep
\atok{M}\atsep\atok{BA}\atsep\atok{C}\atsep\atok{R}\atsep\atok{AD}\atsep\atok{AV}\atsep\atok{Q}\atsep\atok{C}\atsep\atok{M}\atsep\atok{BD}\atsep\atok{R}\atsep\atok{C}\atsep\atok{BA}\atsep\atok{BG}\atsep\atok{AD}\atsep\atok{C}\atsep
\atok{M}\atsep\atok{Q}\atsep\atok{R}\atsep\atok{S}\atsep\atok{C}\atsep\atok{AD}\atsep\atok{K}\atsep\atok{BA}\atsep\atok{C}\atsep\atok{M}\atsep\atok{R}\atsep\atok{BE}\atsep\atok{C}\atsep\atok{Q}\atsep\atok{AD}\atsep\atok{BF}\atsep
\atok{M}\atsep\atok{C}\atsep\atok{R}\atsep\atok{AO}\atsep\atok{BA}\atsep\atok{AD}\atsep\atok{C}\atsep\atok{M}\atsep\atok{Q}\atsep\atok{S}\atsep\atok{R}\atsep\atok{C}\atsep\atok{BJ}\atsep\atok{AD}\atsep\atok{M}\atsep\atok{C}\atsep
\atok{BA}\atsep\atok{R}\atsep\atok{BI}\atsep\atok{Q}\atsep\atok{C}\atsep\atok{M}\atsep\atok{AD}\atsep\atok{R}\atsep\atok{C}\atsep\abend{}
\end{tcolorbox}
\begin{tcolorbox}[exAbsResponse,title={Abstract CoT Response}]
Applying Heron's formula with semi-perimeter $s = 21$, the area of triangle $ABC$ is $\boxed{84}$.

The altitude from vertex $A$ to side $BC$ (length 14) is then $h = \frac{2 \times 84}{14} = \boxed{12}$.
\end{tcolorbox}
\vspace{0.5em}\noindent\hdashrule{\linewidth}{0.4pt}{2pt 2pt}\vspace{0.3em}
\begin{tcolorbox}[exPrompt,title={Prompt}]
\textbf{Triangle $ABC$ has sides of length $a=13$, $b=14$, and $c=15$. Find the area of the triangle and the length of the altitude from vertex $A$ to side $b$.}
\end{tcolorbox}
\begin{tcolorbox}[exVerbal,title={Verbal CoT}]
\small
\textbf{Triangle Area Calculation.}
To determine the area of triangle $ABC$ with sides $a=13$, $b=14$, and $c=15$, we use Heron's Formula. Before applying the formula, we must calculate the semi-perimeter $s$ of the triangle, defined as:
\[
  s = \frac{a+b+c}{2}.
\]
Substituting the given values:
\[
  s = \frac{13+14+15}{2} = \frac{42}{2} = 21.
\]
Then, using Heron's Formula, the area $A$ of the triangle is:
\[
  A = \sqrt{s(s-a)(s-b)(s-c)}.
\]
Substitute the known values:
\[
  A = \sqrt{21(21-13)(21-14)(21-15)} = \sqrt{21\times 8\times 7\times 6}.
\]
Performing the multiplication step-by-step:
\[
  21\times 8 = 168, \quad 7\times 6 = 42, \quad 168\times 42 = 7056.
\]
Thus:
\[
  A = \sqrt{7056} = 84.
\]

\textbf{Altitude from Vertex $A$ to Side $b$.}
Next, we are asked to determine the length of the altitude from vertex $A$ to side $b$.

Assuming the triangle is labeled in the standard way, where:
\begin{itemize}\setlength\itemsep{1pt}
  \item Side $a = BC$ (opposite vertex $A$),
  \item Side $b = AC$ (opposite vertex $B$),
  \item Side $c = AB$ (opposite vertex $C$),
\end{itemize}
then side $b = AC$, and the altitude from vertex $A$ to side $b$ ($AC$) would not make sense, since vertex $A$ lies on side $AC$ itself, making the altitude zero---this is a contradiction.

Therefore, we consider an alternative labeling that matches the question's intent: if side $b = BC$, then the altitude from vertex $A$ to side $b$ (which is $BC$) is a meaningful and non-zero quantity.

To compute this, recall the formula for the area of a triangle as:
\[
  \text{Area} = \tfrac{1}{2}\times\text{base}\times\text{height}.
\]
Given: Area $= 84$, Base $=$ side $b = 14$. We solve for the height (altitude) $h$:
\[
  84 = \tfrac{1}{2}\times 14\times h \implies 84 = 7h \implies h = \frac{84}{7} = 12.
\]
\end{tcolorbox}
\begin{tcolorbox}[exResponse,title={Verbal CoT Response}]
The area of triangle $ABC$ is $\boxed{84}$.

The altitude from vertex $A$ to side $b$, interpreted as $BC$, is $\boxed{12}$.
\end{tcolorbox}

\exsep

\subsubsection{Example 3: Sequences \& Series}
\begin{tcolorbox}[exPrompt,title={Prompt}]
\textbf{Evaluate $\displaystyle\sum_{n=1}^{\infty} \dfrac{n^2}{2^n}$.}
\end{tcolorbox}
\begin{tcolorbox}[exAbstract,title={Abstract CoT}]
\small\abstart{}\atsep
\atok{M}\atsep\atok{C}\atsep\atok{Q}\atsep\atok{R}\atsep\atok{AD}\atsep\atok{BA}\atsep\atok{C}\atsep\atok{AM}\atsep\atok{M}\atsep\atok{R}\atsep\atok{S}\atsep\atok{C}\atsep\atok{AD}\atsep\atok{Q}\atsep\atok{C}\atsep\atok{BI}\atsep
\atok{BA}\atsep\atok{M}\atsep\atok{R}\atsep\atok{C}\atsep\atok{AD}\atsep\atok{Q}\atsep\atok{AT}\atsep\atok{C}\atsep\atok{M}\atsep\atok{BD}\atsep\atok{R}\atsep\atok{C}\atsep\atok{Q}\atsep\atok{AD}\atsep\atok{BA}\atsep\atok{C}\atsep
\atok{BG}\atsep\atok{M}\atsep\atok{R}\atsep\atok{Q}\atsep\atok{C}\atsep\atok{AD}\atsep\atok{S}\atsep\atok{C}\atsep\atok{BA}\atsep\atok{BK}\atsep\atok{M}\atsep\atok{R}\atsep\atok{C}\atsep\atok{Q}\atsep\atok{AD}\atsep\atok{C}\atsep
\atok{M}\atsep\atok{R}\atsep\atok{BA}\atsep\atok{BL}\atsep\atok{Q}\atsep\atok{C}\atsep\atok{AD}\atsep\atok{M}\atsep\atok{R}\atsep\atok{C}\atsep\atok{S}\atsep\atok{Q}\atsep\atok{AD}\atsep\atok{BA}\atsep\atok{C}\atsep\atok{M}\atsep
\atok{R}\atsep\atok{C}\atsep\atok{Q}\atsep\atok{AD}\atsep\atok{BA}\atsep\atok{AE}\atsep\atok{M}\atsep\atok{R}\atsep\atok{C}\atsep\atok{S}\atsep\atok{Q}\atsep\atok{AD}\atsep\atok{C}\atsep\atok{BH}\atsep\atok{BA}\atsep\atok{M}\atsep
\atok{R}\atsep\atok{Q}\atsep\atok{C}\atsep\atok{AD}\atsep\atok{M}\atsep\atok{R}\atsep\atok{BA}\atsep\atok{C}\atsep\atok{BF}\atsep\atok{Q}\atsep\atok{AD}\atsep\atok{C}\atsep\atok{M}\atsep\atok{R}\atsep\atok{S}\atsep\atok{BA}\atsep
\atok{C}\atsep\atok{Q}\atsep\atok{AD}\atsep\atok{M}\atsep\atok{R}\atsep\atok{C}\atsep\atok{BA}\atsep\atok{BD}\atsep\atok{Q}\atsep\atok{C}\atsep\abend{}
\end{tcolorbox}
\begin{tcolorbox}[exAbsResponse,title={Abstract CoT Response}]
Using the generating function identity $\displaystyle\sum_{n=0}^{\infty} n^2 r^n = \frac{r(1+r)}{(1-r)^3}$ evaluated at $r = \tfrac{1}{2}$, we obtain:
\[
  \sum_{n=1}^{\infty}\frac{n^2}{2^n} = \frac{\tfrac{1}{2}\cdot\tfrac{3}{2}}{\left(\tfrac{1}{2}\right)^3} = \boxed{6}.
\]
\end{tcolorbox}
\vspace{0.5em}\noindent\hdashrule{\linewidth}{0.4pt}{2pt 2pt}\vspace{0.3em}
\begin{tcolorbox}[exPrompt,title={Prompt}]
\textbf{Evaluate $\displaystyle\sum_{n=1}^{\infty} \dfrac{n^2}{2^n}$.}
\end{tcolorbox}
\begin{tcolorbox}[exVerbal,title={Verbal CoT}]
\small
To evaluate the infinite series $\displaystyle\sum_{n=1}^{\infty}\frac{n^2}{2^n}$, we recognize that this is a weighted sum where both the numerator and denominator are functions of $n$, with the exponential decay $2^n$ in the denominator. This type of series is well-suited for analysis using generating functions and differentiation techniques.

\textbf{Step 1: Start with the Basic Generating Function.}
We recall the standard geometric series:
\[
  S = \sum_{n=0}^{\infty} r^n = \frac{1}{1-r}, \quad\text{for } |r|<1.
\]
Differentiating both sides with respect to $r$, we get:
\[
  \frac{d}{dr}\,S = \sum_{n=0}^{\infty} n\,r^{n-1} = \frac{1}{(1-r)^2}.
\]
Multiplying both sides by $r$, we obtain:
\[
  \sum_{n=0}^{\infty} n\,r^n = \frac{r}{(1-r)^2}.
\]
Now we take the derivative of this expression again with respect to $r$, to obtain the generating function for $n^2 r^n$:
\[
  \frac{d}{dr}\!\left(\sum_{n=0}^{\infty} n\,r^n\right) = \sum_{n=0}^{\infty} n^2\,r^{n-1} = \frac{d}{dr}\!\left(\frac{r}{(1-r)^2}\right).
\]
We now compute the right-hand side:
\[
  \frac{d}{dr}\!\left(\frac{r}{(1-r)^2}\right) = \frac{(1-r)^2\cdot 1 + r\cdot 2(1-r)}{(1-r)^4} = \frac{(1-r)^2 + 2r(1-r)}{(1-r)^4}.
\]
Simplify the numerator:
\[
  (1-r)^2 + 2r(1-r) = 1-2r+r^2+2r-2r^2 = 1-r^2.
\]
So, the derivative becomes:
\[
  \sum_{n=0}^{\infty} n^2\,r^{n-1} = \frac{1-r^2}{(1-r)^4} = \frac{1+r}{(1-r)^3}.
\]
Multiplying through by $r$ to adjust the exponent:
\[
  \sum_{n=0}^{\infty} n^2\,r^n = \frac{r(1+r)}{(1-r)^3}.
\]
Thus, we obtain the well-known closed-form generating function $\displaystyle\sum_{n=0}^{\infty} n^2\,r^n = \frac{r(1+r)}{(1-r)^3}$.

\textbf{Step 2: Evaluate the Series for $r = \tfrac{1}{2}$.}
Now plug in $r = \tfrac{1}{2}$ into our closed-form expression:
\[
  \sum_{n=0}^{\infty} n^2\!\left(\frac{1}{2}\right)^{\!n}
  = \frac{\frac{1}{2}\!\left(1+\frac{1}{2}\right)}{\left(1-\frac{1}{2}\right)^3}
  = \frac{\frac{1}{2}\cdot\frac{3}{2}}{\left(\frac{1}{2}\right)^3}
  = \frac{\frac{3}{4}}{\frac{1}{8}}
  = 6.
\]
Note that the original series starts at $n=1$, and the term for $n=0$ is zero ($0^2/2^0 = 0$), so:
\[
  \sum_{n=1}^{\infty}\frac{n^2}{2^n} = \sum_{n=0}^{\infty}\frac{n^2}{2^n} = 6.
\]

\textbf{Step 3: Alternative Verification Using Summation by Parts.}
We can alternatively express $n^2$ as $n(n-1)+n$:
\[
  \sum_{n=1}^{\infty}\frac{n^2}{2^n} = \sum_{n=1}^{\infty}\frac{n(n-1)}{2^n} + \sum_{n=1}^{\infty}\frac{n}{2^n}.
\]
We already know from earlier that $\displaystyle\sum_{n=1}^{\infty}\frac{n}{2^n}=2$ and $\displaystyle\sum_{n=1}^{\infty}\frac{n(n-1)}{2^n}=4$. Adding both parts:
\[
  2 + 4 = 6.
\]
Thus, we again arrive at the same result.

\textbf{Step 4: Numerical Partial Sums Approximation.}
To support our analytical result, we can also compute partial sums of the series:
\[
  \sum_{n=1}^{20}\frac{n^2}{2^n} \approx 5.9996.
\]
Since the terms decay exponentially, the convergence is rapid, and the partial sum approaches 6 closely, reinforcing the analytical result.
\end{tcolorbox}
\begin{tcolorbox}[exResponse,title={Verbal CoT Response}]
Through multiple rigorous and independent approaches---including generating functions, summation techniques, and numerical approximations---we consistently find that the value of the infinite series
\[
  \sum_{n=1}^{\infty}\frac{n^2}{2^n}
\]
is:
\[
  \boxed{6}
\]
\end{tcolorbox}

\exsep

\vspace{-3mm}
\subsubsection{Example 4: Probability}
\begin{tcolorbox}[exPrompt,title={Prompt}]
\textbf{A fair eight-sided die is rolled repeatedly until every face has appeared at least once. What is the expected total number of rolls?}
\end{tcolorbox}
\begin{tcolorbox}[exAbstract,title={Abstract CoT}]
\small\abstart{}\atsep
\atok{C}\atsep\atok{M}\atsep\atok{Q}\atsep\atok{R}\atsep\atok{BA}\atsep\atok{AD}\atsep\atok{S}\atsep\atok{C}\atsep\atok{AU}\atsep\atok{M}\atsep\atok{R}\atsep\atok{Q}\atsep\atok{C}\atsep\atok{AD}\atsep\atok{BA}\atsep\atok{C}\atsep
\atok{BJ}\atsep\atok{M}\atsep\atok{R}\atsep\atok{C}\atsep\atok{Q}\atsep\atok{S}\atsep\atok{AD}\atsep\atok{C}\atsep\atok{M}\atsep\atok{BA}\atsep\atok{AX}\atsep\atok{R}\atsep\atok{C}\atsep\atok{Q}\atsep\atok{AD}\atsep\atok{C}\atsep
\atok{BD}\atsep\atok{M}\atsep\atok{R}\atsep\atok{BA}\atsep\atok{C}\atsep\atok{Q}\atsep\atok{AD}\atsep\atok{S}\atsep\atok{C}\atsep\atok{BF}\atsep\atok{M}\atsep\atok{R}\atsep\atok{C}\atsep\atok{Q}\atsep\atok{BA}\atsep\atok{AD}\atsep
\atok{M}\atsep\atok{C}\atsep\atok{R}\atsep\atok{BH}\atsep\atok{Q}\atsep\atok{C}\atsep\atok{AD}\atsep\atok{BA}\atsep\atok{M}\atsep\atok{R}\atsep\atok{S}\atsep\atok{C}\atsep\atok{Q}\atsep\atok{AD}\atsep\atok{BL}\atsep\atok{C}\atsep
\atok{M}\atsep\atok{R}\atsep\atok{Q}\atsep\atok{C}\atsep\atok{BA}\atsep\atok{AD}\atsep\atok{AJ}\atsep\atok{M}\atsep\atok{R}\atsep\atok{C}\atsep\atok{S}\atsep\atok{Q}\atsep\atok{AD}\atsep\atok{C}\atsep\atok{BE}\atsep\atok{BA}\atsep
\atok{M}\atsep\atok{R}\atsep\atok{Q}\atsep\atok{C}\atsep\atok{AD}\atsep\atok{M}\atsep\atok{BA}\atsep\atok{C}\atsep\atok{R}\atsep\atok{BG}\atsep\atok{Q}\atsep\abend{}
\end{tcolorbox}
\begin{tcolorbox}[exAbsResponse,title={Abstract CoT Response}]
By the coupon collector formula $E(n) = n \cdot H_n$ with $n = 8$ and $H_8 = \tfrac{761}{280}$, the expected number of rolls to see all eight faces is:
\[
  \boxed{\dfrac{761}{35}} \approx 21.74.
\]
\end{tcolorbox}
\vspace{0.5em}\noindent\hdashrule{\linewidth}{0.4pt}{2pt 2pt}\vspace{0.3em}
\begin{tcolorbox}[exPrompt,title={Prompt}]
\textbf{A fair eight-sided die is rolled repeatedly until every face has appeared at least once. What is the expected total number of rolls?}
\end{tcolorbox}
\begin{tcolorbox}[exVerbal,title={Verbal CoT}]
\small
To determine the expected total number of rolls required to roll a fair eight-sided die until all eight faces have appeared at least once, we approach this problem using the coupon collector problem, a well-known stochastic model in probability theory.

\textbf{Problem Understanding.}
We are rolling a fair eight-sided die repeatedly, and we want to know the expected number of rolls required to observe all eight sides. Each roll independently produces one of the eight faces, each with equal probability (i.e., $\tfrac{1}{8}$). We want to find the expected value of the number of rolls until all outcomes have been observed. This is a direct application of the coupon collector problem with $n=8$.

\textbf{Modeling the Process.}
Let us define the expected number of rolls needed to observe all $n$ distinct outcomes as $E(n)$. In general, the formula for this expectation is:
\[
  E(n) = n \cdot H_n
\]
where $H_n = 1+\tfrac{1}{2}+\tfrac{1}{3}+\cdots+\tfrac{1}{n}$ is the $n$th harmonic number. For $n=8$, this becomes:
\[
  E(8) = 8\left(1+\frac{1}{2}+\frac{1}{3}+\frac{1}{4}+\frac{1}{5}+\frac{1}{6}+\frac{1}{7}+\frac{1}{8}\right).
\]

\textbf{Computing the Harmonic Number.}
We compute $H_8$ by summing the reciprocals:
\[
  H_8 = 1+\frac{1}{2}+\frac{1}{3}+\frac{1}{4}+\frac{1}{5}+\frac{1}{6}+\frac{1}{7}+\frac{1}{8}.
\]
To compute this exactly, we find a common denominator. The least common multiple of the denominators from 1 to 8 is 840. Converting each term:
\[
\renewcommand{\arraystretch}{1.2}
\begin{array}{rlrl}
  1 &= \dfrac{840}{840}, &\quad \dfrac{1}{2} &= \dfrac{420}{840}, \\[4pt]
  \dfrac{1}{3} &= \dfrac{280}{840}, &\quad \dfrac{1}{4} &= \dfrac{210}{840}, \\[4pt]
  \dfrac{1}{5} &= \dfrac{168}{840}, &\quad \dfrac{1}{6} &= \dfrac{140}{840}, \\[4pt]
  \dfrac{1}{7} &= \dfrac{120}{840}, &\quad \dfrac{1}{8} &= \dfrac{105}{840}.
\end{array}
\]
Adding the numerators:
\[
  840+420+280+210+168+140+120+105 = 2283.
\]
Thus:
\[
  H_8 = \frac{2283}{840}.
\]
Simplifying by dividing numerator and denominator by 3:
\[
  H_8 = \frac{761}{280}.
\]

\textbf{Final Computation.}
Now multiply by $n=8$ to get the expected number of rolls:
\[
  E(8) = 8\cdot\frac{761}{280} = \frac{6088}{280}.
\]
Simplify the fraction by dividing numerator and denominator by 8:
\[
  \frac{6088}{280} = \frac{761}{35}.
\]

\emph{Sanity check.} $761/35 \approx 21.743$, which is reasonable for 8 coupons (the answer for 6 coupons is $14.7$, and the harmonic series grows logarithmically, so we expect a moderate increase). $\checkmark$
\end{tcolorbox}
\begin{tcolorbox}[exResponse,title={Verbal CoT Response}]
Through the application of the coupon collector model, harmonic numbers, and careful arithmetic, we have derived that the expected number of rolls required to see all eight faces of a fair eight-sided die is:
\[
  \boxed{\dfrac{761}{35}} \approx 21.74.
\]
\end{tcolorbox}

\subsection{General Instruction-Following}

\subsubsection{Example 1: Lifestyle Advice}

\begin{tcolorbox}[exPrompt,title={Prompt}]
\textbf{I’m starting high school and I procrastinate a lot. Can you give me a simple plan to stop procrastinating that I can actually follow, even on busy weeks?}
\end{tcolorbox}
\begin{tcolorbox}[exAbstract,title={Abstract CoT}]
\small\abstart{}\atsep
\atok{T}\atsep\atok{F}\atsep\atok{C}\atsep\atok{L}\atsep\atok{AD}\atsep\atok{T}\atsep\atok{C}\atsep\atok{AE}\atsep\atok{F}\atsep\atok{L}\atsep\atok{R}\atsep\atok{T}\atsep\atok{C}\atsep\atok{F}\atsep\atok{BA}\atsep\atok{L}\atsep
\atok{T}\atsep\atok{AD}\atsep\atok{F}\atsep\atok{C}\atsep\atok{L}\atsep\atok{T}\atsep\atok{AN}\atsep\atok{F}\atsep\atok{C}\atsep\atok{L}\atsep\atok{BB}\atsep\atok{T}\atsep\atok{AD}\atsep\atok{F}\atsep\atok{C}\atsep\atok{M}\atsep
\atok{L}\atsep\atok{T}\atsep\atok{BC}\atsep\atok{F}\atsep\atok{S}\atsep\atok{C}\atsep\atok{L}\atsep\atok{BA}\atsep\atok{T}\atsep\atok{F}\atsep\atok{BD}\atsep\atok{C}\atsep\atok{AE}\atsep\atok{L}\atsep\atok{T}\atsep\atok{F}\atsep
\atok{AD}\atsep\atok{BF}\atsep\atok{C}\atsep\atok{L}\atsep\atok{T}\atsep\atok{F}\atsep\atok{L}\atsep\atok{BG}\atsep\atok{C}\atsep\atok{T}\atsep\atok{AD}\atsep\atok{F}\atsep\atok{C}\atsep\atok{BH}\atsep\atok{I}\atsep\atok{L}\atsep
\atok{T}\atsep\atok{F}\atsep\atok{BA}\atsep\atok{AE}\atsep\atok{C}\atsep\atok{T}\atsep\atok{W}\atsep\atok{L}\atsep\atok{F}\atsep\atok{C}\atsep\atok{AD}\atsep\atok{T}\atsep\atok{F}\atsep\atok{L}\atsep\abend{}
\end{tcolorbox}
\begin{tcolorbox}[exAbsResponse,title={Abstract CoT Response}]
\small
Here's a realistic, low-effort system that works even on your craziest weeks.
\\ \\
\textbf{The core idea:} don't try to ``fix'' procrastination---just make starting easier than avoiding. Every morning, write down one thing that matters most today. That's your anchor task. Before anything else, spend 15 minutes on it. Not an hour, not ``until it's done''---just 15 minutes. Most days, you'll keep going once you've started. On bad days, you still moved the needle.
\\ \\
For everything else: if it takes under two minutes, do it now. If it feels too big, shrink it (``study for the test'' becomes ``re-read one page of notes''). Set a 30--60 minute block each day where your phone is out of reach---this is your focus window. On busy weeks, cut it to 15 minutes. That still counts.
\\ \\
The secret nobody tells you: you'll still procrastinate sometimes. That's fine. Don't beat yourself up about it---guilt just makes the cycle worse. Just reset and pick it up tomorrow. Consistency over time beats perfection on any single day.
\end{tcolorbox}
\vspace{0.5em}\noindent\hdashrule{\linewidth}{0.4pt}{2pt 2pt}\vspace{0.3em}
\begin{tcolorbox}[exPrompt,title={Prompt}]
\textbf{I’m starting high school and I procrastinate a lot. Can you give me a simple plan to stop procrastinating that I can actually follow, even on busy weeks?}
\end{tcolorbox}
\begin{tcolorbox}[exVerbal,title={Verbal CoT}]
\small
Starting high school is a big transition, and it's completely normal to feel overwhelmed, especially when you're trying to manage new responsibilities, social dynamics, and academic expectations. Procrastination often stems from a mix of factors: fear of failure, perfectionism, feeling overwhelmed by tasks, or simply not knowing where to start. The key to overcoming procrastination isn't about eliminating it entirely (which is impossible) but about managing it through intentional habits and mindset shifts. The goal is to create a plan that's flexible, simple, and adaptable to even the busiest weeks. Let's break this down step by step.
\\ \\
First, you need to understand why you procrastinate. Are you avoiding tasks because they feel too hard? Do you fear that your work won't be perfect? Are you overwhelmed by the sheer volume of things to do? Once you identify the root causes, you can tailor your strategy. But even if you're unsure, starting with small, actionable steps will help you build momentum. The plan needs to be simple, relatable, and non-judgmental---because if it feels like a burden, you'll be less likely to stick with it.
\\ \\
The most effective anti-procrastination strategies often involve breaking tasks into smaller parts, creating structure, and using time management techniques like the Pomodoro Technique or time blocking. However, these methods can feel overwhelming if you're not used to them. The solution is to start gradually and pair each habit with a specific, low-stakes goal. For example, instead of trying to ``organize your entire schedule,'' begin with one small task, like setting a daily 15-minute goal to plan your homework.
\\ \\
Another critical factor is self-compassion. Procrastination is often linked to perfectionism or self-criticism. If you're too hard on yourself for not completing tasks, you're more likely to procrastinate again. The plan must include a way to reframe setbacks as learning opportunities rather than failures.
\\ \\
Finally, the plan needs to account for real-life interruptions and busy weeks. High school can be unpredictable, so your strategy should allow for flexibility. For instance, if a task takes longer than expected, you shouldn't feel guilty---adjust your schedule and move forward.
\end{tcolorbox}
\begin{tcolorbox}[exResponse,title={Verbal CoT Response}]
\small
Starting high school can be exciting, but procrastination can feel overwhelming. Here’s a simple, flexible plan to help you manage your time and stay on track:
\begin{enumerate}
    \item \textbf{Start Small --- The 2-Minute Rule:} If a task takes less than 2 minutes, do it immediately (e.g., answer an email, organize your desk). This prevents small tasks from piling up.
    \item \textbf{Prioritize with ``Most Important Task First’’ (MIT):} Every day, identify one task that is the most important (e.g., finishing a big project, studying for a test). Do this task first, even if it is short.
    \item \textbf{Break Tasks into Smaller Steps:} If a task feels too big, split it into tiny steps (e.g., ``Write a 2-page essay’’ $\rightarrow$ ``Outline 1 paragraph today’’). Check off each step as you go.
    \item \textbf{Use the 15-Minute Rule:} If you are stuck, promise yourself: ``I’ll work on this for 15 minutes. If I still don’t want to, I’ll stop.’’ This tricks your brain into starting --- and often you will keep going!
    \item \textbf{Schedule Daily ``Focus Time’’:} Block 30--60 minutes each day for focused work (e.g., homework, studying). If you are busy, shorten the time or do it in chunks.
\end{enumerate}
You’ve got this! Start with one or two strategies, and adjust as you go. Over time, you will build habits that make high school manageable and even enjoyable!
\end{tcolorbox}

\exsep
\subsubsection{Example 2: Workplace Communications}

\begin{tcolorbox}[exPrompt,title={Prompt}]
\textbf{A colleague of mine frequently interrupts me mid-sentence during team meetings. I want to address it directly but without damaging the relationship. Give me a short script I can actually use.}
\end{tcolorbox}
\begin{tcolorbox}[exAbstract,title={Abstract CoT}]
\small\abstart{}\atsep
\atok{F}\atsep\atok{T}\atsep\atok{C}\atsep\atok{L}\atsep\atok{AD}\atsep\atok{F}\atsep\atok{C}\atsep\atok{BA}\atsep\atok{T}\atsep\atok{L}\atsep\atok{AE}\atsep\atok{F}\atsep\atok{C}\atsep\atok{Q}\atsep\atok{T}\atsep\atok{L}\atsep
\atok{F}\atsep\atok{BB}\atsep\atok{T}\atsep\atok{AD}\atsep\atok{C}\atsep\atok{L}\atsep\atok{F}\atsep\atok{T}\atsep\atok{C}\atsep\atok{AQ}\atsep\atok{L}\atsep\atok{BC}\atsep\atok{F}\atsep\atok{C}\atsep\atok{AD}\atsep\atok{T}\atsep
\atok{L}\atsep\atok{F}\atsep\atok{C}\atsep\atok{BD}\atsep\atok{S}\atsep\atok{T}\atsep\atok{BA}\atsep\atok{C}\atsep\atok{L}\atsep\atok{BE}\atsep\atok{F}\atsep\atok{T}\atsep\atok{C}\atsep\atok{AE}\atsep\atok{L}\atsep\atok{F}\atsep
\atok{AD}\atsep\atok{BG}\atsep\atok{T}\atsep\atok{C}\atsep\atok{L}\atsep\atok{F}\atsep\atok{W}\atsep\atok{T}\atsep\atok{BH}\atsep\atok{C}\atsep\atok{L}\atsep\atok{AD}\atsep\atok{F}\atsep\atok{C}\atsep\atok{BI}\atsep\atok{T}\atsep
\atok{L}\atsep\atok{F}\atsep\atok{BA}\atsep\atok{BJ}\atsep\atok{C}\atsep\atok{T}\atsep\atok{W}\atsep\atok{L}\atsep\atok{F}\atsep\atok{AD}\atsep\atok{C}\atsep\atok{T}\atsep\atok{L}\atsep\atok{F}\atsep\atok{C}\atsep\abend{}
\end{tcolorbox}
\begin{tcolorbox}[exAbsResponse,title={Abstract CoT Response}]
\small
\textit{[One-on-one, relaxed setting]}
\\ \\
\smallskip
``Hey [Name], I wanted to bring up something small---during the brainstorming session last week, I noticed I got cut off mid-thought a couple of times. I know you're really engaged and eager to jump in, which I genuinely appreciate. But when I lose my train of thought, it's harder for me to contribute effectively. Could we try letting each other finish before responding? And honestly, if I ever do the same to you, please call me out---I want this to go both ways.''
\\ \\
\smallskip
\textit{[Pause, then warmly]}
\\ \\
\smallskip
``I really value how much energy you bring to our meetings---that's what makes them work.''
\end{tcolorbox}
\vspace{0.5em}\noindent\hdashrule{\linewidth}{0.4pt}{2pt 2pt}\vspace{0.3em}
\begin{tcolorbox}[exPrompt,title={Prompt}]
\textbf{A colleague of mine frequently interrupts me mid-sentence during team meetings. I want to address it directly but without damaging the relationship. Give me a short script I can actually use.}
\end{tcolorbox}
\begin{tcolorbox}[exVerbal,title={Verbal CoT}]
\small
This is a classic example of a nuanced communication challenge---balancing honesty with empathy, directness with diplomacy. The colleague’s interruption isn’t necessarily malicious; it could be a habit, a sign of eagerness to contribute, or even a lack of awareness about how their behavior affects others. The key is to address it without framing it as a personal attack. The stakes are high: if handled poorly, this can escalate into resentment or defensiveness, creating a rift in a professional relationship. But if approached with care, it can become an opportunity to model mutual respect and improve team dynamics.
\\ \\
Let’s break down the elements of this scenario. First, the context---team meetings are collaborative spaces where clarity and inclusion matter. When someone interrupts, it disrupts the flow of ideas and can leave others feeling unheard. The colleague might not realize they’re doing it, or they might assume their input is urgent. Either way, interrupting mid-sentence can undermine the meeting’s purpose. The challenge is to redirect that behavior without making the colleague feel criticized.
\\ \\
Now, the tone needs to be collaborative. Starting with a complaint (``You’re always cutting me off’’) immediately puts the other person on the defensive. Instead, framing the issue as a shared goal---like ``I want to make sure we’re all heard’’ or ``I need to finish my thoughts to make sure we’re aligned’’---shifts the focus from blame to partnership. It’s not about pointing fingers; it’s about ensuring the team functions smoothly.
\\ \\
The setting is critical. Public confrontation risks embarrassment or defensiveness. A private, one-on-one conversation allows for a more thoughtful exchange without putting the colleague on the spot. Timing matters too---addressing it immediately after an interruption might feel accusatory, but a calm, separate moment gives both people space to reflect.
\\ \\
The language needs to be constructive. Using ``I’’ statements (e.g., ``I find it hard to finish my thoughts when interrupted’’) avoids sounding accusatory. Specificity is key too: referencing a recent instance (``Last week in the brainstorming session’’) grounds the feedback in reality rather than vague frustration. It also shows you’ve paid attention, which builds trust.
\\ \\
Charitable framing is another layer---acknowledging that the colleague likely doesn’t intend to be dismissive helps them see the issue as a shared challenge. Ending with a reciprocity offer (``Would you let me know if you feel like I’m interrupting you?’’) turns the conversation into a mutual check-in, reinforcing collaboration.
\\ \\
The script needs to be concise but meaningful. It should feel natural, not rehearsed. The three beats---open (acknowledge the moment and low stakes), deliver (specific observation + impact + framing), close (clear ask + reciprocity)---create a rhythm that guides the conversation without forcing it. The goal is not to ``win’’ the conversation but to create a space where both people feel heard and respected.
\\ \\
So, the script must balance clarity with care, specificity with simplicity, and directness with diplomacy. It’s not about fixing the colleague’s behavior in a single interaction but planting the seed of awareness and mutual accountability. The more the colleague feels seen and respected, the more likely they’ll adjust their behavior.
\end{tcolorbox}
\begin{tcolorbox}[exResponse,title={Verbal CoT Response}]
\small
\textit{[Private, calm tone]}
\\ \\
\smallskip
``Hey [Name], just wanted to check in---was there a moment earlier when you interrupted me mid-sentence? I noticed it happened a couple of times during the brainstorming session last week. I know you’re passionate about contributing, and I really value your input, but it’s hard for me to finish my thoughts when I’m cut off. I don’t want to assume you’re dismissive---maybe it’s just a habit or a natural way you communicate. If we could both wait for each other to finish speaking, I think it’d make our discussions even more productive. Would you be open to giving me a heads-up if you ever feel like I’m interrupting you? It’s a small adjustment, but I think it’d help us both feel more heard.’’
\\ \\
\smallskip
\textit{[Pause, then smile]}
\\ \\
\smallskip
``Thanks for being so engaged---I think that’s exactly what makes our meetings work.’’
\end{tcolorbox}

\exsep

\vspace{-3mm}
\subsubsection{Example 3: Technical Explanation}

\begin{tcolorbox}[exPrompt,title={Prompt}]
\textbf{Explain how public-key cryptography works to someone who understands basic algebra but has no computer science background.}
\end{tcolorbox}
\begin{tcolorbox}[exAbstract,title={Abstract CoT}]
\small\abstart{}\atsep
\atok{T}\atsep\atok{C}\atsep\atok{F}\atsep\atok{L}\atsep\atok{AD}\atsep\atok{AE}\atsep\atok{T}\atsep\atok{F}\atsep\atok{C}\atsep\atok{L}\atsep\atok{W}\atsep\atok{T}\atsep\atok{F}\atsep\atok{BA}\atsep\atok{C}\atsep\atok{L}\atsep
\atok{T}\atsep\atok{BC}\atsep\atok{F}\atsep\atok{AD}\atsep\atok{C}\atsep\atok{L}\atsep\atok{T}\atsep\atok{F}\atsep\atok{AR}\atsep\atok{L}\atsep\atok{BD}\atsep\atok{T}\atsep\atok{C}\atsep\atok{F}\atsep\atok{L}\atsep\atok{C}\atsep
\atok{BE}\atsep\atok{T}\atsep\atok{F}\atsep\atok{C}\atsep\atok{AE}\atsep\atok{L}\atsep\atok{F}\atsep\atok{T}\atsep\atok{BG}\atsep\atok{C}\atsep\atok{L}\atsep\atok{T}\atsep\atok{F}\atsep\atok{BH}\atsep\atok{AD}\atsep\atok{C}\atsep
\atok{L}\atsep\atok{T}\atsep\atok{F}\atsep\atok{C}\atsep\atok{BI}\atsep\atok{W}\atsep\atok{L}\atsep\atok{T}\atsep\atok{F}\atsep\atok{C}\atsep\atok{L}\atsep\atok{T}\atsep\atok{F}\atsep\atok{AD}\atsep\atok{C}\atsep\atok{AE}\atsep
\atok{L}\atsep\atok{T}\atsep\atok{F}\atsep\atok{BJ}\atsep\atok{C}\atsep\atok{L}\atsep\atok{AD}\atsep\atok{T}\atsep\atok{F}\atsep\atok{C}\atsep\atok{W}\atsep\atok{L}\atsep\atok{T}\atsep\atok{F}\atsep\atok{BA}\atsep\atok{C}\atsep
\atok{L}\atsep\atok{AD}\atsep\atok{T}\atsep\atok{F}\atsep\atok{BK}\atsep\atok{C}\atsep\atok{L}\atsep\atok{T}\atsep\atok{AE}\atsep\atok{F}\atsep\atok{C}\atsep\atok{L}\atsep\atok{T}\atsep\atok{BL}\atsep\atok{F}\atsep\atok{AD}\atsep
\atok{C}\atsep\atok{L}\atsep\atok{T}\atsep\atok{F}\atsep\abend{}
\end{tcolorbox}
\begin{tcolorbox}[exAbsResponse,title={Abstract CoT Response}]
\small
Public-key cryptography lets two people exchange secrets over a completely open channel---no pre-shared password required. The core idea is a pair of mathematically linked keys: one \emph{public}, one \emph{private}.
\\ \\
\textbf{The Mailbox Analogy.}
Picture an open mailbox with a narrow slot. Anyone walking by can drop a letter in (that's the public key), but only you have the key to open the door and read what's inside (that's the private key). You publish the slot; you guard the key.
\\ \\
\textbf{Encryption in Practice.}
When Alice wants to message Bob privately, she encrypts it with Bob's public key. The result is gibberish to everyone---including Alice herself. Only Bob's private key can reverse the scramble. Even if an attacker captures every byte in transit, they learn nothing without the private key.
\\ \\
\textbf{Signatures: Authentication in Reverse.}
The same key pair works the other way around. Bob can ``sign'' a document with his private key, producing a tag that anyone can verify using his public key. A valid signature proves two things at once: the message came from Bob, and nobody altered it along the way.
\\ \\
\textbf{Why It Holds Up.}
Security comes from a mathematical one-way street. Multiplying two large primes is trivial; factoring their product back into primes is, for numbers of practical size, computationally infeasible. Current estimates suggest that breaking a standard 2048-bit key by brute-force factoring would take billions of years.
\\ \\
In short: one key locks, the other unlocks; one key signs, the other verifies. The math is easy forward, impossible backward---and that asymmetry is the entire foundation.
\end{tcolorbox}
\vspace{0.5em}\noindent\hdashrule{\linewidth}{0.4pt}{2pt 2pt}\vspace{0.3em}
\begin{tcolorbox}[exPrompt,title={Prompt}]
\textbf{Explain how public-key cryptography works to someone who understands basic algebra but has no computer science background.}
\end{tcolorbox}
\begin{tcolorbox}[exVerbal,title={Verbal CoT}]
\small
Alright, so I need to explain public-key cryptography to someone who knows basic algebra but isn’t familiar with computer science. Let’s start by thinking about the problem: How do two people communicate securely without sharing a secret key beforehand? In traditional methods, like symmetric encryption, both parties need to agree on a shared secret key, which is risky if someone intercepts it. Public-key cryptography solves this by using two keys---a public one and a private one---linked mathematically. But how to make this clear without jargon?
\\ \\
Maybe start with an analogy. Imagine a lock and key. If someone has the public key (the lock), they can lock a message, but only the private key (the key) can open it. But wait, that’s one part of the story. Public-key systems also allow for digital signatures: the private key can create a ``signature’’ that the public key can verify. That ensures authenticity, not just secrecy.
\\ \\
Now, the math. Public-key relies on mathematical problems that are easy to compute in one direction but hard to reverse. For example, multiplying two large prime numbers is straightforward, but factoring a huge number back into primes is extremely difficult. This is the core of RSA encryption. Let’s break it down step by step.
\\ \\
First, create a public key: take two large primes, multiply them to get a modulus (say, $N$), then choose an exponent ($e$) that’s coprime to $(p-1)(q-1)$. The public key is $(N,\,e)$. The private key is derived from the primes, usually an exponent ($d$) such that $d \cdot e \equiv 1 \pmod{\varphi(N)}$. The security comes from the difficulty of factoring $N$ into $p$ and $q$.
\\ \\
But how to explain this without getting bogged down in equations? Let’s use a simpler example. Suppose $N=15$ (from primes 3 and 5), and $e=7$. The public key is $(15,\,7)$. To encrypt a message, say ``2,’’ you compute $2^7 \bmod 15 = 128 \bmod 15 = 8$. To decrypt, you use $d=7$ (since $7 \times 7 \bmod 8 = 1$), so $8^7 \bmod 15 = 2097152 \bmod 15 = 2$. The math works here, but in reality, primes are huge (like 100 digits), making factoring impractical.
\\ \\
Digital signatures are another layer. If Bob wants to sign a message, he uses his private key to create a signature, and others can verify it with his public key. This proves the message came from Bob and hasn’t been altered.
\\ \\
But why is this secure? Because even if someone knows the public key $(N,\,e)$, factoring $N$ requires finding the primes $p$ and $q$, which is computationally infeasible for large $N$. This makes public-key cryptography reliable for secure communication and authentication.
\\ \\
Now, I need to structure this into a clear, step-by-step explanation, avoiding technical terms, using analogies, and ensuring the key points (asymmetric keys, one-way functions with trapdoors, encryption/signature processes, and security based on hard math problems) are covered without overwhelming the reader.
\end{tcolorbox}
\begin{tcolorbox}[exResponse,title={Verbal CoT Response}]
\small
Public-key cryptography is a method to securely exchange information over a public channel without needing to share a secret key beforehand. It uses two mathematically linked keys: a \emph{public key} and a \emph{private key}. Here’s how it works:
\\ \\
\textbf{The Key Pair.}
Think of the public key as a lock and the private key as the unique key that fits it. The public key is shared openly with everyone, while the private key is kept secret by the owner (like a treasure map only you have).
\\ \\
\textbf{Encrypting a Message.}
If Alice wants to send a secret message to Bob, she uses Bob’s public key to encrypt the message. This is like locking the message in a box with Bob’s lock. Only Bob’s private key can open this box. Even if someone intercepts the encrypted message, they can’t read it without Bob’s private key.
\\ \\
\textbf{Decrypting the Message.}
Once Bob receives the encrypted message, he uses his private key to unlock it. This is like using the key to open the lock and reveal the original message. The math behind this relies on a tricky problem: multiplying two large prime numbers is easy, but reversing the process (factoring a huge number back into primes) is extremely hard. This ensures the private key remains secure.
\\ \\
\textbf{Digital Signatures.}
Public-key cryptography also allows for digital signatures. If Bob wants to verify the authenticity of a message, he uses his private key to create a unique ``signature’’ for the message. Anyone with Bob’s public key can check this signature to confirm it was indeed sent by Bob and hasn’t been altered. This acts like a digital fingerprint.
\\ \\
\textbf{Why It’s Secure.}
The security hinges on a mathematical challenge. While it’s easy to multiply two large primes, factoring a huge number (like the modulus $N$ in the example) back into its prime components is computationally impossible with current technology. This makes it nearly impossible for anyone to derive the private key from the public key, even with powerful computers.
\\ \\
\textbf{In Summary.}
Public-key cryptography uses a pair of keys (public and private) to securely encrypt messages and verify identities. The public key is like a lock, the private key is the key. The system relies on mathematical problems that are easy to solve in one direction but nearly impossible to reverse, ensuring your secrets stay safe and your messages are authentic.
\end{tcolorbox}

\exsep

\subsubsection{Example 4: Social Communication}

\begin{tcolorbox}[exPrompt,title={Prompt}]
\textbf{I forgot my friend’s birthday yesterday. Help me write a heartfelt but not overly dramatic apology text.}
\end{tcolorbox}
\begin{tcolorbox}[exAbstract,title={Abstract CoT}]
\small\abstart{}\atsep
\atok{F}\atsep\atok{C}\atsep\atok{T}\atsep\atok{L}\atsep\atok{AD}\atsep\atok{F}\atsep\atok{BA}\atsep\atok{T}\atsep\atok{C}\atsep\atok{L}\atsep\atok{AE}\atsep\atok{F}\atsep\atok{T}\atsep\atok{R}\atsep\atok{C}\atsep\atok{L}\atsep
\atok{F}\atsep\atok{T}\atsep\atok{BB}\atsep\atok{C}\atsep\atok{L}\atsep\atok{AD}\atsep\atok{F}\atsep\atok{T}\atsep\atok{C}\atsep\atok{L}\atsep\atok{BC}\atsep\atok{AE}\atsep\atok{F}\atsep\atok{T}\atsep\atok{AD}\atsep\atok{C}\atsep
\atok{L}\atsep\atok{BD}\atsep\atok{F}\atsep\atok{S}\atsep\atok{T}\atsep\atok{BA}\atsep\atok{C}\atsep\atok{L}\atsep\atok{F}\atsep\atok{T}\atsep\atok{C}\atsep\atok{BE}\atsep\atok{AE}\atsep\atok{F}\atsep\atok{T}\atsep\atok{W}\atsep
\atok{BF}\atsep\atok{C}\atsep\atok{L}\atsep\atok{T}\atsep\atok{F}\atsep\atok{L}\atsep\atok{BH}\atsep\atok{C}\atsep\atok{T}\atsep\atok{AD}\atsep\atok{F}\atsep\atok{C}\atsep\atok{BI}\atsep\atok{L}\atsep\atok{T}\atsep\atok{BA}\atsep
\atok{BJ}\atsep\atok{F}\atsep\atok{T}\atsep\atok{W}\atsep\atok{C}\atsep\atok{L}\atsep\atok{AD}\atsep\atok{F}\atsep\atok{C}\atsep\atok{T}\atsep\atok{L}\atsep\atok{F}\atsep\abend{}
\end{tcolorbox}
\begin{tcolorbox}[exAbsResponse,title={Abstract CoT Response}]
\small
Hey [Friend's Name], I'm so sorry---I completely blanked on your birthday and I feel awful about it. No excuses, I just dropped the ball, and you deserved better than that.
\\ \\
You're one of the most important people in my life and I hate that I made you feel otherwise, even for a second. You've always been there for me, and the fact that I let your day slip by without even a text doesn't reflect how much I value you. I really am sorry.
\\ \\
I'd love to make it up to you---can I take you out for [coffee/dinner/your favourite thing] this week to properly celebrate? I want to hear all about your birthday and give you the celebration you actually deserve.
\\ \\
Happy belated birthday---you mean a lot to me and I'm really glad you're my friend.
\end{tcolorbox}
\vspace{0.5em}\noindent\hdashrule{\linewidth}{0.4pt}{2pt 2pt}\vspace{0.3em}
\begin{tcolorbox}[exPrompt,title={Prompt}]
\textbf{I forgot my friend’s birthday yesterday. Help me write a heartfelt but not overly dramatic apology text.}
\end{tcolorbox}
\begin{tcolorbox}[exVerbal,title={Verbal CoT}]
\small
When crafting an apology for forgetting a friend’s birthday, the goal is to balance sincerity with humility, avoiding both insincerity and over-the-top theatrics. The key is to acknowledge the mistake without defensiveness, express genuine remorse, and reaffirm the value of the friendship. Let’s break this down:
\\ \\
Start with a clear, direct apology. A vague or passive apology (``I’m sorry if you’re upset’’) might feel dismissive. Instead, take ownership of the mistake immediately. This shows accountability.
\\ \\
Acknowledge the impact of the mistake. It’s important to validate the friend’s feelings without making them feel like the error was intentional. Phrases like ``I completely understand if you’re upset’’ or ``I know how much this means to you’’ can help bridge the gap between your actions and their emotions.
\\ \\
Explain the mistake briefly (if needed). The user might want to add context---e.g., ``I’ve been swamped with work’’ or ``I’ve been juggling a lot lately.’’ However, this should be done sparingly. Over-explaining can shift the focus to your excuses rather than their feelings.
\\ \\
Reaffirm the significance of the friendship. Emphasize why the friend matters to you. This could include specific memories, shared values, or qualities the friend embodies. For example, ``You’ve always been there for me through [specific event], and I value that deeply.’’
\\ \\
Offer a gesture to make amends. A thoughtful plan to compensate---like a surprise celebration, a gift, or a dedicated time to catch up---shows you’re taking responsibility. Avoid grandiose gestures; instead, focus on sincerity.
\\ \\
Close with warmth and openness. Invite them to share their feelings or let you know how you can make things right. Ending with a lighthearted or affectionate remark can soften the tone.
\\ \\
Now, let’s piece this into a cohesive message:
\end{tcolorbox}
\begin{tcolorbox}[exResponse,title={Verbal CoT Response}]
\small
Hey [Friend’s Name],
\\ \\
I’ve been doing a lot of reflecting today, and I wanted to take a moment to sincerely apologize for forgetting your birthday. I know how much this day means to you, and I’m truly sorry I missed the chance to celebrate you properly. I completely understand if you’re upset---I know how much thought and effort goes into birthdays, and I feel awful for not even realizing it was coming.
\\ \\
Honestly, I’ve been so caught up in [briefly mention a reason, e.g., ``work deadlines’’ or ``personal obligations’’] that I completely overlooked the date. But I’m not excusing my mistake---I know it’s not an excuse. You’re someone who’s always been there for me, and I take that for granted. You’ve made such a positive impact on my life, and I want to make sure you know how much you mean to me.
\\ \\
I’d really like to make it up to you. Whether it’s a surprise celebration, a thoughtful gift, or just a heartfelt conversation, I’m happy to do whatever makes you feel appreciated. If there’s anything specific you’d like, please let me know---I’m all ears.
\\ \\
I know words can’t fully undo this, but I hope you can forgive me. I value our friendship deeply, and I’d hate for this to strain anything between us. Let’s catch up soon---I’d love to hear your thoughts and maybe plan something special.
\\ \\
Thanks for being you, and for putting up with my occasional slip-ups.
\\ \\
Warmly,\\
{[Your Name]}
\end{tcolorbox}

\end{document}